\documentclass[11pt]{article}

\usepackage[]{acl}
\usepackage{algorithm}
\usepackage{subcaption}
\usepackage{graphicx}
\usepackage{amssymb}
\usepackage{amsthm}
\usepackage{hyperref}
\usepackage{booktabs}
\usepackage{url}
\usepackage{xcolor}
\usepackage[table]{xcolor}
\usepackage{listings}
\usepackage{float}
\usepackage{amsmath}
\usepackage{times}
\usepackage{latexsym}
\usepackage{orcidlink}
\usepackage[table]{xcolor}   
\usepackage{graphicx}        
\usepackage{textcomp}        
\usepackage{booktabs}
\definecolor{heat}{HTML}{2A78D6}
\usepackage[most]{tcolorbox}

\usepackage[T1]{fontenc}

\usepackage[utf8]{inputenc}
\usepackage{multirow}
\usepackage{microtype}

\usepackage{inconsolata}

\usepackage{graphicx}

\title{Beyond Single-Turn Confidence: Trajectory-Adapted Uncertainty Quantification for LLM Agents}

\author{ Dylan Bouchard\thanks{Correspondence to \texttt{dbouchard92@gmail.com}}  \orcidlink{0009-0004-9233-2324}\qquad Mohit Singh Chauhan \orcidlink{0000-0002-7817-0427}
}

\begin{document}

\maketitle

\begin{abstract}
Uncertainty quantification (UQ) methods for language models are typically
evaluated on single-turn outputs, where uncertainty is attached to one
generated answer. For LLM agents, however, the unit of observation is an
interactive trajectory, where the model can ask clarifying questions, call tools,
update state, and make intermediate decisions whose errors propagate to the
final outcome. We study whether three common families of single-turn UQ methods
transfer to this setting. Across five LLMs and four multi-turn
tool-use datasets from BFCL-v4 and $\tau^2$-bench, we evaluate white-box scorers
based on action-token probabilities, black-box consistency scorers based on
resampled trajectories, and reflexive scorers based on model self-assessment of
the trajectory. We find that transfer is often useful but uneven. Token-probability
scores are highly sensitive to the choice of aggregator used across turns,
reflexive scores provide the strongest low-cost baseline in most evaluated
settings, and black-box self-consistency is often the strongest UQ
family, with trajectory-equivalence and action-set consistency typically
ranking highest among its variants. These results suggest that UQ methods developed for single
generations should be revalidated at the trajectory level, with careful
attention to the consistency measurement, aggregator choice, and computational budget.
\end{abstract}

\section{Introduction}
Uncertainty quantification (UQ) methods have been widely shown to classify
the factual correctness of large language model responses in single-turn
settings \citep{kadavath2022languagemodelsmostlyknow,
kuhn2023semanticuncertaintylinguisticinvariances,
manakul2023selfcheckgptzeroresourceblackboxhallucination, Farquhar2024,
lin2024generatingconfidenceuncertaintyquantification}. Meanwhile, language
models are increasingly deployed as agents that interact with users, call
external tools, update intermediate state, and take multiple decisions for a single task. 
Along an agentic trajectory, uncertainty arises at many points, including whether a
request is underspecified, which action or tool to choose, how to fill
arguments, and whether the trajectory achieved the user's goal. Quantifying
uncertainty for these systems therefore requires more than confidence in a single-generation response.

Single-turn UQ methods generally fall into three families: white-box scorers
that use token probabilities, black-box consistency scorers that
sample multiple responses and measure self-consistency, and reflexive scorers that
ask a model to assess its own confidence or correctness. These scorers may
degrade when uncertainty accumulates over a trajectory, when an early mistake
propagates into later decisions, or when the model must choose among answering,
clarifying, and acting. Prior work on agent UQ has proposed propagation methods such as SAUP and UProp
\citep{zhao-etal-2025-uncertainty,
duan2025upropinvestigatinguncertaintypropagation}, while recent work has
formalized the problem with trajectory-adapted notation and terminology
\citep{oh2026uncertaintyquantificationllmagents}. What remains less clear is
whether these three UQ families still provide useful failure signals once the
prediction unit is an interactive trajectory. We study this question through
experiments spanning five LLMs and four agentic datasets from 
BFCL-v4 \citep{patil2025the} and $\tau^2$-bench
\citep{barres2025tau2benchevaluatingconversationalagents}, evaluating a broad suite of UQ scorers for their
ability to predict trajectory success across models, domains, and
selective-prediction budgets.

Our contributions are as follows. First, we provide a controlled empirical
comparison of these UQ families in multi-turn tool-use agents, reporting
discrimination, calibration, and selective prediction. Second, we adapt 
them to trajectory-level scoring through cross-turn aggregation of 
token-probability scores and new measures of black-box consistency, 
including a model-based trajectory-equivalence scorer.
 Third, we find that transferability is mixed 
across scoring methods. Specifically, in our evaluated settings, white-box reliability depends strongly on the choice of
aggregator,  reflexive scorers work well for most model-dataset pairs considered, and black-box self-consistency is often the strongest family,
with trajectory-equivalence rate and action-set consistency typically ranking
highest among its variants. Importantly, no family is
uniformly reliable across all models and domains in our evaluations.

\section{Related Work}

\paragraph{Confidence Estimation for Language Model Outputs.}
Prior work has developed several families of post-hoc uncertainty scores for language model generations. Output-consistency methods sample the model multiple times and estimate uncertainty from disagreement among generations, using exact agreement, lexical overlap, embedding similarity, semantic clustering, or entailment-based equivalence \citep{cole-etal-2023-selectively, manakul2023selfcheckgptzeroresourceblackboxhallucination,bouchard2025uncertainty, zhang2020bertscoreevaluatingtextgeneration, chen2023quantifyinguncertaintyanswerslanguage, kuhn2023semanticuncertaintylinguisticinvariances, lin2024generatingconfidenceuncertaintyquantification, Farquhar2024}. Long-form variants apply similar techniques at the claim level \citep{jiang2024graphbaseduncertaintymetricslongform, bouchard2026finegraineduncertaintyquantificationlongform, zhang2024luqlongtextuncertaintyquantification, zhang2025atomiccalibrationllmslongform}. Probability-based methods use access to model likelihoods and aggregate token-level quantities into response-level scores, such as sequence probability, perplexity, entropy, probability margins, or related summaries of the predictive distribution \citep{malinin2021uncertaintyestimationautoregressivestructured, fadeeva2024factcheckingoutputlargelanguage, farr2024redctsystemsdesignmethodology}. Reflexive methods instead ask the model itself to report confidence or assess whether the generated output is correct \citep{kadavath2022languagemodelsmostlyknow, tian2023justaskcalibrationstrategies, xiong2024llmsexpressuncertaintyempirical}. These methods are most often evaluated on static prediction tasks, where uncertainty is attached to a single generated response. Agentic systems require a different unit of analysis because trajectory success depends on a sequence of actions and interactions rather than only a single generation.

\paragraph{Uncertainty in Interactive and Tool-Using Agents.}
A growing body of work studies uncertainty in LLM agents. UALA incorporates uncertainty into the thought-action-observation loop and uses it to guide tool use \citep{han-etal-2024-towards}. ProbeCal calibrates tool-using agents using execution traces and embedding-based probes \citep{liu-etal-2024-uncertainty}. Other work studies uncertainty propagation across multi-step trajectories. SAUP propagates step-level uncertainty through agent reasoning traces using situation-dependent weights \citep{zhao-etal-2025-uncertainty}, while UProp separates uncertainty intrinsic to the current decision from uncertainty inherited from earlier steps \citep{duan2025upropinvestigatinguncertaintypropagation}. Multi-turn confidence estimation has also been studied in dialogue settings, with per-turn calibration and monotonicity as information accumulates being the key desiderata \citep{zhang2026confidenceestimationllmsmultiturn}. Recent agent-specific work further treats uncertainty as a control signal: Agentic UQ uses verbalized uncertainty to guide memory and reflection \citep{zhang2026agenticuncertaintyquantification}, a tool-use calibration study finds that evidence tools induce overconfidence while verification tools can mitigate miscalibration \citep{xuan2026confidencedichotomyanalyzingmitigating}, and clarification-oriented agents use structured uncertainty over tool-call parameters to decide when to ask the user for missing information \citep{suri2026structureduncertaintyguidedclarification}. Recent foundation-oriented work formalizes trajectory-level agent uncertainty as joint uncertainty over the full trajectory, with contributions from initial-query, action, and observation uncertainty \citep{oh2026uncertaintyquantificationllmagents}.
While these studies establish uncertainty as central for LLM agents, to our knowledge, prior work has not systematically compared probability-based, sampling-based, and reflexive methods under matched conditions across a range of LLM agents and datasets.

\section{Methods}
\label{sec:methods}

We evaluate UQ-based confidence scorers spanning white-box,
black-box, and reflexive families as predictors of agentic trajectory success.
We build on the agent-UQ formulation of
\citet{oh2026uncertaintyquantificationllmagents}, where an episode is a trajectory
$F_{\leq T}$ of $T$ turns in which turn $i$ emits an action $A_i$ (a tool call
or a message to the user) and receives an environment observation $O_i$. Upon completion of the trajectory, a
binary reward $r(F_{\leq T})$ records success. We write $\mathcal{F}$ for the
space of such trajectories and, for notational simplicity, abbreviate
$F_{\leq T}$ to $F$. We aim to estimate the uncertainty of the full trajectory $F$, where a useful uncertainty
score should rank successful trajectories above failed ones.\footnote{Following \citet{oh2026uncertaintyquantificationllmagents}, we do not measure observation uncertainty in this work.}

Let $y$ be a generation for prompt $x$ with tokens
$\{t_1, \ldots, t_L\}$, where $L$ is the token count and $p_j$ is the
probability of token $t_j$ conditional on the previous tokens in the generated sequence. At turn $i$ this generation is the action $A_i$.
We treat each confidence scorer as a map $C: \mathcal{F} \to [0,1]$, with
larger values indicating higher confidence that the trajectory succeeded.
In particular, white-box scorers are computed from per-turn token probabilities and aggregated over the
trajectory, black-box self-consistency scorers sample $m$ additional candidate trajectories
$\tilde{\mathbf{F}} = \{\tilde{F}_1, \ldots, \tilde{F}_m\}$ under stochastic decoding and measure trajectory consistency, extending the signature to
$C: \mathcal{F} \times \mathcal{F}^m \to [0,1]$, and reflexive scorers use self-evaluation on the transcript prefix available at the final agent action.
We detail various scorers from each family below.

\subsection{White-Box Scorers}

White-box scorers use the model's token probabilities to compute confidence on
a single generation. We generalize these methods to trajectory-level scoring using various cross-turn aggregators.

\subsubsection{Single-Generation Scorers}

Here we define three single-generation white-box scorers from the literature, with additional scorers defined in Appendix~\ref{app:whitebox-scorers}. 
Sequence probability $\text{SP}(y) =
\prod_{j=1}^{L} p_j$ \citep{Vashurin_2025} is computed as the product of token probabilities in a generated sequence. The normalized version, called Length Normalized Sequence Probability,
$\text{LNSP}(y) = (\prod_{j=1}^{L} p_j)^{1/L}$, is a per-token geometric mean that
does not directly penalize longer generations
\citep{malinin2021uncertaintyestimationautoregressivestructured,bouchard2025uncertainty}. The third single-turn
scorer uses the top-$K$ token probabilities $\{p_{j,1}, \ldots, p_{j,K}\}$ at each
position, normalized to sum to 1. Average token negentropy averages the per-position
entropy-based confidence $\text{TN@}K(t_j) = 1 - \text{TE@}K(t_j)/\log K$ over the generation
\citep{bouchard2025uqlmpythonpackageuncertainty},
where $\text{TE@}K(t_j) = -\sum_{k=1}^{K} p_{j,k} \log p_{j,k}$ is the truncated
top-$K$ entropy \citep{scalena2025eagerentropyawaregenerationadaptive, manakul2023selfcheckgptzeroresourceblackboxhallucination}:
$$\text{ATN@}K(y) = \frac{1}{L} \sum_{j=1}^{L} \text{TN@}K(t_j).$$

\subsubsection{Token Span and Trajectory Aggregation}
\label{subsec:aggregation}

Applying the scorers above along a trajectory requires two agentic choices left
open by the framework: 1) which tokens of a turn to score, and 2) how to combine the
per-turn values into a trajectory-level estimate.

\paragraph{Token span.} The framework from \citet{oh2026uncertaintyquantificationllmagents} treats the action $A_i$ atomically, but a
turn's generation interleaves free-form reasoning with the committed action.
All white-box scores in this work are computed over the action span of each
turn (i.e., the tokens of the committed action, a tool call or a message to the
user).\footnote{Variants that score the reasoning tokens, or
reasoning and action together, are left to future work.}

\paragraph{Cross-turn aggregation.} \citet{oh2026uncertaintyquantificationllmagents} express trajectory
uncertainty as a generalized aggregation of these per-turn uncertainty scores, encompassing special cases such as a maximum or minimum over turns,  arithmetic mean,
and convex weightings
\citep{zhao-etal-2025-uncertainty, duan2025upropinvestigatinguncertaintypropagation}. Letting $s_i$ denote the per-turn confidence of a
white-box scorer at turn $i$, we evaluate the
confidence-side counterparts, including the mean $g_{\text{mean}} = \frac{1}{T}\sum_{i} s_i$
(the length-normalized reduction), the minimum $g_{\text{min}} = \min_{i} s_i$
(the confidence counterpart of the most-uncertain-step rule), and the first-
and last-turn values $g_{\text{first}} = s_1$ and $g_{\text{last}} = s_T$. A
general position-weighted aggregator with early or late weights is defined in
Appendix~\ref{app:whitebox-scorers}. We note that under $g_{\text{mean}}$, LNSP and ATN@$K$
are the confidence analogs of the two trajectory-level white-box
scorers evaluated in \citet{oh2026uncertaintyquantificationllmagents}.

\subsection{Black-Box Consistency Scorers}

Black-box scorers sample $m$ additional trajectories and measure
agreement between the original and sampled trajectories, without access to token probabilities. Focusing on where consistency measurements are well-defined, we consider consistency over the
final message, over the action sequence, and over the full trajectory text.

\subsubsection{Final-Message Consistency}

The agent's closing message is the closest agentic analogue of a single-turn
answer, but in a tool-using agent it is often only a self-report of
reward-relevant actions. With this in mind, we aim to assess whether standard
single-turn consistency UQ transfers to agents directly off the shelf. In particular, we compare the reference
final message $y$ with those of the resampled trajectories,
$\tilde{\mathbf{y}} = \{\tilde{y}_1, \ldots, \tilde{y}_m\}$, exactly as it
would compare resampled answers to a single-turn question.\footnote{When a trajectory
ends without a closing message, its last user-directed message is used.} The exact scorer definition is provided below.

\paragraph{Non-Contradiction Probability (NCP).} The mean probability that the reference and a sample do not contradict, averaged over both directions of an NLI model \citep{chen2023quantifyinguncertaintyanswerslanguage}:
$$\text{NCP}(y; \tilde{\mathbf{y}}) = 1 - \frac{1}{m} \sum_{j=1}^{m} \frac{p_c(y, \tilde{y}_j) + p_c(\tilde{y}_j, y)}{2}$$
where $p_c(\cdot, \cdot)$ is the NLI contradiction probability.

\subsubsection{Action-Structured Consistency}

Final-message consistency ignores the agent's action structure, which is often
a major determinant of trajectory reward. To address this shortcoming, we consider consistency over that structure, comparing
the $m$ resampled trajectories $\tilde{\mathbf{F}}$ to the reference $F$
through their actions. Let $\kappa(A)$ be the
action type (i.e., the agent's choice among tools or a user message). For a trajectory $F$ with $T_F$ actions, let
$\sigma_F = (\kappa(A_1^F), \ldots, \kappa(A_{T_F}^F))$ be its
ordered action-type sequence, $S_F$ its set of unique action types chosen, and $q_F$
its action-type distribution, $q_F(v) = c_F(v) / \sum_{v'} c_F(v')$,
with $c_F(v)$ the count of type $v$.

\paragraph{First-Action Consistency (FAC).} Since sampled trajectories must start from the same prompt, this scorer guarantees prefix-aligned consistency measurement. Specifically, FAC computes a first-action match rate between original and sampled trajectories:
$$\text{FAC}(F; \tilde{\mathbf{F}}) = \frac{1}{m} \sum_{j=1}^{m} \mathbb{I}\!\left(\kappa(A_1^{\tilde{F}_j}) = \kappa(A_1^{F})\right).$$

\paragraph{Action-Set Consistency (ASC).} A trajectory-wide measure, this scorer computes the mean Jaccard similarity of the action-type sets, which is invariant to order and repetition:
$$\text{ASC}(F; \tilde{\mathbf{F}}) = \frac{1}{m} \sum_{j=1}^{m} \frac{|S_F \cap S_{\tilde{F}_j}|}{|S_F \cup S_{\tilde{F}_j}|}.$$

\paragraph{Action-Distribution Consistency (ADC).} This scorer measures agreement between the distributions of action frequencies of sampled trajectories, measured as one minus the mean Jensen-Shannon divergence between action-type distributions (base-2 logarithm, so $\text{JSD} \in [0,1]$). Because the distributions are normalized, the score reflects the usage profile rather than trajectory length, and unlike ASC it retains repetition:
$$\text{ADC}(F; \tilde{\mathbf{F}}) = 1 - \frac{1}{m} \sum_{j=1}^{m} \text{JSD}\!\left(q_F \,\|\, q_{\tilde{F}_j}\right).$$

\paragraph{Action-Edit Consistency (AEC).} Unlike ASC and ADC, AEC is sensitive to the order in which actions are taken and to differences in trajectory length. It is computed as one minus the mean normalized Levenshtein edit distance between the ordered action-type sequences:
$$\text{AEC}(F; \tilde{\mathbf{F}}) = 1 - \frac{1}{m} \sum_{j=1}^{m} \frac{\text{ED}\!\left(\sigma_F, \sigma_{\tilde{F}_j}\right)}{\max\!\left(T_F,\, T_{\tilde{F}_j}\right)},$$
where $\text{ED}(\cdot, \cdot)$ is the Levenshtein edit distance over two action-type sequences. The distance is normalized by the length of the longer sequence so the score lies in $[0,1]$.

\subsubsection{Judged Consistency}

The action-structured scorers compare action types directly, so they ignore tool arguments and may also penalize valid alternative action structures.
Our last consistency scorer replaces the direct action comparison with a
model-based trajectory-equivalence judgment. In particular, a separate LLM judge $J$ reads two full trajectories and decides
whether they reach the same task-relevant outcome. To avoid information leakage, the trajectory-equivalence judge
 is shown only the
trajectory transcripts and never observes the hidden task instructions or the reward. Importantly, the judge assesses outcome equivalence rather
than correctness, so TER remains a consistency measure rather than an explicit LLM-as-a-judge correctness score. Its prompt is
provided in Table~\ref{tab:ter-judge-prompt}.

\paragraph{Trajectory Equivalence Rate (TER).} This scorer measures the fraction of resampled trajectories the judge deems outcome-equivalent to the reference trajectory:
$$\text{TER}(F; \tilde{\mathbf{F}}) = \frac{1}{m} \sum_{j=1}^{m} \mathbb{I}[F \equiv \tilde{F}_j ],$$
where $\equiv$ denotes an equivalence judgment.
\subsection{Reflexive Scorers}
\label{sec:reflexive}
These scorers ask the model to judge its own trajectory. Each is applied once to
the trajectory transcript rather than turn by turn. Specifically, the model receives the transcript prefix ending at the agent's
final action, together with the same domain policy and tool schemas available
to the agent. This boundary excludes any subsequent user confirmation, so the
scorer sees only information available when the final action was taken.\footnote{The
hidden task specification is withheld to prevent the scorer from seeing the
benchmark's acceptance criteria.} The prompts are
provided in Appendix~\ref{app:prompts}.

\paragraph{P(True).} This scorer asks the model whether the trajectory succeeded, expressed as ``True'' or ``False.'' The confidence score is the probability assigned to the ``True'' token \citep{kadavath2022languagemodelsmostlyknow}.

\paragraph{Verbalized Confidence (VC).} Without access to token probabilities, this scorer asks the model for a Yes/No determination of whether the trajectory succeeded together with the probability, from 0 to 1, that its guess is correct \citep{tian2023justaskcalibrationstrategies, xiong2024llmsexpressuncertaintyempirical}. The confidence is the stated probability when the guess is ``Yes'' and its complement when the guess is ``No''.

\section{Experiments}

\subsection{Experimental Setup}
\label{sec:setup}

We evaluate on four multi-turn tool-use datasets: the multi-turn subset of BFCL-v4 
\citep{patil2025the} (200 tasks) and the retail, airline, and telecom text
datasets from $\tau^2$-bench \citep{barres2025tau2benchevaluatingconversationalagents}
(114, 50, and 114 tasks, respectively). Each dataset scores a completed
trajectory with a binary success reward based on state, response,
or required-action checks. We use this reward as the trajectory-success label
and detail per-dataset grading in Appendix~\ref{app:grading}.

The system prompt contains the available tool schemas and any benchmark-provided domain policy. The agent emits one action as text per turn, making the same action-token surface observable across models.
 We provide more detail on
this interface, its rationale, and a tool-calling ablation in
Appendix~\ref{app:native-tool-calling}.
For $\tau^2$-bench, the simulated user is fixed to \texttt{gpt-4.1-mini} at
temperature 0,  minimizing
simulator-side sampling variance. NCP uses
\texttt{microsoft/deberta-large-mnli} for the NLI model, following past work \citep{kuhn2023semanticuncertaintylinguisticinvariances,chen2023quantifyinguncertaintyanswerslanguage,lin2024generatingconfidenceuncertaintyquantification}. For TER, which requires an auxiliary
LLM judge, we use \texttt{gemini-flash-lite}.

We evaluate our suite of UQ scorers on five models spanning a wide range of competence:
\texttt{Qwen2.5-7B}, \texttt{gpt-oss-20b}, \texttt{Qwen3.5-9B}, and \texttt{MiniMax-M3} served by
Together AI and \texttt{gpt-4o-mini} served by OpenAI. For each task we record a
greedy trajectory (the reference run with temperature 0) with $K=5$ top logprobs per
token, plus three sampled trajectories at temperature 0.7 for black-box
consistency measurement.\footnote{Typical UQ sampling uses temperature 1.0 \citep{bouchard2025uncertainty, chen2023quantifyinguncertaintyanswerslanguage}. Given the multi-turn stochastic nature of trajectories, we opted for a slightly lower temperature.} In the main results, every scorer
is evaluated against the greedy trajectory's success label.  In addition to the scorers defined in Section~\ref{sec:methods},
we evaluate SAUP-inspired step-level propagation controls adapted to our
action-level traces, detailed in Appendix~\ref{app:saup}.\footnote{We treat these as
controls rather than a direct SAUP reproduction because the original method is
defined for ReAct-style thought--action--observation trajectories and, for its
learned variant, situational-state supervision that is not available in
our benchmark traces.}
Each uncertainty estimator (Section~\ref{sec:methods}) is
evaluated by how well it classifies trajectory success. We use AUROC as the
primary metric, with bootstrap intervals,
and also report AUPRC, selective-prediction risk-coverage,
and expected calibration error (ECE).

\subsection{Results}
\label{sec:results}

Table~\ref{tab:auroc-main} reports AUROC for predicting trajectory success on
BFCL-v4 and the three $\tau^2$-bench datasets, while Appendix~\ref{app:full-results} reports the results for the complete grid of white-box scorers. 
Scorer rankings under AUPRC and the
prediction rejection ratio (PRR) agree closely with AUROC (median Spearman
rank correlation 0.94 with AUPRC and 0.95 with PRR across model-dataset
cells), so we use AUROC as the primary metric and report PRR and ECE in the appendices (Tables~\ref{tab:prr} and~\ref{tab:ece}). 
Figure~\ref{fig:method-family-prr} in the
appendix summarizes the highest-PRR scorer in each family for every
model-dataset cell.

The main table reports AUROC point estimates, while Table~\ref{tab:auroc-main-ci} repeats
each cell with a 95\% task-level bootstrap interval. The intervals are often
wide, with median half-width 0.10 in columns with at least 20 minority-class
trajectories. We therefore treat within-column rankings as descriptive and focus below on recurring
patterns and the paired comparisons in
Appendix~\ref{app:statistical-uncertainty}.

\begin{table*}[t]
\centering
\scriptsize
\setlength{\tabcolsep}{1.2pt}
\resizebox{\textwidth}{!}{%
\begin{tabular}{lcccccccccccccccccccc}
\toprule
 & \multicolumn{5}{c}{BFCL-v4} & \multicolumn{5}{c}{Airline} & \multicolumn{5}{c}{Telecom} & \multicolumn{5}{c}{Retail} \\
\cmidrule(lr){2-6}
\cmidrule(lr){7-11}
\cmidrule(lr){12-16}
\cmidrule(lr){17-21}
Scorer & \rotatebox{90}{\texttt{gpt-4o-mini}} & \rotatebox{90}{\texttt{gpt-oss-20b}} & \rotatebox{90}{\texttt{Qwen2.5-7B}} & \rotatebox{90}{\texttt{Qwen3.5-9B}} & \rotatebox{90}{\texttt{MiniMax-M3}} & \rotatebox{90}{\texttt{gpt-4o-mini}} & \rotatebox{90}{\texttt{gpt-oss-20b}} & \rotatebox{90}{\texttt{Qwen2.5-7B}} & \rotatebox{90}{\texttt{Qwen3.5-9B}} & \rotatebox{90}{\texttt{MiniMax-M3}} & \rotatebox{90}{\texttt{gpt-4o-mini}} & \rotatebox{90}{\texttt{gpt-oss-20b}} & \rotatebox{90}{\texttt{Qwen2.5-7B}} & \rotatebox{90}{\texttt{Qwen3.5-9B}} & \rotatebox{90}{\texttt{MiniMax-M3}} & \rotatebox{90}{\texttt{gpt-4o-mini}} & \rotatebox{90}{\texttt{gpt-oss-20b}} & \rotatebox{90}{\texttt{Qwen2.5-7B}} & \rotatebox{90}{\texttt{Qwen3.5-9B}} & \rotatebox{90}{\texttt{MiniMax-M3}} \\
\midrule
SP$_{\mathrm{first}}$ & \cellcolor{heat!47}0.66 & \cellcolor{heat!39}0.53 & \cellcolor{heat!39}0.53 & \cellcolor{heat!31}0.52 & \cellcolor{heat!54}0.69 & \cellcolor{heat!31}0.53 & \cellcolor{heat!39}0.56 & \cellcolor{heat!31}0.50 & \cellcolor{heat!16}0.33 & \cellcolor{heat!31}0.51 & \cellcolor{heat!47}0.61 & \cellcolor{heat!31}0.45 & \cellcolor{heat!47}0.64 & \cellcolor{heat!8}0.23 & \cellcolor{heat!39}0.57 & \cellcolor{heat!39}0.60 & \cellcolor{heat!39}0.55 & \cellcolor{heat!39}0.53 & \cellcolor{heat!39}0.56 & \cellcolor{heat!31}0.52 \\
SP$_{\mathrm{mean}}$ & \cellcolor{heat!47}0.62 & \cellcolor{heat!31}0.52 & \cellcolor{heat!47}0.65 & \cellcolor{heat!39}0.59 & \cellcolor{heat!47}0.66 & \cellcolor{heat!31}0.47 & \cellcolor{heat!31}0.52 & \cellcolor{heat!16}0.30 & \cellcolor{heat!31}0.47 & \cellcolor{heat!16}0.33 & \cellcolor{heat!23}0.38 & \cellcolor{heat!16}0.36 & \cellcolor{heat!31}0.52 & \cellcolor{heat!54}0.73 & \cellcolor{heat!23}0.43 & \cellcolor{heat!47}0.64 & \cellcolor{heat!16}0.32 & \cellcolor{heat!31}0.48 & \cellcolor{heat!39}0.56 & \cellcolor{heat!39}0.56 \\
SP$_{\mathrm{min}}$ & \cellcolor{heat!39}0.59 & \cellcolor{heat!31}0.45 & \cellcolor{heat!47}0.62 & \cellcolor{heat!39}0.60 & \cellcolor{heat!39}0.54 & \cellcolor{heat!23}0.45 & \cellcolor{heat!39}0.58 & \cellcolor{heat!31}0.48 & \cellcolor{heat!39}0.59 & \cellcolor{heat!39}0.59 & \cellcolor{heat!8}0.24 & \cellcolor{heat!47}0.61 & \cellcolor{heat!47}0.63 & \cellcolor{heat!31}0.50 & \cellcolor{heat!16}0.36 & \cellcolor{heat!39}0.57 & \cellcolor{heat!39}0.55 & \cellcolor{heat!31}0.49 & \cellcolor{heat!39}0.60 & \cellcolor{heat!31}0.50 \\
SP$_{\mathrm{last}}$ & \cellcolor{heat!39}0.53 & \cellcolor{heat!23}0.39 & \cellcolor{heat!31}0.51 & \cellcolor{heat!39}0.55 & \cellcolor{heat!31}0.51 & \cellcolor{heat!39}0.56 & \cellcolor{heat!31}0.46 & \cellcolor{heat!39}\textbf{0.58} & \cellcolor{heat!39}0.58 & \cellcolor{heat!47}0.62 & \cellcolor{heat!47}0.63 & \cellcolor{heat!31}0.50 & \cellcolor{heat!31}0.49 & \cellcolor{heat!23}0.42 & \cellcolor{heat!39}0.59 & \cellcolor{heat!39}0.60 & \cellcolor{heat!23}0.42 & \cellcolor{heat!31}0.51 & \cellcolor{heat!31}0.46 & \cellcolor{heat!31}0.48 \\
LNSP$_{\mathrm{first}}$ & \cellcolor{heat!47}0.66 & \cellcolor{heat!39}0.56 & \cellcolor{heat!39}0.54 & \cellcolor{heat!31}0.53 & \cellcolor{heat!47}0.65 & \cellcolor{heat!39}0.54 & \cellcolor{heat!39}0.55 & \cellcolor{heat!31}0.53 & \cellcolor{heat!16}0.33 & \cellcolor{heat!39}0.55 & \cellcolor{heat!54}0.70 & \cellcolor{heat!31}0.47 & \cellcolor{heat!39}0.57 & \cellcolor{heat!8}0.23 & \cellcolor{heat!39}0.58 & \cellcolor{heat!47}0.61 & \cellcolor{heat!39}0.59 & \cellcolor{heat!31}0.52 & \cellcolor{heat!39}0.59 & \cellcolor{heat!31}0.53 \\
LNSP$_{\mathrm{mean}}$ & \cellcolor{heat!54}0.70 & \cellcolor{heat!31}0.52 & \cellcolor{heat!47}0.68 & \cellcolor{heat!39}0.60 & \cellcolor{heat!47}0.65 & \cellcolor{heat!23}0.41 & \cellcolor{heat!31}0.52 & 0.17 & \cellcolor{heat!16}0.34 & \cellcolor{heat!8}0.25 & \cellcolor{heat!16}0.37 & \cellcolor{heat!23}0.39 & \cellcolor{heat!31}0.53 & \cellcolor{heat!16}0.34 & \cellcolor{heat!23}0.42 & \cellcolor{heat!54}0.69 & \cellcolor{heat!31}0.47 & \cellcolor{heat!31}0.50 & \cellcolor{heat!39}0.60 & \cellcolor{heat!31}0.50 \\
LNSP$_{\mathrm{min}}$ & \cellcolor{heat!47}0.68 & \cellcolor{heat!23}0.45 & \cellcolor{heat!47}0.65 & \cellcolor{heat!47}0.62 & \cellcolor{heat!47}0.64 & \cellcolor{heat!23}0.45 & \cellcolor{heat!39}0.59 & \cellcolor{heat!23}0.39 & \cellcolor{heat!23}0.37 & \cellcolor{heat!23}0.39 & \cellcolor{heat!8}0.24 & \cellcolor{heat!39}0.60 & \cellcolor{heat!39}0.59 & \cellcolor{heat!31}0.51 & \cellcolor{heat!23}0.37 & \cellcolor{heat!54}0.72 & \cellcolor{heat!39}0.59 & \cellcolor{heat!39}0.55 & \cellcolor{heat!39}0.59 & \cellcolor{heat!31}0.52 \\
LNSP$_{\mathrm{last}}$ & \cellcolor{heat!39}0.56 & \cellcolor{heat!16}0.36 & \cellcolor{heat!31}0.51 & \cellcolor{heat!39}0.56 & \cellcolor{heat!31}0.52 & \cellcolor{heat!39}0.55 & \cellcolor{heat!31}0.50 & \cellcolor{heat!31}0.48 & \cellcolor{heat!31}0.49 & \cellcolor{heat!31}0.51 & \cellcolor{heat!47}0.64 & \cellcolor{heat!31}0.47 & \cellcolor{heat!23}0.43 & \cellcolor{heat!23}0.43 & \cellcolor{heat!23}0.45 & \cellcolor{heat!47}0.67 & \cellcolor{heat!23}0.44 & \cellcolor{heat!39}0.53 & \cellcolor{heat!23}0.44 & \cellcolor{heat!31}0.52 \\
ATN@5$_{\mathrm{first}}$ & \cellcolor{heat!47}0.67 & \cellcolor{heat!39}0.55 & \cellcolor{heat!39}0.55 & \cellcolor{heat!39}0.53 & \cellcolor{heat!47}0.64 & \cellcolor{heat!39}0.54 & \cellcolor{heat!39}0.54 & \cellcolor{heat!39}0.53 & \cellcolor{heat!16}0.32 & \cellcolor{heat!39}0.54 & \cellcolor{heat!47}0.62 & \cellcolor{heat!31}0.47 & \cellcolor{heat!39}0.61 & 0.20 & \cellcolor{heat!39}0.57 & \cellcolor{heat!39}0.60 & \cellcolor{heat!39}0.58 & \cellcolor{heat!39}0.56 & \cellcolor{heat!39}0.60 & \cellcolor{heat!39}0.54 \\
ATN@5$_{\mathrm{mean}}$ & \cellcolor{heat!54}0.71 & \cellcolor{heat!31}0.50 & \cellcolor{heat!54}0.70 & \cellcolor{heat!47}0.62 & \cellcolor{heat!47}0.66 & \cellcolor{heat!23}0.41 & \cellcolor{heat!31}0.50 & 0.18 & \cellcolor{heat!16}0.36 & \cellcolor{heat!8}0.24 & \cellcolor{heat!23}0.38 & \cellcolor{heat!23}0.38 & \cellcolor{heat!39}0.54 & \cellcolor{heat!16}0.35 & \cellcolor{heat!23}0.43 & \cellcolor{heat!47}0.68 & \cellcolor{heat!23}0.44 & \cellcolor{heat!31}0.49 & \cellcolor{heat!39}0.60 & \cellcolor{heat!31}0.49 \\
ATN@5$_{\mathrm{min}}$ & \cellcolor{heat!47}0.68 & \cellcolor{heat!23}0.42 & \cellcolor{heat!47}0.65 & \cellcolor{heat!47}0.63 & \cellcolor{heat!47}0.65 & \cellcolor{heat!23}0.45 & \cellcolor{heat!39}0.59 & \cellcolor{heat!23}0.37 & \cellcolor{heat!23}0.38 & \cellcolor{heat!16}0.34 & \cellcolor{heat!8}0.28 & \cellcolor{heat!39}0.57 & \cellcolor{heat!47}0.66 & \cellcolor{heat!31}0.50 & \cellcolor{heat!16}0.37 & \cellcolor{heat!54}0.71 & \cellcolor{heat!39}0.59 & \cellcolor{heat!39}0.55 & \cellcolor{heat!39}0.58 & \cellcolor{heat!31}0.50 \\
ATN@5$_{\mathrm{last}}$ & \cellcolor{heat!39}0.58 & \cellcolor{heat!16}0.35 & \cellcolor{heat!39}0.54 & \cellcolor{heat!39}0.57 & \cellcolor{heat!31}0.52 & \cellcolor{heat!39}0.55 & \cellcolor{heat!31}0.49 & \cellcolor{heat!31}0.49 & \cellcolor{heat!31}0.49 & \cellcolor{heat!31}0.51 & \cellcolor{heat!47}0.63 & \cellcolor{heat!31}0.48 & \cellcolor{heat!23}0.40 & \cellcolor{heat!23}0.42 & \cellcolor{heat!31}0.49 & \cellcolor{heat!47}0.67 & \cellcolor{heat!23}0.42 & \cellcolor{heat!31}0.51 & \cellcolor{heat!23}0.43 & \cellcolor{heat!31}0.51 \\
\cmidrule[0.01em](lr){1-21}
Prop-Dist & \cellcolor{heat!47}0.69 & \cellcolor{heat!23}0.45 & \cellcolor{heat!47}0.65 & \cellcolor{heat!39}0.59 & \cellcolor{heat!47}0.65 & \cellcolor{heat!23}0.42 & \cellcolor{heat!31}0.51 & \cellcolor{heat!8}0.23 & \cellcolor{heat!16}0.35 & \cellcolor{heat!8}0.25 & \cellcolor{heat!8}0.29 & \cellcolor{heat!23}0.41 & \cellcolor{heat!31}0.51 & \cellcolor{heat!8}0.27 & \cellcolor{heat!23}0.38 & \cellcolor{heat!54}0.69 & \cellcolor{heat!39}0.55 & \cellcolor{heat!31}0.47 & \cellcolor{heat!39}0.57 & \cellcolor{heat!31}0.48 \\
Prop-HMM & \cellcolor{heat!47}0.67 & \cellcolor{heat!23}0.41 & \cellcolor{heat!47}0.65 & \cellcolor{heat!39}0.58 & \cellcolor{heat!47}0.62 & \cellcolor{heat!23}0.43 & \cellcolor{heat!39}0.54 & \cellcolor{heat!16}0.34 & \cellcolor{heat!23}0.39 & \cellcolor{heat!8}0.25 & \cellcolor{heat!8}0.27 & \cellcolor{heat!23}0.44 & \cellcolor{heat!23}0.43 & 0.19 & \cellcolor{heat!23}0.38 & \cellcolor{heat!54}0.71 & \cellcolor{heat!39}0.54 & \cellcolor{heat!31}0.50 & \cellcolor{heat!39}0.56 & \cellcolor{heat!23}0.44 \\
\cmidrule[0.01em](lr){1-21}
P(True) & \cellcolor{heat!62}\textbf{0.77} & \cellcolor{heat!54}0.75 & \cellcolor{heat!62}\textbf{0.82} & \cellcolor{heat!54}\textbf{0.75} & \cellcolor{heat!70}0.85 & \cellcolor{heat!23}0.40 & \cellcolor{heat!39}0.55 & \cellcolor{heat!8}0.23 & \cellcolor{heat!16}0.36 & \cellcolor{heat!47}0.66 & \cellcolor{heat!70}\textbf{0.85} & \cellcolor{heat!47}0.61 & \cellcolor{heat!47}0.61 & \cellcolor{heat!47}0.66 & \cellcolor{heat!54}0.73 & \cellcolor{heat!54}0.71 & \cellcolor{heat!39}0.59 & \cellcolor{heat!47}0.67 & \cellcolor{heat!47}0.62 & \cellcolor{heat!47}0.66 \\
VC & \cellcolor{heat!54}0.75 & \cellcolor{heat!62}\textbf{0.83} & \cellcolor{heat!54}0.69 & \cellcolor{heat!39}0.59 & \cellcolor{heat!70}\textbf{0.88} & \cellcolor{heat!39}0.55 & \cellcolor{heat!54}0.70 & \cellcolor{heat!23}0.40 & \cellcolor{heat!23}0.43 & \cellcolor{heat!39}0.54 & \cellcolor{heat!47}0.67 & \cellcolor{heat!47}0.67 & \cellcolor{heat!54}0.72 & \cellcolor{heat!47}0.65 & \cellcolor{heat!54}0.73 & \cellcolor{heat!47}0.64 & \cellcolor{heat!54}0.71 & \cellcolor{heat!39}0.53 & \cellcolor{heat!39}0.57 & \cellcolor{heat!47}0.66 \\
\cmidrule[0.01em](lr){1-21}
NCP & \cellcolor{heat!47}0.66 & \cellcolor{heat!39}0.54 & \cellcolor{heat!47}0.61 & \cellcolor{heat!39}0.59 & \cellcolor{heat!47}0.68 & \cellcolor{heat!31}0.51 & \cellcolor{heat!54}0.71 & \cellcolor{heat!31}0.49 & \cellcolor{heat!62}0.83 & \cellcolor{heat!47}0.62 & \cellcolor{heat!16}0.31 & \cellcolor{heat!16}0.36 & \cellcolor{heat!31}0.46 & \cellcolor{heat!62}0.80 & \cellcolor{heat!31}0.50 & \cellcolor{heat!54}\textbf{0.76} & \cellcolor{heat!47}0.68 & \cellcolor{heat!39}0.57 & \cellcolor{heat!47}0.67 & \cellcolor{heat!54}0.71 \\
FAC & \cellcolor{heat!39}0.56 & \cellcolor{heat!54}0.71 & \cellcolor{heat!31}0.53 & \cellcolor{heat!39}0.58 & \cellcolor{heat!39}0.58 & \cellcolor{heat!39}0.54 & \cellcolor{heat!47}0.62 & \cellcolor{heat!31}0.50 & \cellcolor{heat!47}0.62 & \cellcolor{heat!39}0.58 & \cellcolor{heat!31}0.45 & \cellcolor{heat!39}0.56 & \cellcolor{heat!47}0.65 & \cellcolor{heat!31}0.50 & \cellcolor{heat!31}0.51 & \cellcolor{heat!31}0.51 & \cellcolor{heat!39}0.56 & \cellcolor{heat!31}0.50 & \cellcolor{heat!31}0.49 & \cellcolor{heat!31}0.50 \\
ASC & \cellcolor{heat!54}0.69 & \cellcolor{heat!47}0.69 & \cellcolor{heat!54}0.75 & \cellcolor{heat!47}0.68 & \cellcolor{heat!62}0.77 & \cellcolor{heat!31}0.50 & \cellcolor{heat!62}\textbf{0.82} & \cellcolor{heat!8}0.22 & \cellcolor{heat!39}0.55 & \cellcolor{heat!39}0.61 & \cellcolor{heat!23}0.38 & \cellcolor{heat!54}\textbf{0.76} & \cellcolor{heat!54}\textbf{0.72} & \cellcolor{heat!54}0.76 & \cellcolor{heat!70}\textbf{0.85} & \cellcolor{heat!47}0.64 & \cellcolor{heat!54}0.71 & \cellcolor{heat!54}\textbf{0.73} & \cellcolor{heat!47}0.68 & \cellcolor{heat!54}0.74 \\
ADC & \cellcolor{heat!54}0.71 & \cellcolor{heat!16}0.31 & \cellcolor{heat!54}0.75 & \cellcolor{heat!54}0.70 & \cellcolor{heat!54}0.76 & \cellcolor{heat!39}0.55 & \cellcolor{heat!62}0.80 & \cellcolor{heat!16}0.34 & \cellcolor{heat!39}0.58 & \cellcolor{heat!47}0.67 & \cellcolor{heat!31}0.53 & \cellcolor{heat!54}0.69 & \cellcolor{heat!39}0.58 & \cellcolor{heat!47}0.66 & \cellcolor{heat!62}0.81 & \cellcolor{heat!47}0.64 & \cellcolor{heat!54}\textbf{0.72} & \cellcolor{heat!47}0.68 & \cellcolor{heat!54}0.70 & \cellcolor{heat!54}0.74 \\
AEC & \cellcolor{heat!54}0.71 & \cellcolor{heat!8}0.29 & \cellcolor{heat!54}0.75 & \cellcolor{heat!54}0.71 & \cellcolor{heat!54}0.76 & \cellcolor{heat!39}0.56 & \cellcolor{heat!62}0.77 & \cellcolor{heat!23}0.42 & \cellcolor{heat!47}0.63 & \cellcolor{heat!54}\textbf{0.74} & \cellcolor{heat!31}0.52 & \cellcolor{heat!54}0.71 & \cellcolor{heat!31}0.50 & \cellcolor{heat!23}0.43 & \cellcolor{heat!47}0.61 & \cellcolor{heat!47}0.68 & \cellcolor{heat!54}0.71 & \cellcolor{heat!47}0.65 & \cellcolor{heat!54}0.69 & \cellcolor{heat!54}0.71 \\
TER & \cellcolor{heat!54}0.71 & \cellcolor{heat!39}0.58 & \cellcolor{heat!54}0.74 & \cellcolor{heat!47}0.67 & \cellcolor{heat!62}0.77 & \cellcolor{heat!47}\textbf{0.65} & \cellcolor{heat!47}0.66 & \cellcolor{heat!39}0.58 & \cellcolor{heat!70}\textbf{0.87} & \cellcolor{heat!47}0.67 & \cellcolor{heat!16}0.34 & \cellcolor{heat!47}0.62 & \cellcolor{heat!54}0.69 & \cellcolor{heat!62}\textbf{0.81} & \cellcolor{heat!54}0.76 & \cellcolor{heat!54}0.76 & \cellcolor{heat!47}0.68 & \cellcolor{heat!47}0.65 & \cellcolor{heat!62}\textbf{0.77} & \cellcolor{heat!54}\textbf{0.75} \\
\midrule[\heavyrulewidth]
Success rate & 54\% & 5\% & 26\% & 65\% & 49\% & 24\% & 52\% & 20\% & 74\% & 74\% & 11\% & 25\% & 11\% & 80\% & 61\% & 37\% & 45\% & 16\% & 68\% & 55\% \\
Minority $n$ & 92 & 10$^\dagger$ & 53 & 70 & 98 & 12$^\dagger$ & 24 & 10$^\dagger$ & 13$^\dagger$ & 13$^\dagger$ & 12$^\dagger$ & 28 & 13$^\dagger$ & 23 & 45 & 42 & 51 & 18$^\dagger$ & 36 & 51 \\
\bottomrule
\end{tabular}
}
\caption{AUROC point estimates for predicting greedy-trajectory success. Bold marks the highest point estimate per model-domain column; full task-clustered bootstrap intervals are in Table~\ref{tab:auroc-main-ci}. \textsuperscript{$\dagger$}Minority class $n<20$; point estimates and bold marks in these columns should not be read as resolved rankings.}
\label{tab:auroc-main}
\end{table*}

\begin{figure*}[t]
\centering
\includegraphics[width=\textwidth]{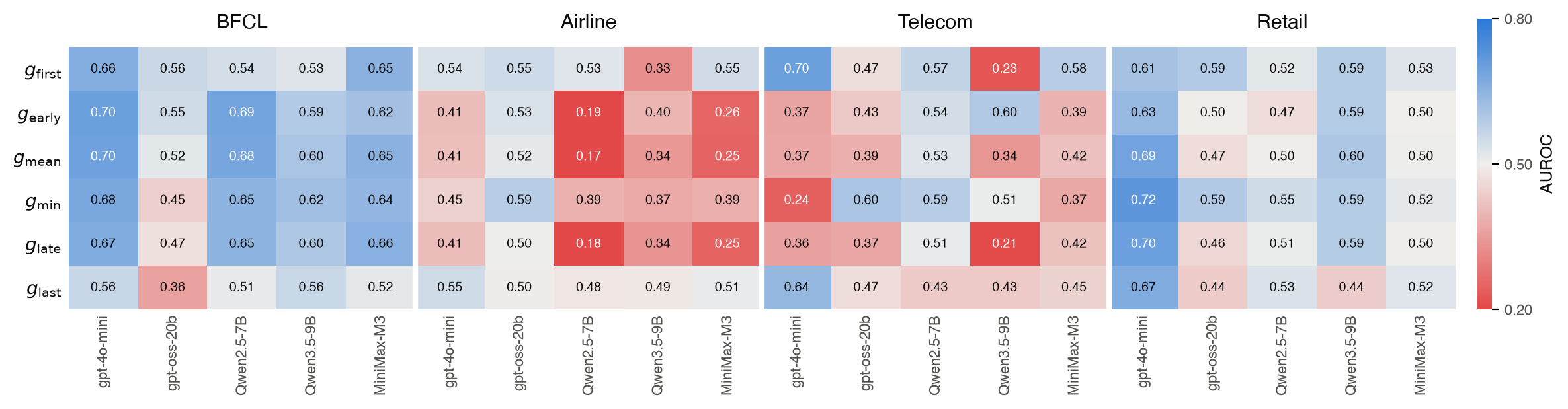}
\caption{White-box AUROC for LNSP under the six cross-turn aggregations
(rows) across the BFCL-v4 and $\tau^2$ model-dataset cells (columns), on the greedy
reference trajectories. Colour is diverging about chance (blue predictive, red
anti-correlated, grey no skill). Aggregation can swing AUROC sharply within a
column. ATN@5 and probability margin broadly track LNSP, whereas SP can diverge
where sequence-length effects matter (Table~\ref{tab:auroc-full-greedy}). No row
is reliably predictive, and the highest-AUROC aggregation is
dataset- and model-specific: the min rule is strong on retail for \texttt{gpt-4o-mini}
(0.72) but anti-correlated on telecom (0.24), and the mean-style rules
collapse across airline.}
\label{fig:whitebox-agg}
\end{figure*}

\paragraph{White-Box Scorers.} Table~\ref{tab:auroc-main} reports SP, LNSP,
and the entropy-based ATN@5 under the first, mean, min, and last aggregators. 
White-box performance varies
with both base score and aggregator. BFCL-v4 is the most favorable setting, with
mean or minimum entropy reaching 0.65--0.71 AUROC for \texttt{gpt-4o-mini}, \texttt{Qwen2.5-7B},
and \texttt{MiniMax-M3}. Among the four aggregation rules in Table~\ref{tab:auroc-main},
the largest $\tau^2$-bench white-box AUROC point estimate is 0.725 (SP$_{\mathrm{mean}}$,
\texttt{Qwen3.5-9B} on telecom), but switching the aggregation for
that same base score moves AUROC to 0.228. On airline, the mean aggregator applied to SP and LNSP, the
scorers used in \citet{oh2026uncertaintyquantificationllmagents}, 
is approximately at or below chance for every model. \texttt{MiniMax-M3}'s highest AUROC is 0.62 across white-box scorers in Table~\ref{tab:auroc-main}
on $\tau^2$-bench and 0.69 on BFCL-v4.

Figure~\ref{fig:whitebox-agg} shows this instability across the whole grid. For
LNSP, changing the aggregation rule produces large AUROC swings within
model-dataset columns, and no aggregation is predictive everywhere. Across
the white-box grid, 91 of 240 cells fall below 0.5, and 21 have bootstrap
intervals lying entirely below 0.5
(Table~\ref{tab:auroc-main-ci}). Selective prediction shows the same sign
instability (Table~\ref{tab:prr}), with LNSP$_{\mathrm{min}}$ attaining
0.44 PRR on retail but $-0.28$ on telecom for \texttt{gpt-4o-mini}, where abstaining on
its least-confident trajectories retains a worse success rate than
abstaining at random. The SAUP-inspired propagation controls
(Appendix~\ref{app:saup}) also underperform across the three $\tau^2$-bench
datasets. The strongest control averages 0.434 AUROC across the fifteen
$\tau^2$-bench model-dataset cells, below the strongest fixed naive aggregation
(0.53, SP$_{\mathrm{last}}$).

\paragraph{Black-Box Consistency Scorers.} The consistency scorers fall into 
three design patterns: 1) NCP compares final messages with an
off-the-shelf NLI model, 2) the action-structured scorers (FAC, ASC, ADC, AEC)
compare action types symbolically, and 3) TER replaces the symbolic comparison
with a \texttt{gemini-flash-lite} judge of whole-trajectory outcome
equivalence. NCP ranges from 0.832 for \texttt{Qwen3.5-9B} on airline to at or below
chance on telecom for several models. On the $\tau^2$-bench datasets, FAC typically stays near chance
because the first action is usually a message to the user. On BFCL-v4, action- or trajectory-level
agreement reaches about 0.71--0.77 AUROC for four models. \texttt{gpt-oss-20b} is more
mixed, with FAC at 0.714 and ASC at 0.685 but the order- and
repetition-sensitive scores below chance (ADC 0.31, AEC 0.29).
On airline, consistency has the largest family-level point estimate for four of five models, with TER reaching 0.868 for \texttt{Qwen3.5-9B} and ASC reaching 0.819 for
\texttt{gpt-oss-20b}. Across the black-box consistency variants, TER and ASC most often
occupy the top ranks, as TER is strongest in several airline and retail cells,
while ASC is competitive or best in multiple BFCL-v4, airline, and telecom cells. Despite strong performance elsewhere,
TER falls to 0.342 on \texttt{gpt-4o-mini} telecom, where the 11\% success
rate leaves the judge comparing mostly failure-failure pairs. Overall, no single
consistency scorer dominates across datasets.

\paragraph{Reflexive Scorers.} Reflexive scoring has its strongest AUROC point
estimates on BFCL-v4, ranging from
0.746--0.852 across models for P(True) and reaching at least 0.75 on three models for VC. The \texttt{gpt-oss-20b} column is one of
the small-minority-class columns flagged in Table~\ref{tab:auroc-main}, since
that model succeeds on only 5\% of BFCL-v4 tasks, but its VC interval
[0.749, 0.906] still separates from chance. Across the $\tau^2$-bench datasets, the best reflexive score
is 0.710 on retail, 0.852 on telecom, and 0.703 on airline. Airline shows the
widest cross-model spread, with P(True) ranging from
0.225 to 0.659 and VC from 0.400 to 0.703 (notably lower than the best black-box consistency AUROC of 0.87). 
Calibration roughly follows this pattern, with 
reflexive ECE (Table~\ref{tab:ece}) being low for \texttt{gpt-4o-mini} on telecom
(0.050 for VC and 0.104 for P(True)) but much worse for several airline and
telecom model-scorer cells, reaching 0.560. Selective prediction is similarly
mixed, with airline PRR ranging from negative for several P(True) columns to 0.381
for \texttt{gpt-oss-20b} VC and 0.303 for \texttt{MiniMax-M3} P(True). Which models retain
airline reflexive signal does not appear to track task success. For instance, \texttt{Qwen3.5-9B} and
\texttt{MiniMax-M3} both succeed on 74\% of airline tasks, yet \texttt{Qwen3.5-9B} has weak
reflexive AUROC (0.356 for P(True) and 0.430 for VC) while \texttt{MiniMax-M3} retains a stronger
P(True) signal (0.659).\footnote{Retail labels are partly based on benchmark-provided \texttt{gpt-4.1} judgments of
whether the final response satisfies required natural-language assertions
(Appendix~\ref{app:grading}). Reflexive results on retail should therefore be
read with this judge-dependence caveat in mind.}


\section{Discussion and Conclusions}
\label{sec:discussion}
In our experiments across four datasets and five LLMs, our results indicate that 
trajectory-adapted variants of single-turn UQ have mixed performance. In particular, no single
family or scorer is reliable across benchmarks and models,  and performance depends
on where uncertainty is read, how per-turn signals are aggregated, the domain, and the model being scored.

\paragraph{Transferability by UQ family.} The three families fail in different
ways, which suggests that agent UQ is not just single-turn UQ with a longer
context. Token-probability scores are cheap and theoretically grounded, but in
our trajectories they are dominated by the choice of where and how to aggregate, as
early, mean, minimum, and final-turn aggregators can imply different rankings for
the same model and domain. This makes raw white-box confidence a poor
drop-in method unless the aggregation rule is validated for the deployment
setting. Reflexive scores are more robust in aggregate and require only one
extra pass over the transcript, making them a strong practical baseline. Among sampled-trajectory
consistency scorers, final-message agreement is weakest on average but remains
competitive in some model-dataset cells. Among trajectory-structured measures,
trajectory equivalence rate and action-set agreement are the two that most often lead.
Black-box self-consistency,
however, incurs the cost of additional sampled trajectories.

\paragraph{Robustness checks.} We use two targeted ablations to assess whether
the main findings depend on the action interface or on resampling the simulated
user. First, the results do not appear to be
an artifact of the text-action interface. On $\tau^2$-retail with \texttt{gpt-4o-mini},
native tool calling gives similar greedy success and nearly identical TER and
AEC AUROC (Appendix~\ref{app:native-tool-calling}). Second, full-trajectory sampling
appears to provide a stronger signal than turn-wise prefix conditioning. We conduct a
prefix-conditioned ablation (Appendix~\ref{app:prefix-action-consistency}) on $\tau^2$-airline, where 
we hold the transcript fixed and resample
only \texttt{gpt-4o-mini}'s next action, finding only 
modest ranking ability (up to 0.588 AUROC). The gap is consistent with
downstream trajectory dynamics contributing additional signal beyond local
next-action stability.

\begin{table}[t]
\centering
\small
\setlength{\tabcolsep}{3pt}
\begin{tabular}{lccccc}
\toprule
Family & Extra & Aux. & Aux. & \multicolumn{2}{c}{AUROC} \\
\cmidrule(lr){5-6}
 & traj. & model & calls & mean & peak \\
\midrule
White-box & 0 & -- & 0 & 0.628 & 0.725 \\
Reflexive & 0 & agent & 1 & 0.691 & 0.885 \\
Action consistency & $m$ & -- & 0 & 0.705 & 0.849 \\
Message consistency & $m$ & NLI & $2m$ & 0.603 & 0.832 \\
Judged consistency & $m$ & \texttt{Gemini} & $m$ & 0.686 & 0.868 \\
\bottomrule
\end{tabular}
\caption{Compute budget and discrimination of each scorer family, per greedy-referenced task, with $m{=}3$ resampled trajectories. Budget is given in provider-independent units: extra agent trajectories and auxiliary model calls per task. The $m$ draws dominate the cost and are shared across all consistency scorers, so the judging layered on them adds only a constant factor; white-box logprobs add no extra model calls when the interface exposes action-token probabilities, and reflexive scoring adds one self-evaluation pass over the last-action transcript prefix. Auxiliary calls exclude the agent-trajectory cost, which dominates and is model- and domain-dependent. The AUROC columns give the mean over available model-benchmark cells of each family's highest point estimate per cell, and its single highest cell (peak), from Table~\ref{tab:auroc-main}.}
\label{tab:cost}
\end{table}

\paragraph{Cost and guidance.} White-box scores add no extra model calls when
action-token log probabilities are available, reflexive scoring adds one
self-evaluation pass, and consistency scorers incur the cost of the
$m$ resampled trajectories, with TER's \texttt{gemini-flash-lite} judge adding a negligible
\$0.002 per task. Table~\ref{tab:cost} summarizes these tradeoffs. The main
cost distinction is not the cheap equivalence judge but the need to run additional
agent trajectories. Once the $m$ sampled trajectories are cached, action-set and 
action-distribution consistency require no additional model calls, NCP’s $2m$
directional NLI evaluations are batched locally, and TER adds one auxiliary judge call per sample. 
The practical choice therefore depends on when the signal is needed. If
token log probabilities over actions are available, white-box scores are the
only zero-extra-call option, but our results suggest they require per-domain and per-model
validation. Reflexive scoring is the lowest-cost, broadly useful baseline when
a pre-outcome transcript assessment is acceptable. Sampled-trajectory
consistency is most attractive when repeated runs are already part of the
evaluation budget, when pass@k-style sampling is required for another reason,
or when a high-stakes deployment can afford extra trajectories for a stronger
failure-ranking signal.

 \paragraph{Future work.} The results point to agent-adapted uncertainty as a
productive direction.
We recommend experiments with step-level labels that enable evaluation by action type and tool family. 
We also recommend testing whether acting on per-turn scores improves outcomes during execution.
Finally, exploring different types of agentic environments exposed by other benchmarks (e.g., web-browsing) 
would provide insights on generalizability of our findings beyond the settings we explored.


\section*{Limitations}
\paragraph{Benchmark scope and label noise.} Our evaluation covers the multi-turn subset of BFCL-v4 and 
three $\tau^2$-bench text domains, so it is a focused study of
tool-using benchmarks rather than a claim of breadth across all agentic
environments. How far the conditional picture we report extends to other
tool-use benchmarks, to web or embodied agents, or to more open-ended tasks is
untested. The prediction target is the benchmark's trajectory reward, computed
from database-state, assertion, action, and final-message checks depending on
the dataset, and this label is imperfect. Reward noise upper-bounds the
discrimination any scorer can achieve and can distort calibration.

\paragraph{User-simulator confound.} Trajectories are produced against a
simulated user, and a fraction of failures may originate in the simulator rather
than the agent. We hold the simulator fixed across all models and run it at
temperature 0, which minimizes simulator-side sampling variance and prevents
cross-model comparisons from being confounded by different user models. Even
so, simulator-induced failures can still add label noise that affects the
measured reliability of every method, and a different simulator could change
absolute success rates and with them the difficulty of the prediction target.
The simulated user is also resampled
together with the agent when we draw the consistency trajectories, so the
black-box consistency scorers measure the variability of the joint agent-user
system rather than the agent's uncertainty in isolation. We include a targeted
agent-only prefix-conditioned ablation on airline
(Appendix~\ref{app:prefix-action-consistency}) that diagnoses the concern by
holding the transcript fixed and resampling only the next agent action.

\paragraph{Action interface scope.} The agent emits each action as text rather
than through native function-calling, which is what makes action-token log
probabilities observable for every model
(Appendix~\ref{app:native-tool-calling}). Our
white-box results are therefore specific to that action surface. In a native tool-calling ablation on
retail with \texttt{gpt-4o-mini}, the greedy success-rate difference between interfaces
is not significant at the 5\% level and the strongest consistency scores are
stable across interfaces. That ablation covers one model and one
domain, and native tool-call support was not uniformly usable across the model
ladder, so we treat the interface as a deployment variable rather than a clean
cross-model control.

\paragraph{Statistical power.} Per-domain task counts are
modest, so the bootstrap intervals in Appendix~\ref{app:statistical-uncertainty}
are wide, especially on airline (50 tasks). Small differences between scorers
should therefore be read as suggestive rather than established, as many per-column
best-scorer margins are plausibly within sampling noise, and we do not claim a
scorer is significantly best where the intervals do not separate it. 

\paragraph{Judge dependence.} The judged trajectory-equivalence scorer relies on
a single external judge model. On the
database-graded domains we validate the equivalence judge against a database-state ground
truth (Appendix~\ref{app:ter-prompt}), where it attains 0.88 precision and 0.93
recall over $1{,}365$ pairs.
 An ablation that replaces
the judge with the larger \texttt{gemini-flash}  finds the two judges agreeing on 91\% of pairs and moves TER's
AUROC by at most 0.062, so the airline TER conclusions are robust to this
particular judge swap.
We do not, however, test every domain or more distant judge models, and the
scorer's reliability still inherits the judge's competence.

\paragraph{Scope of the uncertainty target.} We study the action term of the
trajectory and score the action span of each turn's generation. Observation
uncertainty, white-box scores over the reasoning tokens rather than the
committed action, and step-level labels that would allow evaluation by action
type and tool family are all outside the present scope. We also evaluate
whether scores predict completed-trajectory success, not whether acting on
them through abstention, clarification, or escalation improves outcomes during
execution.

\section*{Ethical Considerations}
This work studies uncertainty quantification for tool-using agents with the aim
of making such systems safer to deploy: a reliable uncertainty signal lets an
agent abstain or escalate to a human before taking a harmful or irreversible
action. Our measurements do not reach that use directly. Every score we
evaluate is computed over a completed trajectory, and the consistency scorers
additionally require resampled trajectories that have run to completion, so
what we establish is how well these signals classify trajectory success after
the fact, not whether acting on them mid-run would prevent a harmful action.
Our findings carry a corresponding caution. Because several of the
signals we evaluate transfer unreliably, and some anti-correlate with success,
treating any of them as a trustworthy safety gate without validation in the
target domain and model could produce false assurance and permit confident but
incorrect state-changing actions. We therefore report where each method fails 
as prominently as where it
works, and recommend per-deployment validation rather than transfer by
default.


\bibliography{refs.bib}

\appendix


\appendix

\section{Additional White-Box Scorers and Propagation Baseline}
\label{app:whitebox-scorers}

The main text evaluates SP, LNSP, and ATN@5 under the first, mean, min, and
last cross-turn aggregations. The full grid in
Table~\ref{tab:auroc-full-greedy} additionally covers the probability margin
score and the position-weighted early and late aggregations, defined here.
Notation follows Section~\ref{sec:methods}: $y$ is a generation
with tokens $\{t_1, \ldots, t_L\}$, per-token probabilities $p_j$, top-$K$
probabilities $\{p_{j,1}, \ldots, p_{j,K}\}$ at position $j$.

\paragraph{Probability Margin (PM).} The mean separation between the two most probable tokens at each position \citep{farr2024redctsystemsdesignmethodology}:
$$\text{PM}(y) = \frac{1}{L} \sum_{j=1}^{L} (p_{j,1} - p_{j,2})$$

\paragraph{Position-weighted aggregation.} With per-turn confidences $s_i$ as
in Section~\ref{subsec:aggregation}, the position-weighted aggregator is
$$
g_{\text{pos}} = \frac{\sum_{i=1}^{T} w_i\, s_i}{\sum_{i=1}^{T} w_i},
$$
with either early weights $w_i = T - i + 1$ or late weights $w_i = i$, a
general early- or late-weighted convex combination of the per-turn scores.

\subsection{SAUP-Inspired Propagation Controls}
\label{app:saup}

SAUP \citep{zhao-etal-2025-uncertainty} motivates step-level uncertainty
propagation: rather than reducing per-turn uncertainties by a fixed rule as in
Section~\ref{subsec:aggregation}, it weights each step's uncertainty by a
situation-dependent surrogate and aggregates by root mean square. We include
SAUP-inspired propagation controls adapted to our recorded action traces, not a
faithful reproduction of SAUP. The exact SAUP setting assumes ReAct trajectories
with separate thought, action, and observation spans, and its learned HMM
variant relies on situational-state supervision. Our traces expose the
committed action span but not a clean thought span, and we do not have
step-level situational labels.

The per-step uncertainty $U_n$ is therefore the length-normalized negative
log-likelihood of the action span, the same quantity that underlies our
white-box scores. For distance-based controls, we compute an inquiry drift
$D_a^n = 1 - \cos(Z_n, Q)$ between the running trajectory state $Z_n$ and the
task $Q$, and an inference gap $D_o^n = 1 - \cos(A_n, O_n)$ between the action
$A_n$ and its observation $O_n$. States, actions, observations, and the task are
embedded with a local sentence encoder (\texttt{all-MiniLM-L6-v2}); $Z_n$ is
the mean of its message embeddings. Each trajectory score is
$U_{\text{agent}} = \sqrt{\frac{1}{N}\sum_n (W_n U_n)^2}$, oriented as a
confidence by negation. Prop-Pos uses a position surrogate with larger weights
near the final step; Prop-Dist uses normalized $D_a^n + D_o^n$ weights;
Prop-PD multiplies the position and distance weights; Prop-HMM fits an
unsupervised three-state Gaussian HMM to the two-dimensional $(D_a,D_o)$
sequences, orders states by mean total drift, and uses the posterior-expected
state severity as $W_n$. Prop-Uniform sets $W_n=1$, giving the unweighted RMS
of per-step uncertainty. All nonuniform weights are normalized within a trajectory so that
trajectory length does not dominate the score scale; the HMM is fit with
five-fold cross-fitting across tasks.

Averaged over the fifteen $\tau^2$ model-domain cells the AUROCs are 0.434 (Prop-PD),
0.433 (Prop-Pos), 0.427 (Prop-HMM), and 0.426 (Prop-Dist and Prop-Uniform),
against a best fixed naive aggregation of 0.53 and a per-cell oracle of 0.62.
The HMM weighting beats the unweighted RMS in 7 of 15 cells and none of the
propagation controls beats the per-cell naive oracle. We therefore interpret
these rows only as propagation-style controls adapted to our action-level
traces, and not as evidence for or against the original SAUP method. 

\section{Trajectory Success Grading}
\label{app:grading}

\paragraph{Dataset licenses.}
The BFCL-v4 data are distributed through the Berkeley Function
Calling Leaderboard/Gorilla release, whose dataset card and repository list an
Apache-2.0 license. The $\tau^2$-bench domains used here--airline, telecom, and
retail--are distributed with the $\tau^2$-bench repository under the MIT license.

\paragraph{BFCL-v4 grading.}
For BFCL-v4, we use the benchmark's official binary pass/fail
grading for each completed trajectory. A trajectory is successful when the
model's tool-call behavior and final response satisfy the task-specific
expected function-call constraints in the BFCL-v4 evaluator; otherwise it is
marked as failed. We use this binary pass/fail label as the trajectory-success
target for all BFCL-v4 UQ metrics, matching the use of $\tau^2$ reward labels for the
three $\tau^2$ domains.

\paragraph{$\tau^2$ grading.}
$\tau^2$ grades a trajectory with a product of binary component checks, with each
task specifying which components enter its reward. A trajectory that ends
without a proper agent or user stop, for example at a step limit, is graded 0
without evaluation. The components are as follows. \textbf{Database state}:
$\tau^2$ replays the trajectory's tool calls in a fresh copy of the environment
and compares a hash of the resulting database against the state produced by
replaying the task's gold actions, passing only on an exact match.
\textbf{Environment assertions}: programmatic predicates evaluated on the
trajectory's final environment state, all of which must pass. \textbf{Required
actions}: each gold tool call must appear among the trajectory's tool calls,
compared by name and arguments, in any order. \textbf{Communicated
information}: each required information string must appear as a
case-insensitive substring of some agent message; the check scans every agent
message, not only the final one. \textbf{Natural-language assertions}: an LLM
judge (\texttt{gpt-4.1}, temperature 0) reads the transcript and decides
whether each behavioral assertion about the agent is met, all of which must
be.

The composition over our task sets is domain-specific. Every retail task is
graded on database state, and 112 of 114 additionally include the
natural-language assertions in their basis (non-vacuous on 40 tasks). Every
airline task is graded on database state and communicated information
(non-vacuous on 6 of 50). Every telecom task is graded on environment
assertions, with 20 of 114 additionally requiring action checks. The retail
label is therefore partly judge-derived: on the 40 retail tasks with
assertions, an LLM judgment can decide success when the database check
passes. This judge enters only the label, never any scorer, but its model
family overlaps the \texttt{gpt-4o-mini} agent and the \texttt{gpt-4.1-mini} user
simulator, a dependence we inherit from the benchmark.

\section{Full Results}
\label{app:full-results}

\begin{figure*}[t]
\centering
\includegraphics[width=0.95\textwidth]{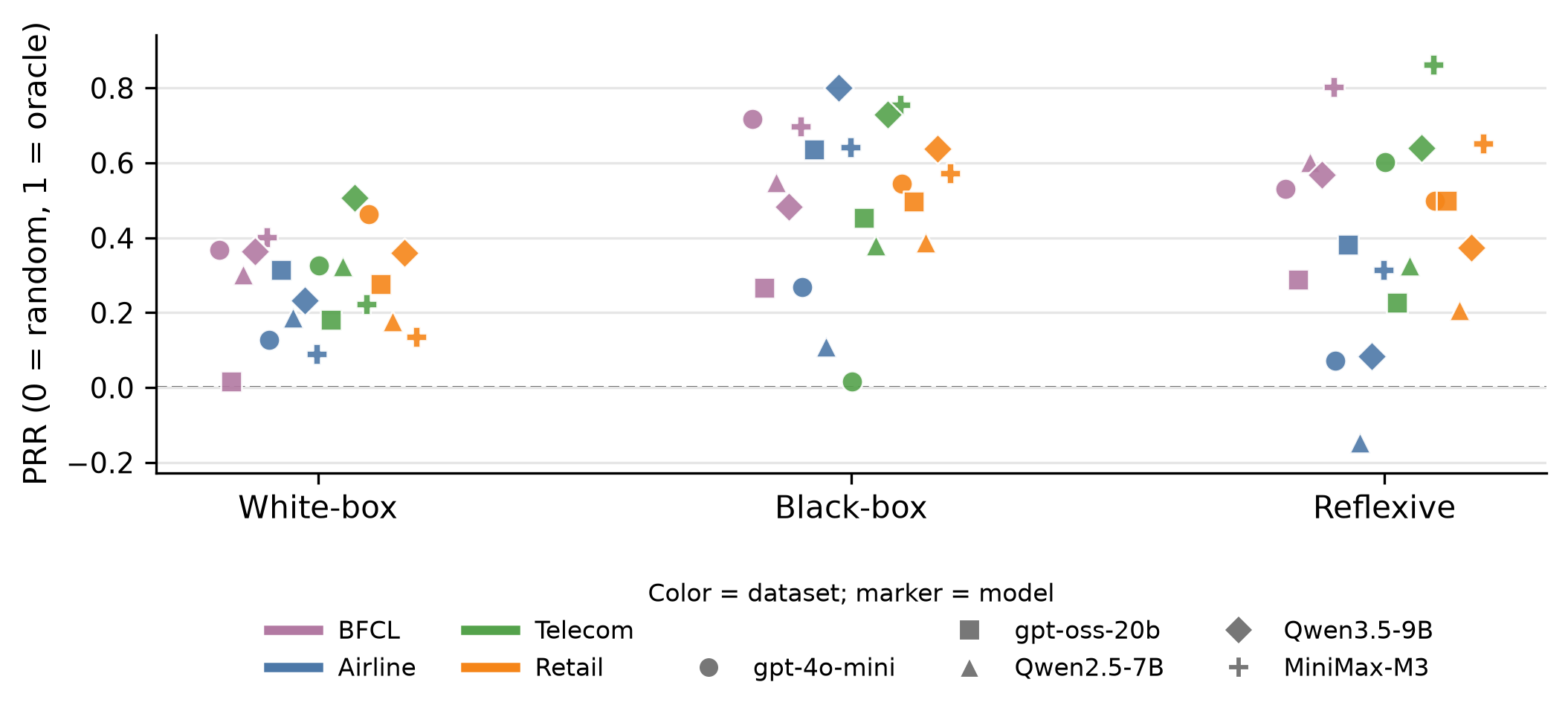}
\caption{Selective-prediction performance by method family. Each point is the
highest-PRR scorer for one model-domain cell, selected descriptively within
family, so families with more candidate scorers get more selection latitude
(24 white-box scorers, excluding the propagation controls; six consistency;
two reflexive).}
\label{fig:method-family-prr}
\end{figure*}

Table~\ref{tab:auroc-main-ci} repeats the main AUROC table with a
task-clustered bootstrap interval for every cell; it is discussed in
Appendix~\ref{app:statistical-uncertainty}.
Table~\ref{tab:auroc-full-greedy} reports AUROC for the full white-box grid on
BFCL-v4 and the three $\tau^2$-bench domains, covering all (base score, cross-turn
aggregation) pairs including PM and the early and late position-weighted
aggregations. It uses the greedy reference trajectories, matching the main
table; the reflexive and consistency scorers appear in the main table and in
Tables~\ref{tab:prr}--\ref{tab:ece}. Table~\ref{tab:auprc}
reports AUPRC (average precision), where the no-skill baseline is the success
rate rather than 0.5. Table~\ref{tab:prr} reports the prediction rejection
ratio and Table~\ref{tab:ece} the expected calibration error, both on the
greedy reference basis matching the main table. ECE covers every scorer whose
values lie in $[0,1]$: the reflexive scorers, the consistency scorers, and the
per-token SP and LNSP confidences (mean aggregation), with the caveat that the
raw SP/LNSP values and consistency agreement rates are not calibrated success
probabilities. Table~\ref{tab:cost}
in the main text reports the compute budget of each scorer family against its
discrimination. Figure~\ref{fig:method-family-prr} summarizes the highest-PRR scorer of each
family per model-domain cell. The white-box aggregation instability these
tables record is plotted over the greedy-basis grid in
Figure~\ref{fig:whitebox-agg} in the main text.

\begin{table*}[t]
\centering
\scriptsize
\setlength{\tabcolsep}{1.5pt}
\resizebox{\textwidth}{!}{%
\begin{tabular}{lcccccccccccccccccccc}
\toprule
 & \multicolumn{5}{c}{BFCL-v4} & \multicolumn{5}{c}{Airline} & \multicolumn{5}{c}{Telecom} & \multicolumn{5}{c}{Retail} \\
\cmidrule(lr){2-6}
\cmidrule(lr){7-11}
\cmidrule(lr){12-16}
\cmidrule(lr){17-21}
Scorer & \rotatebox{90}{\texttt{gpt-4o-mini}} & \rotatebox{90}{\texttt{gpt-oss-20b}} & \rotatebox{90}{\texttt{Qwen2.5-7B}} & \rotatebox{90}{\texttt{Qwen3.5-9B}} & \rotatebox{90}{\texttt{MiniMax-M3}} & \rotatebox{90}{\texttt{gpt-4o-mini}} & \rotatebox{90}{\texttt{gpt-oss-20b}} & \rotatebox{90}{\texttt{Qwen2.5-7B}} & \rotatebox{90}{\texttt{Qwen3.5-9B}} & \rotatebox{90}{\texttt{MiniMax-M3}} & \rotatebox{90}{\texttt{gpt-4o-mini}} & \rotatebox{90}{\texttt{gpt-oss-20b}} & \rotatebox{90}{\texttt{Qwen2.5-7B}} & \rotatebox{90}{\texttt{Qwen3.5-9B}} & \rotatebox{90}{\texttt{MiniMax-M3}} & \rotatebox{90}{\texttt{gpt-4o-mini}} & \rotatebox{90}{\texttt{gpt-oss-20b}} & \rotatebox{90}{\texttt{Qwen2.5-7B}} & \rotatebox{90}{\texttt{Qwen3.5-9B}} & \rotatebox{90}{\texttt{MiniMax-M3}} \\
\midrule
SP$_{\mathrm{first}}$ & \cellcolor{heat!47}.66{\tiny\textpm.08} & \cellcolor{heat!39}.53{\tiny\textpm.19} & \cellcolor{heat!39}.53{\tiny\textpm.08} & \cellcolor{heat!31}.52{\tiny\textpm.09} & \cellcolor{heat!54}.69{\tiny\textpm.07} & \cellcolor{heat!31}.53{\tiny\textpm.17} & \cellcolor{heat!39}.56{\tiny\textpm.17} & \cellcolor{heat!31}.50{\tiny\textpm.25} & \cellcolor{heat!16}.33{\tiny\textpm.19} & \cellcolor{heat!31}.51{\tiny\textpm.20} & \cellcolor{heat!47}.61{\tiny\textpm.14} & \cellcolor{heat!31}.45{\tiny\textpm.12} & \cellcolor{heat!47}.64{\tiny\textpm.19} & \cellcolor{heat!8}.23{\tiny\textpm.10} & \cellcolor{heat!39}.57{\tiny\textpm.11} & \cellcolor{heat!39}.60{\tiny\textpm.11} & \cellcolor{heat!39}.55{\tiny\textpm.11} & \cellcolor{heat!39}.53{\tiny\textpm.14} & \cellcolor{heat!39}.56{\tiny\textpm.11} & \cellcolor{heat!31}.52{\tiny\textpm.11} \\
SP$_{\mathrm{mean}}$ & \cellcolor{heat!47}.62{\tiny\textpm.08} & \cellcolor{heat!31}.52{\tiny\textpm.17} & \cellcolor{heat!47}.65{\tiny\textpm.08} & \cellcolor{heat!39}.59{\tiny\textpm.09} & \cellcolor{heat!47}.66{\tiny\textpm.08} & \cellcolor{heat!31}.47{\tiny\textpm.18} & \cellcolor{heat!31}.52{\tiny\textpm.16} & \cellcolor{heat!16}.30{\tiny\textpm.17} & \cellcolor{heat!31}.47{\tiny\textpm.18} & \cellcolor{heat!16}.33{\tiny\textpm.19} & \cellcolor{heat!23}.38{\tiny\textpm.10} & \cellcolor{heat!16}.36{\tiny\textpm.11} & \cellcolor{heat!31}.52{\tiny\textpm.15} & \cellcolor{heat!54}.73{\tiny\textpm.15} & \cellcolor{heat!23}.43{\tiny\textpm.12} & \cellcolor{heat!47}.64{\tiny\textpm.10} & \cellcolor{heat!16}.32{\tiny\textpm.10} & \cellcolor{heat!31}.48{\tiny\textpm.13} & \cellcolor{heat!39}.56{\tiny\textpm.11} & \cellcolor{heat!39}.56{\tiny\textpm.11} \\
SP$_{\mathrm{min}}$ & \cellcolor{heat!39}.59{\tiny\textpm.08} & \cellcolor{heat!31}.45{\tiny\textpm.20} & \cellcolor{heat!47}.62{\tiny\textpm.09} & \cellcolor{heat!39}.60{\tiny\textpm.09} & \cellcolor{heat!39}.54{\tiny\textpm.08} & \cellcolor{heat!23}.45{\tiny\textpm.18} & \cellcolor{heat!39}.58{\tiny\textpm.17} & \cellcolor{heat!31}.48{\tiny\textpm.19} & \cellcolor{heat!39}.59{\tiny\textpm.19} & \cellcolor{heat!39}.59{\tiny\textpm.20} & \cellcolor{heat!8}.24{\tiny\textpm.10} & \cellcolor{heat!47}.61{\tiny\textpm.12} & \cellcolor{heat!47}.63{\tiny\textpm.15} & \cellcolor{heat!31}.50{\tiny\textpm.11} & \cellcolor{heat!16}.36{\tiny\textpm.10} & \cellcolor{heat!39}.57{\tiny\textpm.11} & \cellcolor{heat!39}.55{\tiny\textpm.11} & \cellcolor{heat!31}.49{\tiny\textpm.15} & \cellcolor{heat!39}.60{\tiny\textpm.11} & \cellcolor{heat!31}.50{\tiny\textpm.11} \\
SP$_{\mathrm{last}}$ & \cellcolor{heat!39}.53{\tiny\textpm.08} & \cellcolor{heat!23}.39{\tiny\textpm.15} & \cellcolor{heat!31}.51{\tiny\textpm.09} & \cellcolor{heat!39}.55{\tiny\textpm.08} & \cellcolor{heat!31}.51{\tiny\textpm.08} & \cellcolor{heat!39}.56{\tiny\textpm.22} & \cellcolor{heat!31}.46{\tiny\textpm.16} & \cellcolor{heat!39}\textbf{.58}{\tiny\textpm.23} & \cellcolor{heat!39}.58{\tiny\textpm.17} & \cellcolor{heat!47}.62{\tiny\textpm.22} & \cellcolor{heat!47}.63{\tiny\textpm.18} & \cellcolor{heat!31}.50{\tiny\textpm.11} & \cellcolor{heat!31}.49{\tiny\textpm.13} & \cellcolor{heat!23}.42{\tiny\textpm.15} & \cellcolor{heat!39}.59{\tiny\textpm.11} & \cellcolor{heat!39}.60{\tiny\textpm.11} & \cellcolor{heat!23}.42{\tiny\textpm.11} & \cellcolor{heat!31}.51{\tiny\textpm.13} & \cellcolor{heat!31}.46{\tiny\textpm.12} & \cellcolor{heat!31}.48{\tiny\textpm.11} \\
LNSP$_{\mathrm{first}}$ & \cellcolor{heat!47}.66{\tiny\textpm.08} & \cellcolor{heat!39}.56{\tiny\textpm.17} & \cellcolor{heat!39}.54{\tiny\textpm.08} & \cellcolor{heat!31}.53{\tiny\textpm.09} & \cellcolor{heat!47}.65{\tiny\textpm.08} & \cellcolor{heat!39}.54{\tiny\textpm.20} & \cellcolor{heat!39}.55{\tiny\textpm.16} & \cellcolor{heat!31}.53{\tiny\textpm.24} & \cellcolor{heat!16}.33{\tiny\textpm.20} & \cellcolor{heat!39}.55{\tiny\textpm.18} & \cellcolor{heat!54}.70{\tiny\textpm.12} & \cellcolor{heat!31}.47{\tiny\textpm.12} & \cellcolor{heat!39}.57{\tiny\textpm.18} & \cellcolor{heat!8}.23{\tiny\textpm.10} & \cellcolor{heat!39}.58{\tiny\textpm.11} & \cellcolor{heat!47}.61{\tiny\textpm.11} & \cellcolor{heat!39}.59{\tiny\textpm.10} & \cellcolor{heat!31}.52{\tiny\textpm.15} & \cellcolor{heat!39}.59{\tiny\textpm.11} & \cellcolor{heat!31}.53{\tiny\textpm.11} \\
LNSP$_{\mathrm{mean}}$ & \cellcolor{heat!54}.70{\tiny\textpm.08} & \cellcolor{heat!31}.52{\tiny\textpm.17} & \cellcolor{heat!47}.68{\tiny\textpm.08} & \cellcolor{heat!39}.60{\tiny\textpm.09} & \cellcolor{heat!47}.65{\tiny\textpm.07} & \cellcolor{heat!23}.41{\tiny\textpm.19} & \cellcolor{heat!31}.52{\tiny\textpm.15} & .17{\tiny\textpm.12} & \cellcolor{heat!16}.34{\tiny\textpm.17} & \cellcolor{heat!8}.25{\tiny\textpm.15} & \cellcolor{heat!16}.37{\tiny\textpm.10} & \cellcolor{heat!23}.39{\tiny\textpm.12} & \cellcolor{heat!31}.53{\tiny\textpm.13} & \cellcolor{heat!16}.34{\tiny\textpm.12} & \cellcolor{heat!23}.42{\tiny\textpm.12} & \cellcolor{heat!54}.69{\tiny\textpm.10} & \cellcolor{heat!31}.47{\tiny\textpm.11} & \cellcolor{heat!31}.50{\tiny\textpm.14} & \cellcolor{heat!39}.60{\tiny\textpm.11} & \cellcolor{heat!31}.50{\tiny\textpm.11} \\
LNSP$_{\mathrm{min}}$ & \cellcolor{heat!47}.68{\tiny\textpm.08} & \cellcolor{heat!23}.45{\tiny\textpm.19} & \cellcolor{heat!47}.65{\tiny\textpm.09} & \cellcolor{heat!47}.62{\tiny\textpm.08} & \cellcolor{heat!47}.64{\tiny\textpm.08} & \cellcolor{heat!23}.45{\tiny\textpm.20} & \cellcolor{heat!39}.59{\tiny\textpm.16} & \cellcolor{heat!23}.39{\tiny\textpm.20} & \cellcolor{heat!23}.37{\tiny\textpm.17} & \cellcolor{heat!23}.39{\tiny\textpm.17} & \cellcolor{heat!8}.24{\tiny\textpm.09} & \cellcolor{heat!39}.60{\tiny\textpm.12} & \cellcolor{heat!39}.59{\tiny\textpm.18} & \cellcolor{heat!31}.51{\tiny\textpm.13} & \cellcolor{heat!23}.37{\tiny\textpm.11} & \cellcolor{heat!54}.72{\tiny\textpm.09} & \cellcolor{heat!39}.59{\tiny\textpm.10} & \cellcolor{heat!39}.55{\tiny\textpm.16} & \cellcolor{heat!39}.59{\tiny\textpm.11} & \cellcolor{heat!31}.52{\tiny\textpm.12} \\
LNSP$_{\mathrm{last}}$ & \cellcolor{heat!39}.56{\tiny\textpm.08} & \cellcolor{heat!16}.36{\tiny\textpm.17} & \cellcolor{heat!31}.51{\tiny\textpm.08} & \cellcolor{heat!39}.56{\tiny\textpm.08} & \cellcolor{heat!31}.52{\tiny\textpm.08} & \cellcolor{heat!39}.55{\tiny\textpm.23} & \cellcolor{heat!31}.50{\tiny\textpm.15} & \cellcolor{heat!31}.48{\tiny\textpm.23} & \cellcolor{heat!31}.49{\tiny\textpm.16} & \cellcolor{heat!31}.51{\tiny\textpm.19} & \cellcolor{heat!47}.64{\tiny\textpm.18} & \cellcolor{heat!31}.47{\tiny\textpm.12} & \cellcolor{heat!23}.43{\tiny\textpm.15} & \cellcolor{heat!23}.43{\tiny\textpm.14} & \cellcolor{heat!23}.45{\tiny\textpm.12} & \cellcolor{heat!47}.67{\tiny\textpm.10} & \cellcolor{heat!23}.44{\tiny\textpm.10} & \cellcolor{heat!39}.53{\tiny\textpm.12} & \cellcolor{heat!23}.44{\tiny\textpm.12} & \cellcolor{heat!31}.52{\tiny\textpm.12} \\
ATN@5$_{\mathrm{first}}$ & \cellcolor{heat!47}.67{\tiny\textpm.08} & \cellcolor{heat!39}.55{\tiny\textpm.17} & \cellcolor{heat!39}.55{\tiny\textpm.08} & \cellcolor{heat!39}.53{\tiny\textpm.09} & \cellcolor{heat!47}.64{\tiny\textpm.08} & \cellcolor{heat!39}.54{\tiny\textpm.19} & \cellcolor{heat!39}.54{\tiny\textpm.17} & \cellcolor{heat!39}.53{\tiny\textpm.23} & \cellcolor{heat!16}.32{\tiny\textpm.20} & \cellcolor{heat!39}.54{\tiny\textpm.19} & \cellcolor{heat!47}.62{\tiny\textpm.14} & \cellcolor{heat!31}.47{\tiny\textpm.11} & \cellcolor{heat!39}.61{\tiny\textpm.19} & .20{\tiny\textpm.09} & \cellcolor{heat!39}.57{\tiny\textpm.11} & \cellcolor{heat!39}.60{\tiny\textpm.11} & \cellcolor{heat!39}.58{\tiny\textpm.10} & \cellcolor{heat!39}.56{\tiny\textpm.14} & \cellcolor{heat!39}.60{\tiny\textpm.11} & \cellcolor{heat!39}.54{\tiny\textpm.11} \\
ATN@5$_{\mathrm{mean}}$ & \cellcolor{heat!54}.71{\tiny\textpm.08} & \cellcolor{heat!31}.50{\tiny\textpm.16} & \cellcolor{heat!54}.70{\tiny\textpm.07} & \cellcolor{heat!47}.62{\tiny\textpm.09} & \cellcolor{heat!47}.66{\tiny\textpm.07} & \cellcolor{heat!23}.41{\tiny\textpm.18} & \cellcolor{heat!31}.50{\tiny\textpm.15} & .18{\tiny\textpm.12} & \cellcolor{heat!16}.36{\tiny\textpm.17} & \cellcolor{heat!8}.24{\tiny\textpm.15} & \cellcolor{heat!23}.38{\tiny\textpm.10} & \cellcolor{heat!23}.38{\tiny\textpm.12} & \cellcolor{heat!39}.54{\tiny\textpm.13} & \cellcolor{heat!16}.35{\tiny\textpm.13} & \cellcolor{heat!23}.43{\tiny\textpm.12} & \cellcolor{heat!47}.68{\tiny\textpm.10} & \cellcolor{heat!23}.44{\tiny\textpm.11} & \cellcolor{heat!31}.49{\tiny\textpm.13} & \cellcolor{heat!39}.60{\tiny\textpm.11} & \cellcolor{heat!31}.49{\tiny\textpm.11} \\
ATN@5$_{\mathrm{min}}$ & \cellcolor{heat!47}.68{\tiny\textpm.08} & \cellcolor{heat!23}.42{\tiny\textpm.19} & \cellcolor{heat!47}.65{\tiny\textpm.09} & \cellcolor{heat!47}.63{\tiny\textpm.08} & \cellcolor{heat!47}.65{\tiny\textpm.07} & \cellcolor{heat!23}.45{\tiny\textpm.20} & \cellcolor{heat!39}.59{\tiny\textpm.16} & \cellcolor{heat!23}.37{\tiny\textpm.19} & \cellcolor{heat!23}.38{\tiny\textpm.18} & \cellcolor{heat!16}.34{\tiny\textpm.16} & \cellcolor{heat!8}.28{\tiny\textpm.10} & \cellcolor{heat!39}.57{\tiny\textpm.12} & \cellcolor{heat!47}.66{\tiny\textpm.17} & \cellcolor{heat!31}.50{\tiny\textpm.15} & \cellcolor{heat!16}.37{\tiny\textpm.11} & \cellcolor{heat!54}.71{\tiny\textpm.10} & \cellcolor{heat!39}.59{\tiny\textpm.11} & \cellcolor{heat!39}.55{\tiny\textpm.16} & \cellcolor{heat!39}.58{\tiny\textpm.11} & \cellcolor{heat!31}.50{\tiny\textpm.11} \\
ATN@5$_{\mathrm{last}}$ & \cellcolor{heat!39}.58{\tiny\textpm.08} & \cellcolor{heat!16}.35{\tiny\textpm.16} & \cellcolor{heat!39}.54{\tiny\textpm.08} & \cellcolor{heat!39}.57{\tiny\textpm.08} & \cellcolor{heat!31}.52{\tiny\textpm.08} & \cellcolor{heat!39}.55{\tiny\textpm.23} & \cellcolor{heat!31}.49{\tiny\textpm.16} & \cellcolor{heat!31}.49{\tiny\textpm.24} & \cellcolor{heat!31}.49{\tiny\textpm.16} & \cellcolor{heat!31}.51{\tiny\textpm.19} & \cellcolor{heat!47}.63{\tiny\textpm.18} & \cellcolor{heat!31}.48{\tiny\textpm.11} & \cellcolor{heat!23}.40{\tiny\textpm.14} & \cellcolor{heat!23}.42{\tiny\textpm.15} & \cellcolor{heat!31}.49{\tiny\textpm.13} & \cellcolor{heat!47}.67{\tiny\textpm.10} & \cellcolor{heat!23}.42{\tiny\textpm.10} & \cellcolor{heat!31}.51{\tiny\textpm.13} & \cellcolor{heat!23}.43{\tiny\textpm.12} & \cellcolor{heat!31}.51{\tiny\textpm.12} \\
\cmidrule[0.01em](lr){1-21}
Prop-Dist & \cellcolor{heat!47}.69{\tiny\textpm.08} & \cellcolor{heat!23}.45{\tiny\textpm.15} & \cellcolor{heat!47}.65{\tiny\textpm.09} & \cellcolor{heat!39}.59{\tiny\textpm.09} & \cellcolor{heat!47}.65{\tiny\textpm.08} & \cellcolor{heat!23}.42{\tiny\textpm.18} & \cellcolor{heat!31}.51{\tiny\textpm.16} & \cellcolor{heat!8}.23{\tiny\textpm.15} & \cellcolor{heat!16}.35{\tiny\textpm.17} & \cellcolor{heat!8}.25{\tiny\textpm.14} & \cellcolor{heat!8}.29{\tiny\textpm.09} & \cellcolor{heat!23}.41{\tiny\textpm.12} & \cellcolor{heat!31}.51{\tiny\textpm.12} & \cellcolor{heat!8}.27{\tiny\textpm.11} & \cellcolor{heat!23}.38{\tiny\textpm.11} & \cellcolor{heat!54}.69{\tiny\textpm.10} & \cellcolor{heat!39}.55{\tiny\textpm.11} & \cellcolor{heat!31}.47{\tiny\textpm.15} & \cellcolor{heat!39}.57{\tiny\textpm.11} & \cellcolor{heat!31}.48{\tiny\textpm.11} \\
Prop-HMM & \cellcolor{heat!47}.67{\tiny\textpm.08} & \cellcolor{heat!23}.41{\tiny\textpm.20} & \cellcolor{heat!47}.65{\tiny\textpm.08} & \cellcolor{heat!39}.58{\tiny\textpm.08} & \cellcolor{heat!47}.62{\tiny\textpm.08} & \cellcolor{heat!23}.43{\tiny\textpm.18} & \cellcolor{heat!39}.54{\tiny\textpm.17} & \cellcolor{heat!16}.34{\tiny\textpm.18} & \cellcolor{heat!23}.39{\tiny\textpm.17} & \cellcolor{heat!8}.25{\tiny\textpm.14} & \cellcolor{heat!8}.27{\tiny\textpm.09} & \cellcolor{heat!23}.44{\tiny\textpm.12} & \cellcolor{heat!23}.43{\tiny\textpm.14} & .19{\tiny\textpm.09} & \cellcolor{heat!23}.38{\tiny\textpm.11} & \cellcolor{heat!54}.71{\tiny\textpm.09} & \cellcolor{heat!39}.54{\tiny\textpm.11} & \cellcolor{heat!31}.50{\tiny\textpm.15} & \cellcolor{heat!39}.56{\tiny\textpm.11} & \cellcolor{heat!23}.44{\tiny\textpm.11} \\
\cmidrule[0.01em](lr){1-21}
P(True) & \cellcolor{heat!62}\textbf{.77}{\tiny\textpm.07} & \cellcolor{heat!54}.75{\tiny\textpm.15} & \cellcolor{heat!62}\textbf{.82}{\tiny\textpm.06} & \cellcolor{heat!54}\textbf{.75}{\tiny\textpm.07} & \cellcolor{heat!70}.85{\tiny\textpm.06} & \cellcolor{heat!23}.40{\tiny\textpm.21} & \cellcolor{heat!39}.55{\tiny\textpm.16} & \cellcolor{heat!8}.23{\tiny\textpm.18} & \cellcolor{heat!16}.36{\tiny\textpm.17} & \cellcolor{heat!47}.66{\tiny\textpm.22} & \cellcolor{heat!70}\textbf{.85}{\tiny\textpm.11} & \cellcolor{heat!47}.61{\tiny\textpm.12} & \cellcolor{heat!47}.61{\tiny\textpm.19} & \cellcolor{heat!47}.66{\tiny\textpm.12} & \cellcolor{heat!54}.73{\tiny\textpm.10} & \cellcolor{heat!54}.71{\tiny\textpm.10} & \cellcolor{heat!39}.59{\tiny\textpm.10} & \cellcolor{heat!47}.67{\tiny\textpm.14} & \cellcolor{heat!47}.62{\tiny\textpm.11} & \cellcolor{heat!47}.66{\tiny\textpm.11} \\
VC & \cellcolor{heat!54}.75{\tiny\textpm.07} & \cellcolor{heat!62}\textbf{.83}{\tiny\textpm.08} & \cellcolor{heat!54}.69{\tiny\textpm.08} & \cellcolor{heat!39}.59{\tiny\textpm.05} & \cellcolor{heat!70}\textbf{.88}{\tiny\textpm.04} & \cellcolor{heat!39}.55{\tiny\textpm.15} & \cellcolor{heat!54}.70{\tiny\textpm.14} & \cellcolor{heat!23}.40{\tiny\textpm.16} & \cellcolor{heat!23}.43{\tiny\textpm.15} & \cellcolor{heat!39}.54{\tiny\textpm.19} & \cellcolor{heat!47}.67{\tiny\textpm.15} & \cellcolor{heat!47}.67{\tiny\textpm.11} & \cellcolor{heat!54}.72{\tiny\textpm.12} & \cellcolor{heat!47}.65{\tiny\textpm.12} & \cellcolor{heat!54}.73{\tiny\textpm.11} & \cellcolor{heat!47}.64{\tiny\textpm.09} & \cellcolor{heat!54}.71{\tiny\textpm.09} & \cellcolor{heat!39}.53{\tiny\textpm.13} & \cellcolor{heat!39}.57{\tiny\textpm.09} & \cellcolor{heat!47}.66{\tiny\textpm.10} \\
\cmidrule[0.01em](lr){1-21}
NCP & \cellcolor{heat!47}.66{\tiny\textpm.08} & \cellcolor{heat!39}.54{\tiny\textpm.20} & \cellcolor{heat!47}.61{\tiny\textpm.09} & \cellcolor{heat!39}.59{\tiny\textpm.08} & \cellcolor{heat!47}.68{\tiny\textpm.07} & \cellcolor{heat!31}.51{\tiny\textpm.22} & \cellcolor{heat!54}.71{\tiny\textpm.15} & \cellcolor{heat!31}.49{\tiny\textpm.17} & \cellcolor{heat!62}.83{\tiny\textpm.12} & \cellcolor{heat!47}.62{\tiny\textpm.19} & \cellcolor{heat!16}.31{\tiny\textpm.19} & \cellcolor{heat!16}.36{\tiny\textpm.13} & \cellcolor{heat!31}.46{\tiny\textpm.19} & \cellcolor{heat!62}.80{\tiny\textpm.09} & \cellcolor{heat!31}.50{\tiny\textpm.12} & \cellcolor{heat!54}\textbf{.76}{\tiny\textpm.09} & \cellcolor{heat!47}.68{\tiny\textpm.10} & \cellcolor{heat!39}.57{\tiny\textpm.14} & \cellcolor{heat!47}.67{\tiny\textpm.10} & \cellcolor{heat!54}.71{\tiny\textpm.09} \\
FAC & \cellcolor{heat!39}.56{\tiny\textpm.04} & \cellcolor{heat!54}.71{\tiny\textpm.19} & \cellcolor{heat!31}.53{\tiny\textpm.05} & \cellcolor{heat!39}.58{\tiny\textpm.06} & \cellcolor{heat!39}.58{\tiny\textpm.06} & \cellcolor{heat!39}.54{\tiny\textpm.15} & \cellcolor{heat!47}.62{\tiny\textpm.14} & \cellcolor{heat!31}.50{\tiny\textpm.20} & \cellcolor{heat!47}.62{\tiny\textpm.14} & \cellcolor{heat!39}.58{\tiny\textpm.13} & \cellcolor{heat!31}.45{\tiny\textpm.18} & \cellcolor{heat!39}.56{\tiny\textpm.12} & \cellcolor{heat!47}.65{\tiny\textpm.13} & \cellcolor{heat!31}.50{\tiny\textpm.00} & \cellcolor{heat!31}.51{\tiny\textpm.02} & \cellcolor{heat!31}.51{\tiny\textpm.06} & \cellcolor{heat!39}.56{\tiny\textpm.10} & \cellcolor{heat!31}.50{\tiny\textpm.13} & \cellcolor{heat!31}.49{\tiny\textpm.05} & \cellcolor{heat!31}.50{\tiny\textpm.04} \\
ASC & \cellcolor{heat!54}.69{\tiny\textpm.07} & \cellcolor{heat!47}.69{\tiny\textpm.20} & \cellcolor{heat!54}.75{\tiny\textpm.08} & \cellcolor{heat!47}.68{\tiny\textpm.08} & \cellcolor{heat!62}.77{\tiny\textpm.07} & \cellcolor{heat!31}.50{\tiny\textpm.22} & \cellcolor{heat!62}\textbf{.82}{\tiny\textpm.12} & \cellcolor{heat!8}.22{\tiny\textpm.19} & \cellcolor{heat!39}.55{\tiny\textpm.19} & \cellcolor{heat!39}.61{\tiny\textpm.17} & \cellcolor{heat!23}.38{\tiny\textpm.15} & \cellcolor{heat!54}\textbf{.76}{\tiny\textpm.09} & \cellcolor{heat!54}\textbf{.72}{\tiny\textpm.18} & \cellcolor{heat!54}.76{\tiny\textpm.09} & \cellcolor{heat!70}\textbf{.85}{\tiny\textpm.08} & \cellcolor{heat!47}.64{\tiny\textpm.11} & \cellcolor{heat!54}.71{\tiny\textpm.09} & \cellcolor{heat!54}\textbf{.73}{\tiny\textpm.12} & \cellcolor{heat!47}.68{\tiny\textpm.10} & \cellcolor{heat!54}.74{\tiny\textpm.09} \\
ADC & \cellcolor{heat!54}.71{\tiny\textpm.07} & \cellcolor{heat!16}.31{\tiny\textpm.16} & \cellcolor{heat!54}.75{\tiny\textpm.08} & \cellcolor{heat!54}.70{\tiny\textpm.07} & \cellcolor{heat!54}.76{\tiny\textpm.07} & \cellcolor{heat!39}.55{\tiny\textpm.21} & \cellcolor{heat!62}.80{\tiny\textpm.12} & \cellcolor{heat!16}.34{\tiny\textpm.17} & \cellcolor{heat!39}.58{\tiny\textpm.19} & \cellcolor{heat!47}.67{\tiny\textpm.17} & \cellcolor{heat!31}.53{\tiny\textpm.18} & \cellcolor{heat!54}.69{\tiny\textpm.11} & \cellcolor{heat!39}.58{\tiny\textpm.18} & \cellcolor{heat!47}.66{\tiny\textpm.11} & \cellcolor{heat!62}.81{\tiny\textpm.09} & \cellcolor{heat!47}.64{\tiny\textpm.11} & \cellcolor{heat!54}\textbf{.72}{\tiny\textpm.09} & \cellcolor{heat!47}.68{\tiny\textpm.16} & \cellcolor{heat!54}.70{\tiny\textpm.11} & \cellcolor{heat!54}.74{\tiny\textpm.09} \\
AEC & \cellcolor{heat!54}.71{\tiny\textpm.07} & \cellcolor{heat!8}.29{\tiny\textpm.19} & \cellcolor{heat!54}.75{\tiny\textpm.08} & \cellcolor{heat!54}.71{\tiny\textpm.08} & \cellcolor{heat!54}.76{\tiny\textpm.07} & \cellcolor{heat!39}.56{\tiny\textpm.20} & \cellcolor{heat!62}.77{\tiny\textpm.13} & \cellcolor{heat!23}.42{\tiny\textpm.18} & \cellcolor{heat!47}.63{\tiny\textpm.17} & \cellcolor{heat!54}\textbf{.74}{\tiny\textpm.14} & \cellcolor{heat!31}.52{\tiny\textpm.15} & \cellcolor{heat!54}.71{\tiny\textpm.10} & \cellcolor{heat!31}.50{\tiny\textpm.18} & \cellcolor{heat!23}.43{\tiny\textpm.13} & \cellcolor{heat!47}.61{\tiny\textpm.11} & \cellcolor{heat!47}.68{\tiny\textpm.11} & \cellcolor{heat!54}.71{\tiny\textpm.09} & \cellcolor{heat!47}.65{\tiny\textpm.14} & \cellcolor{heat!54}.69{\tiny\textpm.10} & \cellcolor{heat!54}.71{\tiny\textpm.09} \\
TER & \cellcolor{heat!54}.71{\tiny\textpm.06} & \cellcolor{heat!39}.58{\tiny\textpm.19} & \cellcolor{heat!54}.74{\tiny\textpm.07} & \cellcolor{heat!47}.67{\tiny\textpm.07} & \cellcolor{heat!62}.77{\tiny\textpm.06} & \cellcolor{heat!47}\textbf{.65}{\tiny\textpm.17} & \cellcolor{heat!47}.66{\tiny\textpm.15} & \cellcolor{heat!39}.58{\tiny\textpm.19} & \cellcolor{heat!70}\textbf{.87}{\tiny\textpm.12} & \cellcolor{heat!47}.67{\tiny\textpm.16} & \cellcolor{heat!16}.34{\tiny\textpm.16} & \cellcolor{heat!47}.62{\tiny\textpm.12} & \cellcolor{heat!54}.69{\tiny\textpm.15} & \cellcolor{heat!62}\textbf{.81}{\tiny\textpm.10} & \cellcolor{heat!54}.76{\tiny\textpm.09} & \cellcolor{heat!54}.76{\tiny\textpm.08} & \cellcolor{heat!47}.68{\tiny\textpm.09} & \cellcolor{heat!47}.65{\tiny\textpm.14} & \cellcolor{heat!62}\textbf{.77}{\tiny\textpm.09} & \cellcolor{heat!54}\textbf{.75}{\tiny\textpm.09} \\
\midrule[\heavyrulewidth]
Success rate & 54\% & 5\% & 26\% & 65\% & 49\% & 24\% & 52\% & 20\% & 74\% & 74\% & 11\% & 25\% & 11\% & 80\% & 61\% & 37\% & 45\% & 16\% & 68\% & 55\% \\
Minority $n$ & 92 & 10$^\dagger$ & 53 & 70 & 98 & 12$^\dagger$ & 24 & 10$^\dagger$ & 13$^\dagger$ & 13$^\dagger$ & 12$^\dagger$ & 28 & 13$^\dagger$ & 23 & 45 & 42 & 51 & 18$^\dagger$ & 36 & 51 \\
\bottomrule
\end{tabular}
}
\caption{AUROC with uncertainty for every cell of the main table (Table~\ref{tab:auroc-main}): point estimate followed by the half-range $(hi-lo)/2$ of a 95\% task-clustered bootstrap interval (1000 resamples of whole tasks). This compact display does not encode interval asymmetry; exact bounds are in the released numeric grid. Bold marks the highest point estimate per column, as in the main table. Half-ranges of 0.10 to 0.20 are typical, and the widest intervals fall in the columns flagged below, so most within-column differences between the leading scorers are not resolved; the family-level patterns discussed in Section~\ref{sec:results} rest on the direction of these estimates across cells rather than on individual cell rankings. These are marginal intervals, which are conservative for comparing two scorers on the same tasks; the paired bootstrap of Table~\ref{tab:airline-paired-bootstrap} resamples the two scorers together and resolves differences that the marginal intervals here leave overlapping. \textsuperscript{$\dagger$}Minority class $n<20$; point estimates and bold marks in these columns should not be read as resolved rankings.}
\label{tab:auroc-main-ci}
\end{table*}

\begin{table*}[t]
\centering
\scriptsize
\setlength{\tabcolsep}{1.2pt}
\resizebox{\textwidth}{!}{%
\begin{tabular}{lcccccccccccccccccccc}
\toprule
 & \multicolumn{5}{c}{BFCL-v4} & \multicolumn{5}{c}{Airline} & \multicolumn{5}{c}{Telecom} & \multicolumn{5}{c}{Retail} \\
\cmidrule(lr){2-6}
\cmidrule(lr){7-11}
\cmidrule(lr){12-16}
\cmidrule(lr){17-21}
Scorer & \rotatebox{90}{\texttt{gpt-4o-mini}} & \rotatebox{90}{\texttt{gpt-oss-20b}} & \rotatebox{90}{\texttt{Qwen2.5-7B}} & \rotatebox{90}{\texttt{Qwen3.5-9B}} & \rotatebox{90}{\texttt{MiniMax-M3}} & \rotatebox{90}{\texttt{gpt-4o-mini}} & \rotatebox{90}{\texttt{gpt-oss-20b}} & \rotatebox{90}{\texttt{Qwen2.5-7B}} & \rotatebox{90}{\texttt{Qwen3.5-9B}} & \rotatebox{90}{\texttt{MiniMax-M3}} & \rotatebox{90}{\texttt{gpt-4o-mini}} & \rotatebox{90}{\texttt{gpt-oss-20b}} & \rotatebox{90}{\texttt{Qwen2.5-7B}} & \rotatebox{90}{\texttt{Qwen3.5-9B}} & \rotatebox{90}{\texttt{MiniMax-M3}} & \rotatebox{90}{\texttt{gpt-4o-mini}} & \rotatebox{90}{\texttt{gpt-oss-20b}} & \rotatebox{90}{\texttt{Qwen2.5-7B}} & \rotatebox{90}{\texttt{Qwen3.5-9B}} & \rotatebox{90}{\texttt{MiniMax-M3}} \\
\midrule
SP$_{\mathrm{first}}$ & \cellcolor{heat!54}0.66 & \cellcolor{heat!39}0.53 & \cellcolor{heat!39}0.53 & \cellcolor{heat!39}0.52 & \cellcolor{heat!62}\textbf{0.69} & \cellcolor{heat!39}0.53 & \cellcolor{heat!47}0.56 & \cellcolor{heat!39}0.50 & \cellcolor{heat!16}0.33 & \cellcolor{heat!39}0.51 & \cellcolor{heat!47}0.61 & \cellcolor{heat!31}0.45 & \cellcolor{heat!54}0.64 & \cellcolor{heat!8}0.23 & \cellcolor{heat!47}0.57 & \cellcolor{heat!47}0.60 & \cellcolor{heat!39}0.55 & \cellcolor{heat!39}0.53 & \cellcolor{heat!47}0.56 & \cellcolor{heat!39}0.52 \\
SP$_{\mathrm{early}}$ & \cellcolor{heat!54}0.63 & \cellcolor{heat!39}0.51 & \cellcolor{heat!54}0.66 & \cellcolor{heat!47}0.59 & \cellcolor{heat!54}0.63 & \cellcolor{heat!31}0.43 & \cellcolor{heat!47}0.57 & \cellcolor{heat!16}0.30 & \cellcolor{heat!31}0.46 & \cellcolor{heat!16}0.31 & \cellcolor{heat!23}0.38 & \cellcolor{heat!31}0.42 & \cellcolor{heat!47}0.61 & \cellcolor{heat!70}\textbf{0.79} & \cellcolor{heat!31}0.45 & \cellcolor{heat!47}0.61 & \cellcolor{heat!23}0.41 & \cellcolor{heat!31}0.44 & \cellcolor{heat!39}0.53 & \cellcolor{heat!47}\textbf{0.57} \\
SP$_{\mathrm{mean}}$ & \cellcolor{heat!47}0.62 & \cellcolor{heat!39}0.52 & \cellcolor{heat!54}0.65 & \cellcolor{heat!47}0.59 & \cellcolor{heat!54}0.66 & \cellcolor{heat!31}0.47 & \cellcolor{heat!39}0.52 & \cellcolor{heat!16}0.30 & \cellcolor{heat!31}0.47 & \cellcolor{heat!16}0.33 & \cellcolor{heat!23}0.38 & \cellcolor{heat!23}0.36 & \cellcolor{heat!39}0.52 & \cellcolor{heat!62}0.73 & \cellcolor{heat!31}0.43 & \cellcolor{heat!54}0.64 & \cellcolor{heat!16}0.32 & \cellcolor{heat!31}0.48 & \cellcolor{heat!47}0.56 & \cellcolor{heat!47}0.56 \\
SP$_{\mathrm{min}}$ & \cellcolor{heat!47}0.59 & \cellcolor{heat!31}0.45 & \cellcolor{heat!47}0.62 & \cellcolor{heat!47}0.60 & \cellcolor{heat!39}0.54 & \cellcolor{heat!31}0.45 & \cellcolor{heat!47}0.58 & \cellcolor{heat!31}0.48 & \cellcolor{heat!47}\textbf{0.59} & \cellcolor{heat!47}0.59 & \cellcolor{heat!8}0.24 & \cellcolor{heat!47}\textbf{0.61} & \cellcolor{heat!54}0.63 & \cellcolor{heat!39}0.50 & \cellcolor{heat!23}0.36 & \cellcolor{heat!47}0.57 & \cellcolor{heat!39}0.55 & \cellcolor{heat!39}0.49 & \cellcolor{heat!47}\textbf{0.60} & \cellcolor{heat!39}0.50 \\
SP$_{\mathrm{late}}$ & \cellcolor{heat!47}0.58 & \cellcolor{heat!39}0.50 & \cellcolor{heat!47}0.61 & \cellcolor{heat!47}0.57 & \cellcolor{heat!54}0.65 & \cellcolor{heat!39}0.51 & \cellcolor{heat!31}0.47 & \cellcolor{heat!23}0.38 & \cellcolor{heat!39}0.50 & \cellcolor{heat!23}0.35 & \cellcolor{heat!23}0.37 & \cellcolor{heat!16}0.33 & \cellcolor{heat!31}0.47 & \cellcolor{heat!47}0.55 & \cellcolor{heat!23}0.40 & \cellcolor{heat!54}0.65 & \cellcolor{heat!16}0.29 & \cellcolor{heat!39}0.52 & \cellcolor{heat!47}0.59 & \cellcolor{heat!47}0.55 \\
SP$_{\mathrm{last}}$ & \cellcolor{heat!39}0.53 & \cellcolor{heat!23}0.39 & \cellcolor{heat!39}0.51 & \cellcolor{heat!39}0.55 & \cellcolor{heat!39}0.51 & \cellcolor{heat!47}\textbf{0.56} & \cellcolor{heat!31}0.46 & \cellcolor{heat!47}\textbf{0.58} & \cellcolor{heat!47}0.58 & \cellcolor{heat!47}\textbf{0.62} & \cellcolor{heat!54}0.63 & \cellcolor{heat!39}0.50 & \cellcolor{heat!39}0.49 & \cellcolor{heat!31}0.42 & \cellcolor{heat!47}\textbf{0.59} & \cellcolor{heat!47}0.60 & \cellcolor{heat!31}0.42 & \cellcolor{heat!39}0.51 & \cellcolor{heat!31}0.46 & \cellcolor{heat!39}0.48 \\
LNSP$_{\mathrm{first}}$ & \cellcolor{heat!54}0.66 & \cellcolor{heat!47}\textbf{0.56} & \cellcolor{heat!39}0.54 & \cellcolor{heat!39}0.53 & \cellcolor{heat!54}0.65 & \cellcolor{heat!39}0.54 & \cellcolor{heat!47}0.55 & \cellcolor{heat!39}0.53 & \cellcolor{heat!16}0.33 & \cellcolor{heat!39}0.55 & \cellcolor{heat!62}\textbf{0.70} & \cellcolor{heat!31}0.47 & \cellcolor{heat!47}0.57 & \cellcolor{heat!8}0.23 & \cellcolor{heat!47}0.58 & \cellcolor{heat!47}0.61 & \cellcolor{heat!47}0.59 & \cellcolor{heat!39}0.52 & \cellcolor{heat!47}0.59 & \cellcolor{heat!39}0.53 \\
LNSP$_{\mathrm{early}}$ & \cellcolor{heat!62}0.70 & \cellcolor{heat!39}0.55 & \cellcolor{heat!62}0.69 & \cellcolor{heat!47}0.59 & \cellcolor{heat!47}0.62 & \cellcolor{heat!23}0.41 & \cellcolor{heat!39}0.53 & 0.19 & \cellcolor{heat!23}0.40 & \cellcolor{heat!8}0.26 & \cellcolor{heat!23}0.37 & \cellcolor{heat!31}0.43 & \cellcolor{heat!39}0.54 & \cellcolor{heat!47}0.60 & \cellcolor{heat!23}0.39 & \cellcolor{heat!54}0.63 & \cellcolor{heat!39}0.50 & \cellcolor{heat!31}0.47 & \cellcolor{heat!47}0.59 & \cellcolor{heat!39}0.50 \\
LNSP$_{\mathrm{mean}}$ & \cellcolor{heat!62}0.70 & \cellcolor{heat!39}0.52 & \cellcolor{heat!54}0.68 & \cellcolor{heat!47}0.60 & \cellcolor{heat!54}0.65 & \cellcolor{heat!31}0.41 & \cellcolor{heat!39}0.52 & 0.17 & \cellcolor{heat!16}0.34 & \cellcolor{heat!8}0.25 & \cellcolor{heat!23}0.37 & \cellcolor{heat!23}0.39 & \cellcolor{heat!39}0.53 & \cellcolor{heat!16}0.34 & \cellcolor{heat!31}0.42 & \cellcolor{heat!62}0.69 & \cellcolor{heat!31}0.47 & \cellcolor{heat!39}0.50 & \cellcolor{heat!47}0.60 & \cellcolor{heat!39}0.50 \\
LNSP$_{\mathrm{min}}$ & \cellcolor{heat!54}0.68 & \cellcolor{heat!31}0.45 & \cellcolor{heat!54}0.65 & \cellcolor{heat!47}0.62 & \cellcolor{heat!54}0.64 & \cellcolor{heat!31}0.45 & \cellcolor{heat!47}0.59 & \cellcolor{heat!23}0.39 & \cellcolor{heat!23}0.37 & \cellcolor{heat!23}0.39 & \cellcolor{heat!8}0.24 & \cellcolor{heat!47}0.60 & \cellcolor{heat!47}0.59 & \cellcolor{heat!39}0.51 & \cellcolor{heat!23}0.37 & \cellcolor{heat!62}0.72 & \cellcolor{heat!47}\textbf{0.59} & \cellcolor{heat!39}0.55 & \cellcolor{heat!47}0.59 & \cellcolor{heat!39}0.52 \\
LNSP$_{\mathrm{late}}$ & \cellcolor{heat!54}0.67 & \cellcolor{heat!31}0.47 & \cellcolor{heat!54}0.65 & \cellcolor{heat!47}0.60 & \cellcolor{heat!54}0.66 & \cellcolor{heat!23}0.41 & \cellcolor{heat!39}0.50 & 0.18 & \cellcolor{heat!16}0.34 & \cellcolor{heat!8}0.25 & \cellcolor{heat!23}0.36 & \cellcolor{heat!23}0.37 & \cellcolor{heat!39}0.51 & 0.21 & \cellcolor{heat!31}0.42 & \cellcolor{heat!62}0.70 & \cellcolor{heat!31}0.46 & \cellcolor{heat!39}0.51 & \cellcolor{heat!47}0.59 & \cellcolor{heat!39}0.50 \\
LNSP$_{\mathrm{last}}$ & \cellcolor{heat!47}0.56 & \cellcolor{heat!23}0.36 & \cellcolor{heat!39}0.51 & \cellcolor{heat!47}0.56 & \cellcolor{heat!39}0.52 & \cellcolor{heat!39}0.55 & \cellcolor{heat!39}0.50 & \cellcolor{heat!39}0.48 & \cellcolor{heat!39}0.49 & \cellcolor{heat!39}0.51 & \cellcolor{heat!54}0.64 & \cellcolor{heat!31}0.47 & \cellcolor{heat!31}0.43 & \cellcolor{heat!31}0.43 & \cellcolor{heat!31}0.45 & \cellcolor{heat!54}0.67 & \cellcolor{heat!31}0.44 & \cellcolor{heat!39}0.53 & \cellcolor{heat!31}0.44 & \cellcolor{heat!39}0.52 \\
ATN@5$_{\mathrm{first}}$ & \cellcolor{heat!54}0.67 & \cellcolor{heat!39}0.55 & \cellcolor{heat!39}0.55 & \cellcolor{heat!39}0.53 & \cellcolor{heat!54}0.64 & \cellcolor{heat!39}0.54 & \cellcolor{heat!39}0.54 & \cellcolor{heat!39}0.53 & \cellcolor{heat!16}0.32 & \cellcolor{heat!39}0.54 & \cellcolor{heat!47}0.62 & \cellcolor{heat!31}0.47 & \cellcolor{heat!47}0.61 & 0.20 & \cellcolor{heat!47}0.57 & \cellcolor{heat!47}0.60 & \cellcolor{heat!47}0.58 & \cellcolor{heat!47}\textbf{0.56} & \cellcolor{heat!47}0.60 & \cellcolor{heat!39}0.54 \\
ATN@5$_{\mathrm{early}}$ & \cellcolor{heat!62}0.71 & \cellcolor{heat!39}0.53 & \cellcolor{heat!62}\textbf{0.71} & \cellcolor{heat!47}0.60 & \cellcolor{heat!47}0.61 & \cellcolor{heat!23}0.41 & \cellcolor{heat!39}0.53 & 0.20 & \cellcolor{heat!23}0.39 & \cellcolor{heat!8}0.25 & \cellcolor{heat!23}0.37 & \cellcolor{heat!31}0.42 & \cellcolor{heat!47}0.56 & \cellcolor{heat!47}0.62 & \cellcolor{heat!23}0.40 & \cellcolor{heat!54}0.62 & \cellcolor{heat!31}0.47 & \cellcolor{heat!31}0.48 & \cellcolor{heat!47}0.59 & \cellcolor{heat!39}0.49 \\
ATN@5$_{\mathrm{mean}}$ & \cellcolor{heat!62}\textbf{0.71} & \cellcolor{heat!39}0.50 & \cellcolor{heat!62}0.70 & \cellcolor{heat!47}0.62 & \cellcolor{heat!54}0.66 & \cellcolor{heat!23}0.41 & \cellcolor{heat!39}0.50 & 0.18 & \cellcolor{heat!23}0.36 & \cellcolor{heat!8}0.24 & \cellcolor{heat!23}0.38 & \cellcolor{heat!23}0.38 & \cellcolor{heat!39}0.54 & \cellcolor{heat!23}0.35 & \cellcolor{heat!31}0.43 & \cellcolor{heat!54}0.68 & \cellcolor{heat!31}0.44 & \cellcolor{heat!39}0.49 & \cellcolor{heat!47}0.60 & \cellcolor{heat!39}0.49 \\
ATN@5$_{\mathrm{min}}$ & \cellcolor{heat!54}0.68 & \cellcolor{heat!31}0.42 & \cellcolor{heat!54}0.65 & \cellcolor{heat!54}\textbf{0.63} & \cellcolor{heat!54}0.65 & \cellcolor{heat!31}0.45 & \cellcolor{heat!47}\textbf{0.59} & \cellcolor{heat!23}0.37 & \cellcolor{heat!23}0.38 & \cellcolor{heat!16}0.34 & \cellcolor{heat!16}0.28 & \cellcolor{heat!47}0.57 & \cellcolor{heat!54}\textbf{0.66} & \cellcolor{heat!39}0.50 & \cellcolor{heat!23}0.37 & \cellcolor{heat!62}0.71 & \cellcolor{heat!47}0.59 & \cellcolor{heat!39}0.55 & \cellcolor{heat!47}0.58 & \cellcolor{heat!39}0.50 \\
ATN@5$_{\mathrm{late}}$ & \cellcolor{heat!54}0.68 & \cellcolor{heat!31}0.46 & \cellcolor{heat!54}0.66 & \cellcolor{heat!47}0.61 & \cellcolor{heat!54}0.66 & \cellcolor{heat!23}0.41 & \cellcolor{heat!39}0.49 & 0.20 & \cellcolor{heat!23}0.35 & \cellcolor{heat!8}0.25 & \cellcolor{heat!23}0.38 & \cellcolor{heat!23}0.37 & \cellcolor{heat!39}0.50 & 0.21 & \cellcolor{heat!31}0.42 & \cellcolor{heat!62}0.70 & \cellcolor{heat!31}0.43 & \cellcolor{heat!39}0.50 & \cellcolor{heat!47}0.57 & \cellcolor{heat!39}0.49 \\
ATN@5$_{\mathrm{last}}$ & \cellcolor{heat!47}0.58 & \cellcolor{heat!23}0.35 & \cellcolor{heat!39}0.54 & \cellcolor{heat!47}0.57 & \cellcolor{heat!39}0.52 & \cellcolor{heat!47}0.55 & \cellcolor{heat!39}0.49 & \cellcolor{heat!39}0.49 & \cellcolor{heat!39}0.49 & \cellcolor{heat!39}0.51 & \cellcolor{heat!54}0.63 & \cellcolor{heat!31}0.48 & \cellcolor{heat!23}0.40 & \cellcolor{heat!31}0.42 & \cellcolor{heat!39}0.49 & \cellcolor{heat!54}0.67 & \cellcolor{heat!31}0.42 & \cellcolor{heat!39}0.51 & \cellcolor{heat!31}0.43 & \cellcolor{heat!39}0.51 \\
PM$_{\mathrm{first}}$ & \cellcolor{heat!54}0.66 & \cellcolor{heat!47}0.55 & \cellcolor{heat!39}0.54 & \cellcolor{heat!39}0.53 & \cellcolor{heat!54}0.65 & \cellcolor{heat!39}0.54 & \cellcolor{heat!39}0.55 & \cellcolor{heat!39}0.51 & \cellcolor{heat!16}0.32 & \cellcolor{heat!39}0.52 & \cellcolor{heat!47}0.58 & \cellcolor{heat!39}0.49 & \cellcolor{heat!47}0.57 & 0.19 & \cellcolor{heat!47}0.58 & \cellcolor{heat!47}0.59 & \cellcolor{heat!47}0.55 & \cellcolor{heat!47}0.55 & \cellcolor{heat!47}0.57 & \cellcolor{heat!39}0.53 \\
PM$_{\mathrm{early}}$ & \cellcolor{heat!62}0.69 & \cellcolor{heat!39}0.54 & \cellcolor{heat!54}0.67 & \cellcolor{heat!47}0.60 & \cellcolor{heat!47}0.61 & \cellcolor{heat!23}0.41 & \cellcolor{heat!39}0.53 & 0.19 & \cellcolor{heat!23}0.39 & \cellcolor{heat!8}0.27 & \cellcolor{heat!23}0.36 & \cellcolor{heat!31}0.41 & \cellcolor{heat!39}0.54 & \cellcolor{heat!47}0.60 & \cellcolor{heat!23}0.40 & \cellcolor{heat!47}0.60 & \cellcolor{heat!31}0.48 & \cellcolor{heat!31}0.47 & \cellcolor{heat!47}0.58 & \cellcolor{heat!39}0.49 \\
PM$_{\mathrm{mean}}$ & \cellcolor{heat!62}0.69 & \cellcolor{heat!39}0.50 & \cellcolor{heat!54}0.66 & \cellcolor{heat!47}0.61 & \cellcolor{heat!54}0.66 & \cellcolor{heat!23}0.40 & \cellcolor{heat!39}0.51 & 0.18 & \cellcolor{heat!16}0.34 & \cellcolor{heat!8}0.25 & \cellcolor{heat!23}0.36 & \cellcolor{heat!23}0.37 & \cellcolor{heat!39}0.51 & \cellcolor{heat!23}0.35 & \cellcolor{heat!31}0.42 & \cellcolor{heat!54}0.66 & \cellcolor{heat!31}0.45 & \cellcolor{heat!39}0.49 & \cellcolor{heat!47}0.59 & \cellcolor{heat!39}0.49 \\
PM$_{\mathrm{min}}$ & \cellcolor{heat!54}0.67 & \cellcolor{heat!31}0.42 & \cellcolor{heat!54}0.64 & \cellcolor{heat!47}0.61 & \cellcolor{heat!54}0.65 & \cellcolor{heat!31}0.44 & \cellcolor{heat!47}0.57 & \cellcolor{heat!23}0.35 & \cellcolor{heat!23}0.37 & \cellcolor{heat!23}0.35 & \cellcolor{heat!8}0.24 & \cellcolor{heat!47}0.59 & \cellcolor{heat!47}0.58 & \cellcolor{heat!39}0.49 & \cellcolor{heat!23}0.36 & \cellcolor{heat!62}\textbf{0.73} & \cellcolor{heat!47}0.58 & \cellcolor{heat!47}0.55 & \cellcolor{heat!47}0.58 & \cellcolor{heat!39}0.50 \\
PM$_{\mathrm{late}}$ & \cellcolor{heat!54}0.66 & \cellcolor{heat!31}0.46 & \cellcolor{heat!54}0.63 & \cellcolor{heat!47}0.60 & \cellcolor{heat!54}0.66 & \cellcolor{heat!23}0.40 & \cellcolor{heat!39}0.49 & 0.20 & \cellcolor{heat!23}0.35 & \cellcolor{heat!8}0.25 & \cellcolor{heat!23}0.36 & \cellcolor{heat!23}0.37 & \cellcolor{heat!39}0.49 & 0.20 & \cellcolor{heat!31}0.42 & \cellcolor{heat!62}0.69 & \cellcolor{heat!31}0.43 & \cellcolor{heat!39}0.50 & \cellcolor{heat!47}0.57 & \cellcolor{heat!31}0.48 \\
PM$_{\mathrm{last}}$ & \cellcolor{heat!47}0.56 & \cellcolor{heat!23}0.36 & \cellcolor{heat!39}0.51 & \cellcolor{heat!47}0.56 & \cellcolor{heat!39}0.51 & \cellcolor{heat!39}0.54 & \cellcolor{heat!39}0.50 & \cellcolor{heat!39}0.49 & \cellcolor{heat!39}0.49 & \cellcolor{heat!39}0.51 & \cellcolor{heat!54}0.63 & \cellcolor{heat!31}0.47 & \cellcolor{heat!31}0.42 & \cellcolor{heat!31}0.43 & \cellcolor{heat!31}0.47 & \cellcolor{heat!54}0.67 & \cellcolor{heat!31}0.43 & \cellcolor{heat!39}0.51 & \cellcolor{heat!31}0.43 & \cellcolor{heat!39}0.50 \\
\midrule[\heavyrulewidth]
Success rate & 54\% & 5\% & 26\% & 65\% & 49\% & 24\% & 52\% & 20\% & 74\% & 74\% & 11\% & 25\% & 11\% & 80\% & 61\% & 37\% & 45\% & 16\% & 68\% & 55\% \\
Minority $n$ & 92 & 10$^\dagger$ & 53 & 70 & 98 & 12$^\dagger$ & 24 & 10$^\dagger$ & 13$^\dagger$ & 13$^\dagger$ & 12$^\dagger$ & 28 & 13$^\dagger$ & 23 & 45 & 42 & 51 & 18$^\dagger$ & 36 & 51 \\
\bottomrule
\end{tabular}
}
\caption{AUROC for the full white-box grid, all (base score, cross-turn aggregation) pairs including PM and the early and late position-weighted aggregations, on the greedy reference basis. This grid covers the white-box family only; reflexive and consistency scorers appear in the main table and in Tables~\ref{tab:prr} and \ref{tab:ece}. Cell shading is the AUROC relative to the whole table. Bold marks the highest point estimate per column; the bottom row gives each model's success rate. Higher is better; values below 0.5 suggest possible anti-correlation with success.}
\label{tab:auroc-full-greedy}
\end{table*}

\begin{table*}[t]
\centering
\scriptsize
\setlength{\tabcolsep}{1.2pt}
\resizebox{\textwidth}{!}{%
\begin{tabular}{lcccccccccccccccccccc}
\toprule
 & \multicolumn{5}{c}{BFCL-v4} & \multicolumn{5}{c}{Airline} & \multicolumn{5}{c}{Telecom} & \multicolumn{5}{c}{Retail} \\
\cmidrule(lr){2-6}
\cmidrule(lr){7-11}
\cmidrule(lr){12-16}
\cmidrule(lr){17-21}
Scorer & \rotatebox{90}{\texttt{gpt-4o-mini}} & \rotatebox{90}{\texttt{gpt-oss-20b}} & \rotatebox{90}{\texttt{Qwen2.5-7B}} & \rotatebox{90}{\texttt{Qwen3.5-9B}} & \rotatebox{90}{\texttt{MiniMax-M3}} & \rotatebox{90}{\texttt{gpt-4o-mini}} & \rotatebox{90}{\texttt{gpt-oss-20b}} & \rotatebox{90}{\texttt{Qwen2.5-7B}} & \rotatebox{90}{\texttt{Qwen3.5-9B}} & \rotatebox{90}{\texttt{MiniMax-M3}} & \rotatebox{90}{\texttt{gpt-4o-mini}} & \rotatebox{90}{\texttt{gpt-oss-20b}} & \rotatebox{90}{\texttt{Qwen2.5-7B}} & \rotatebox{90}{\texttt{Qwen3.5-9B}} & \rotatebox{90}{\texttt{MiniMax-M3}} & \rotatebox{90}{\texttt{gpt-4o-mini}} & \rotatebox{90}{\texttt{gpt-oss-20b}} & \rotatebox{90}{\texttt{Qwen2.5-7B}} & \rotatebox{90}{\texttt{Qwen3.5-9B}} & \rotatebox{90}{\texttt{MiniMax-M3}} \\
\midrule
SP$_{\mathrm{first}}$ & \cellcolor{heat!54}0.31 & \cellcolor{heat!39}0.01 & \cellcolor{heat!47}0.03 & \cellcolor{heat!39}$-0.01$ & \cellcolor{heat!54}0.40 & \cellcolor{heat!39}$-0.04$ & \cellcolor{heat!47}0.21 & \cellcolor{heat!39}$-0.00$ & \cellcolor{heat!16}$-0.68$ & \cellcolor{heat!39}$-0.04$ & \cellcolor{heat!47}0.10 & \cellcolor{heat!39}$-0.07$ & \cellcolor{heat!47}0.22 & \cellcolor{heat!23}$-0.65$ & \cellcolor{heat!47}0.15 & \cellcolor{heat!47}0.21 & \cellcolor{heat!47}0.21 & \cellcolor{heat!47}0.05 & \cellcolor{heat!47}0.20 & \cellcolor{heat!47}0.06 \\
SP$_{\mathrm{mean}}$ & \cellcolor{heat!47}0.15 & \cellcolor{heat!39}$-0.00$ & \cellcolor{heat!47}0.15 & \cellcolor{heat!47}0.17 & \cellcolor{heat!47}0.21 & \cellcolor{heat!39}$-0.05$ & \cellcolor{heat!47}0.06 & \cellcolor{heat!31}$-0.28$ & \cellcolor{heat!47}0.08 & \cellcolor{heat!16}$-0.69$ & \cellcolor{heat!39}$-0.20$ & \cellcolor{heat!31}$-0.24$ & \cellcolor{heat!39}$-0.04$ & \cellcolor{heat!54}0.26 & \cellcolor{heat!31}$-0.42$ & \cellcolor{heat!47}0.17 & \cellcolor{heat!31}$-0.37$ & \cellcolor{heat!39}$-0.09$ & \cellcolor{heat!47}0.18 & \cellcolor{heat!47}0.14 \\
SP$_{\mathrm{min}}$ & \cellcolor{heat!47}0.10 & \cellcolor{heat!39}$-0.07$ & \cellcolor{heat!47}0.15 & \cellcolor{heat!54}0.30 & \cellcolor{heat!39}$-0.09$ & \cellcolor{heat!39}$-0.08$ & \cellcolor{heat!54}0.29 & \cellcolor{heat!39}$-0.09$ & \cellcolor{heat!47}0.22 & \cellcolor{heat!47}0.06 & \cellcolor{heat!31}$-0.28$ & \cellcolor{heat!47}0.18 & \cellcolor{heat!54}0.25 & \cellcolor{heat!47}0.20 & \cellcolor{heat!31}$-0.43$ & \cellcolor{heat!54}0.25 & \cellcolor{heat!39}$-0.12$ & \cellcolor{heat!47}0.09 & \cellcolor{heat!54}0.25 & \cellcolor{heat!39}$-0.15$ \\
SP$_{\mathrm{last}}$ & \cellcolor{heat!39}$-0.02$ & \cellcolor{heat!39}$-0.13$ & \cellcolor{heat!39}$-0.05$ & \cellcolor{heat!47}0.04 & \cellcolor{heat!39}$-0.09$ & \cellcolor{heat!47}0.12 & \cellcolor{heat!39}0.00 & \cellcolor{heat!47}\textbf{0.19} & \cellcolor{heat!47}0.23 & \cellcolor{heat!39}0.02 & \cellcolor{heat!54}0.31 & \cellcolor{heat!39}0.00 & \cellcolor{heat!39}$-0.10$ & \cellcolor{heat!23}$-0.50$ & \cellcolor{heat!47}0.10 & \cellcolor{heat!47}0.17 & \cellcolor{heat!31}$-0.31$ & \cellcolor{heat!39}$-0.05$ & \cellcolor{heat!39}$-0.20$ & \cellcolor{heat!39}$-0.10$ \\
LNSP$_{\mathrm{first}}$ & \cellcolor{heat!54}0.31 & \cellcolor{heat!47}0.02 & \cellcolor{heat!47}0.03 & \cellcolor{heat!39}$-0.04$ & \cellcolor{heat!54}0.35 & \cellcolor{heat!47}0.03 & \cellcolor{heat!47}0.21 & \cellcolor{heat!47}0.02 & \cellcolor{heat!16}$-0.68$ & \cellcolor{heat!47}0.12 & \cellcolor{heat!47}0.16 & \cellcolor{heat!39}$-0.04$ & \cellcolor{heat!47}0.13 & \cellcolor{heat!23}$-0.65$ & \cellcolor{heat!47}0.15 & \cellcolor{heat!47}0.17 & \cellcolor{heat!54}0.28 & \cellcolor{heat!47}0.05 & \cellcolor{heat!47}0.24 & \cellcolor{heat!47}0.07 \\
LNSP$_{\mathrm{mean}}$ & \cellcolor{heat!54}0.33 & \cellcolor{heat!39}$-0.02$ & \cellcolor{heat!47}0.21 & \cellcolor{heat!54}0.27 & \cellcolor{heat!47}0.11 & \cellcolor{heat!39}$-0.12$ & \cellcolor{heat!47}0.07 & \cellcolor{heat!31}$-0.40$ & \cellcolor{heat!31}$-0.29$ & \cellcolor{heat!16}$-0.81$ & \cellcolor{heat!39}$-0.20$ & \cellcolor{heat!39}$-0.19$ & \cellcolor{heat!39}$-0.05$ & \cellcolor{heat!23}$-0.49$ & \cellcolor{heat!23}$-0.47$ & \cellcolor{heat!54}0.25 & \cellcolor{heat!39}$-0.16$ & \cellcolor{heat!39}$-0.01$ & \cellcolor{heat!54}0.35 & \cellcolor{heat!31}$-0.21$ \\
LNSP$_{\mathrm{min}}$ & \cellcolor{heat!54}0.33 & \cellcolor{heat!39}$-0.05$ & \cellcolor{heat!47}0.21 & \cellcolor{heat!54}0.35 & \cellcolor{heat!47}0.06 & \cellcolor{heat!39}$-0.07$ & \cellcolor{heat!54}0.30 & \cellcolor{heat!39}$-0.18$ & \cellcolor{heat!39}$-0.17$ & \cellcolor{heat!31}$-0.26$ & \cellcolor{heat!31}$-0.28$ & \cellcolor{heat!47}0.17 & \cellcolor{heat!47}0.19 & \cellcolor{heat!47}0.06 & \cellcolor{heat!23}$-0.47$ & \cellcolor{heat!54}0.44 & \cellcolor{heat!39}0.00 & \cellcolor{heat!47}0.16 & \cellcolor{heat!54}0.26 & \cellcolor{heat!39}$-0.17$ \\
LNSP$_{\mathrm{last}}$ & \cellcolor{heat!47}0.04 & \cellcolor{heat!39}$-0.14$ & \cellcolor{heat!39}$-0.09$ & \cellcolor{heat!47}0.10 & \cellcolor{heat!39}$-0.11$ & \cellcolor{heat!47}0.12 & \cellcolor{heat!47}0.04 & \cellcolor{heat!39}0.01 & \cellcolor{heat!47}0.11 & \cellcolor{heat!39}$-0.11$ & \cellcolor{heat!54}0.33 & \cellcolor{heat!39}$-0.01$ & \cellcolor{heat!39}$-0.14$ & \cellcolor{heat!23}$-0.45$ & \cellcolor{heat!31}$-0.39$ & \cellcolor{heat!54}0.28 & \cellcolor{heat!31}$-0.28$ & \cellcolor{heat!39}$-0.05$ & \cellcolor{heat!31}$-0.33$ & \cellcolor{heat!39}$-0.14$ \\
ATN@5$_{\mathrm{first}}$ & \cellcolor{heat!54}0.31 & \cellcolor{heat!39}0.01 & \cellcolor{heat!47}0.05 & \cellcolor{heat!39}$-0.02$ & \cellcolor{heat!54}0.33 & \cellcolor{heat!39}0.02 & \cellcolor{heat!47}0.22 & \cellcolor{heat!47}0.03 & \cellcolor{heat!16}$-0.71$ & \cellcolor{heat!47}0.11 & \cellcolor{heat!47}0.13 & \cellcolor{heat!39}$-0.04$ & \cellcolor{heat!47}0.18 & \cellcolor{heat!16}$-0.83$ & \cellcolor{heat!47}0.05 & \cellcolor{heat!47}0.16 & \cellcolor{heat!54}0.24 & \cellcolor{heat!47}0.09 & \cellcolor{heat!54}0.24 & \cellcolor{heat!47}0.12 \\
ATN@5$_{\mathrm{mean}}$ & \cellcolor{heat!54}0.36 & \cellcolor{heat!39}$-0.03$ & \cellcolor{heat!47}0.23 & \cellcolor{heat!54}0.30 & \cellcolor{heat!47}0.11 & \cellcolor{heat!39}$-0.18$ & \cellcolor{heat!47}0.04 & \cellcolor{heat!31}$-0.40$ & \cellcolor{heat!31}$-0.28$ & \cellcolor{heat!16}$-0.83$ & \cellcolor{heat!39}$-0.20$ & \cellcolor{heat!39}$-0.20$ & \cellcolor{heat!39}$-0.04$ & \cellcolor{heat!23}$-0.51$ & \cellcolor{heat!23}$-0.46$ & \cellcolor{heat!47}0.23 & \cellcolor{heat!39}$-0.20$ & \cellcolor{heat!39}$-0.02$ & \cellcolor{heat!54}0.34 & \cellcolor{heat!31}$-0.24$ \\
ATN@5$_{\mathrm{min}}$ & \cellcolor{heat!54}0.30 & \cellcolor{heat!39}$-0.07$ & \cellcolor{heat!47}0.22 & \cellcolor{heat!54}0.36 & \cellcolor{heat!47}0.08 & \cellcolor{heat!39}$-0.08$ & \cellcolor{heat!54}0.31 & \cellcolor{heat!39}$-0.20$ & \cellcolor{heat!31}$-0.21$ & \cellcolor{heat!31}$-0.35$ & \cellcolor{heat!31}$-0.25$ & \cellcolor{heat!47}0.12 & \cellcolor{heat!54}0.32 & \cellcolor{heat!39}$-0.19$ & \cellcolor{heat!23}$-0.47$ & \cellcolor{heat!54}0.46 & \cellcolor{heat!39}$-0.01$ & \cellcolor{heat!47}0.14 & \cellcolor{heat!54}0.24 & \cellcolor{heat!39}$-0.19$ \\
ATN@5$_{\mathrm{last}}$ & \cellcolor{heat!47}0.10 & \cellcolor{heat!39}$-0.14$ & \cellcolor{heat!39}$-0.05$ & \cellcolor{heat!47}0.09 & \cellcolor{heat!39}$-0.13$ & \cellcolor{heat!47}0.13 & \cellcolor{heat!47}0.03 & \cellcolor{heat!47}0.02 & \cellcolor{heat!47}0.11 & \cellcolor{heat!39}$-0.11$ & \cellcolor{heat!54}0.32 & \cellcolor{heat!39}0.01 & \cellcolor{heat!39}$-0.16$ & \cellcolor{heat!23}$-0.47$ & \cellcolor{heat!31}$-0.36$ & \cellcolor{heat!54}0.29 & \cellcolor{heat!31}$-0.31$ & \cellcolor{heat!39}$-0.06$ & \cellcolor{heat!31}$-0.35$ & \cellcolor{heat!39}$-0.17$ \\
\cmidrule[0.01em](lr){1-21}
Prop-Dist & \cellcolor{heat!54}0.29 & \cellcolor{heat!39}$-0.07$ & \cellcolor{heat!47}0.17 & \cellcolor{heat!54}0.28 & \cellcolor{heat!47}0.08 & \cellcolor{heat!39}$-0.13$ & \cellcolor{heat!47}0.09 & \cellcolor{heat!31}$-0.35$ & \cellcolor{heat!31}$-0.29$ & \cellcolor{heat!16}$-0.72$ & \cellcolor{heat!31}$-0.25$ & \cellcolor{heat!39}$-0.15$ & \cellcolor{heat!39}$-0.07$ & \cellcolor{heat!23}$-0.60$ & \cellcolor{heat!23}$-0.52$ & \cellcolor{heat!54}0.29 & \cellcolor{heat!39}$-0.07$ & \cellcolor{heat!47}0.03 & \cellcolor{heat!47}0.21 & \cellcolor{heat!31}$-0.30$ \\
Prop-HMM & \cellcolor{heat!47}0.23 & \cellcolor{heat!39}$-0.07$ & \cellcolor{heat!47}0.17 & \cellcolor{heat!47}0.21 & \cellcolor{heat!47}0.06 & \cellcolor{heat!39}$-0.13$ & \cellcolor{heat!47}0.16 & \cellcolor{heat!39}$-0.20$ & \cellcolor{heat!39}$-0.15$ & \cellcolor{heat!16}$-0.72$ & \cellcolor{heat!31}$-0.27$ & \cellcolor{heat!39}$-0.08$ & \cellcolor{heat!39}$-0.13$ & \cellcolor{heat!16}$-0.87$ & \cellcolor{heat!23}$-0.53$ & \cellcolor{heat!54}0.28 & \cellcolor{heat!39}$-0.08$ & \cellcolor{heat!47}0.04 & \cellcolor{heat!47}0.21 & \cellcolor{heat!31}$-0.36$ \\
\cmidrule[0.01em](lr){1-21}
P(True) & \cellcolor{heat!62}0.53 & \cellcolor{heat!54}0.24 & \cellcolor{heat!62}\textbf{0.60} & \cellcolor{heat!62}\textbf{0.57} & \cellcolor{heat!62}0.57 & \cellcolor{heat!39}$-0.09$ & \cellcolor{heat!47}0.21 & \cellcolor{heat!31}$-0.28$ & \cellcolor{heat!31}$-0.35$ & \cellcolor{heat!54}0.30 & \cellcolor{heat!62}\textbf{0.60} & \cellcolor{heat!47}0.21 & \cellcolor{heat!54}0.29 & \cellcolor{heat!62}0.49 & \cellcolor{heat!62}0.52 & \cellcolor{heat!62}0.50 & \cellcolor{heat!47}0.21 & \cellcolor{heat!47}0.21 & \cellcolor{heat!54}0.37 & \cellcolor{heat!47}0.21 \\
VC & \cellcolor{heat!62}0.53 & \cellcolor{heat!54}\textbf{0.29} & \cellcolor{heat!47}0.19 & \cellcolor{heat!39}$-0.03$ & \cellcolor{heat!70}\textbf{0.80} & \cellcolor{heat!47}0.07 & \cellcolor{heat!54}0.38 & \cellcolor{heat!39}$-0.15$ & \cellcolor{heat!47}0.08 & \cellcolor{heat!39}$-0.02$ & \cellcolor{heat!62}0.50 & \cellcolor{heat!47}0.23 & \cellcolor{heat!54}0.33 & \cellcolor{heat!62}0.64 & \cellcolor{heat!54}0.28 & \cellcolor{heat!54}0.43 & \cellcolor{heat!62}\textbf{0.50} & \cellcolor{heat!47}0.14 & \cellcolor{heat!47}0.18 & \cellcolor{heat!54}0.41 \\
\cmidrule[0.01em](lr){1-21}
NCP & \cellcolor{heat!47}0.21 & \cellcolor{heat!39}0.01 & \cellcolor{heat!47}0.21 & \cellcolor{heat!47}0.20 & \cellcolor{heat!54}0.37 & \cellcolor{heat!47}0.10 & \cellcolor{heat!62}0.51 & \cellcolor{heat!39}$-0.12$ & \cellcolor{heat!70}0.76 & \cellcolor{heat!54}0.32 & \cellcolor{heat!39}$-0.13$ & \cellcolor{heat!39}$-0.09$ & \cellcolor{heat!47}0.09 & \cellcolor{heat!70}\textbf{0.73} & \cellcolor{heat!47}0.07 & \cellcolor{heat!62}0.51 & \cellcolor{heat!62}0.50 & \cellcolor{heat!47}0.13 & \cellcolor{heat!54}0.45 & \cellcolor{heat!62}0.49 \\
FAC & \cellcolor{heat!54}0.46 & \cellcolor{heat!54}0.27 & \cellcolor{heat!54}0.34 & \cellcolor{heat!54}0.25 & \cellcolor{heat!62}0.49 & \cellcolor{heat!39}$-0.11$ & \cellcolor{heat!54}0.28 & \cellcolor{heat!47}0.02 & \cellcolor{heat!54}0.24 & \cellcolor{heat!39}$-0.10$ & \cellcolor{heat!39}$-0.15$ & \cellcolor{heat!47}0.05 & \cellcolor{heat!47}0.15 & $-1.22$ & \cellcolor{heat!47}0.10 & \cellcolor{heat!39}$-0.04$ & \cellcolor{heat!47}0.09 & \cellcolor{heat!47}0.09 & \cellcolor{heat!39}$-0.13$ & \cellcolor{heat!31}$-0.27$ \\
ASC & \cellcolor{heat!62}0.67 & \cellcolor{heat!47}0.19 & \cellcolor{heat!62}0.55 & \cellcolor{heat!54}0.38 & \cellcolor{heat!62}0.67 & \cellcolor{heat!47}0.10 & \cellcolor{heat!62}0.61 & \cellcolor{heat!31}$-0.31$ & \cellcolor{heat!47}0.07 & \cellcolor{heat!54}0.43 & \cellcolor{heat!39}$-0.16$ & \cellcolor{heat!54}\textbf{0.45} & \cellcolor{heat!54}\textbf{0.38} & \cellcolor{heat!62}0.66 & \cellcolor{heat!70}0.74 & \cellcolor{heat!54}0.38 & \cellcolor{heat!54}0.38 & \cellcolor{heat!54}\textbf{0.39} & \cellcolor{heat!54}0.35 & \cellcolor{heat!62}0.54 \\
ADC & \cellcolor{heat!62}0.67 & \cellcolor{heat!39}$-0.17$ & \cellcolor{heat!62}0.54 & \cellcolor{heat!54}0.46 & \cellcolor{heat!62}0.66 & \cellcolor{heat!54}0.24 & \cellcolor{heat!62}\textbf{0.63} & \cellcolor{heat!31}$-0.25$ & \cellcolor{heat!47}0.22 & \cellcolor{heat!62}0.50 & \cellcolor{heat!39}0.02 & \cellcolor{heat!54}0.41 & \cellcolor{heat!47}0.19 & \cellcolor{heat!62}0.50 & \cellcolor{heat!62}0.63 & \cellcolor{heat!54}0.41 & \cellcolor{heat!54}0.39 & \cellcolor{heat!54}0.37 & \cellcolor{heat!62}0.48 & \cellcolor{heat!62}\textbf{0.57} \\
AEC & \cellcolor{heat!62}0.67 & \cellcolor{heat!39}$-0.17$ & \cellcolor{heat!62}0.53 & \cellcolor{heat!62}0.48 & \cellcolor{heat!62}0.65 & \cellcolor{heat!54}0.25 & \cellcolor{heat!62}0.61 & \cellcolor{heat!39}$-0.14$ & \cellcolor{heat!54}0.44 & \cellcolor{heat!62}\textbf{0.64} & \cellcolor{heat!39}$-0.03$ & \cellcolor{heat!54}0.43 & \cellcolor{heat!47}0.05 & \cellcolor{heat!39}$-0.02$ & \cellcolor{heat!47}0.17 & \cellcolor{heat!62}0.51 & \cellcolor{heat!54}0.40 & \cellcolor{heat!54}0.28 & \cellcolor{heat!62}0.52 & \cellcolor{heat!62}0.50 \\
TER & \cellcolor{heat!70}\textbf{0.72} & \cellcolor{heat!47}0.08 & \cellcolor{heat!62}0.49 & \cellcolor{heat!54}0.41 & \cellcolor{heat!70}0.70 & \cellcolor{heat!54}\textbf{0.27} & \cellcolor{heat!54}0.36 & \cellcolor{heat!47}0.11 & \cellcolor{heat!70}\textbf{0.80} & \cellcolor{heat!54}0.42 & \cellcolor{heat!31}$-0.22$ & \cellcolor{heat!47}0.16 & \cellcolor{heat!47}0.20 & \cellcolor{heat!47}0.15 & \cellcolor{heat!70}\textbf{0.75} & \cellcolor{heat!62}\textbf{0.54} & \cellcolor{heat!54}0.36 & \cellcolor{heat!54}0.26 & \cellcolor{heat!62}\textbf{0.64} & \cellcolor{heat!62}0.52 \\
\midrule[\heavyrulewidth]
Success rate & 54\% & 5\% & 26\% & 65\% & 49\% & 24\% & 52\% & 20\% & 74\% & 74\% & 11\% & 25\% & 11\% & 80\% & 61\% & 37\% & 45\% & 16\% & 68\% & 55\% \\
Minority $n$ & 92 & 10$^\dagger$ & 53 & 70 & 98 & 12$^\dagger$ & 24 & 10$^\dagger$ & 13$^\dagger$ & 13$^\dagger$ & 12$^\dagger$ & 28 & 13$^\dagger$ & 23 & 45 & 42 & 51 & 18$^\dagger$ & 36 & 51 \\
\bottomrule
\end{tabular}
}
\caption{Prediction rejection ratio (PRR) for predicting trajectory success on BFCL-v4 and the three $\tau^2$-bench domains, all scorers on the greedy reference basis matching the main table: white-box (SP, LNSP, ATN@5), SAUP-inspired propagation controls, and reflexive (P(True), VC) rows score the greedy trajectory, and consistency rows (NCP, FAC, ASC, ADC, AEC, TER) compare it with its three sampled draws where available. Bold marks the highest point estimate per column. The bottom row gives each model's success rate. PRR normalizes the scorer's retained-success curve between random (0) and oracle (1) rejection; negative values are worse than random abstention. Cell shading is the PRR relative to the whole table. Higher is better.}
\label{tab:prr}
\end{table*}

\begin{table*}[t]
\centering
\scriptsize
\setlength{\tabcolsep}{1.2pt}
\resizebox{\textwidth}{!}{%
\begin{tabular}{lcccccccccccccccccccc}
\toprule
 & \multicolumn{5}{c}{BFCL-v4} & \multicolumn{5}{c}{Airline} & \multicolumn{5}{c}{Telecom} & \multicolumn{5}{c}{Retail} \\
\cmidrule(lr){2-6}
\cmidrule(lr){7-11}
\cmidrule(lr){12-16}
\cmidrule(lr){17-21}
Scorer & \rotatebox{90}{\texttt{gpt-4o-mini}} & \rotatebox{90}{\texttt{gpt-oss-20b}} & \rotatebox{90}{\texttt{Qwen2.5-7B}} & \rotatebox{90}{\texttt{Qwen3.5-9B}} & \rotatebox{90}{\texttt{MiniMax-M3}} & \rotatebox{90}{\texttt{gpt-4o-mini}} & \rotatebox{90}{\texttt{gpt-oss-20b}} & \rotatebox{90}{\texttt{Qwen2.5-7B}} & \rotatebox{90}{\texttt{Qwen3.5-9B}} & \rotatebox{90}{\texttt{MiniMax-M3}} & \rotatebox{90}{\texttt{gpt-4o-mini}} & \rotatebox{90}{\texttt{gpt-oss-20b}} & \rotatebox{90}{\texttt{Qwen2.5-7B}} & \rotatebox{90}{\texttt{Qwen3.5-9B}} & \rotatebox{90}{\texttt{MiniMax-M3}} & \rotatebox{90}{\texttt{gpt-4o-mini}} & \rotatebox{90}{\texttt{gpt-oss-20b}} & \rotatebox{90}{\texttt{Qwen2.5-7B}} & \rotatebox{90}{\texttt{Qwen3.5-9B}} & \rotatebox{90}{\texttt{MiniMax-M3}} \\
\midrule
SP$_{\mathrm{first}}$ & \cellcolor{heat!47}0.66 & 0.06 & \cellcolor{heat!16}0.28 & \cellcolor{heat!47}0.65 & \cellcolor{heat!47}0.66 & \cellcolor{heat!16}0.26 & \cellcolor{heat!47}0.62 & \cellcolor{heat!16}0.23 & \cellcolor{heat!47}0.64 & \cellcolor{heat!54}0.74 & \cellcolor{heat!8}0.15 & \cellcolor{heat!16}0.25 & \cellcolor{heat!16}0.22 & \cellcolor{heat!54}0.70 & \cellcolor{heat!47}0.66 & \cellcolor{heat!31}0.48 & \cellcolor{heat!39}0.55 & \cellcolor{heat!16}0.19 & \cellcolor{heat!54}0.74 & \cellcolor{heat!39}0.58 \\
SP$_{\mathrm{mean}}$ & \cellcolor{heat!47}0.60 & 0.06 & \cellcolor{heat!23}0.34 & \cellcolor{heat!54}0.71 & \cellcolor{heat!39}0.59 & \cellcolor{heat!16}0.26 & \cellcolor{heat!39}0.56 & \cellcolor{heat!8}0.15 & \cellcolor{heat!54}0.77 & \cellcolor{heat!47}0.64 & 0.09 & \cellcolor{heat!16}0.19 & \cellcolor{heat!8}0.12 & \cellcolor{heat!62}0.86 & \cellcolor{heat!39}0.52 & \cellcolor{heat!31}0.45 & \cellcolor{heat!23}0.35 & \cellcolor{heat!8}0.15 & \cellcolor{heat!54}0.74 & \cellcolor{heat!47}0.61 \\
SP$_{\mathrm{min}}$ & \cellcolor{heat!39}0.58 & 0.05 & \cellcolor{heat!23}0.34 & \cellcolor{heat!54}0.75 & \cellcolor{heat!31}0.49 & \cellcolor{heat!16}0.24 & \cellcolor{heat!47}0.66 & \cellcolor{heat!16}0.20 & \cellcolor{heat!62}0.80 & \cellcolor{heat!54}0.77 & 0.07 & \cellcolor{heat!23}0.34 & \cellcolor{heat!16}0.26 & \cellcolor{heat!62}0.84 & \cellcolor{heat!39}0.51 & \cellcolor{heat!39}0.50 & \cellcolor{heat!31}0.44 & \cellcolor{heat!16}0.25 & \cellcolor{heat!54}0.76 & \cellcolor{heat!39}0.53 \\
SP$_{\mathrm{last}}$ & \cellcolor{heat!39}0.54 & 0.04 & \cellcolor{heat!16}0.26 & \cellcolor{heat!47}0.67 & \cellcolor{heat!31}0.47 & \cellcolor{heat!23}0.33 & \cellcolor{heat!39}0.55 & \cellcolor{heat!23}\textbf{0.32} & \cellcolor{heat!62}0.80 & \cellcolor{heat!54}0.76 & \cellcolor{heat!23}0.30 & \cellcolor{heat!16}0.28 & \cellcolor{heat!8}0.11 & \cellcolor{heat!54}0.73 & \cellcolor{heat!47}0.65 & \cellcolor{heat!31}0.46 & \cellcolor{heat!23}0.38 & \cellcolor{heat!8}0.16 & \cellcolor{heat!47}0.65 & \cellcolor{heat!39}0.54 \\
LNSP$_{\mathrm{first}}$ & \cellcolor{heat!47}0.66 & 0.06 & \cellcolor{heat!16}0.29 & \cellcolor{heat!47}0.65 & \cellcolor{heat!47}0.64 & \cellcolor{heat!16}0.28 & \cellcolor{heat!47}0.62 & \cellcolor{heat!16}0.24 & \cellcolor{heat!47}0.64 & \cellcolor{heat!54}0.78 & \cellcolor{heat!8}0.17 & \cellcolor{heat!16}0.26 & \cellcolor{heat!8}0.19 & \cellcolor{heat!54}0.70 & \cellcolor{heat!47}0.66 & \cellcolor{heat!31}0.46 & \cellcolor{heat!39}0.58 & \cellcolor{heat!16}0.19 & \cellcolor{heat!54}0.75 & \cellcolor{heat!39}0.59 \\
LNSP$_{\mathrm{mean}}$ & \cellcolor{heat!47}0.67 & 0.06 & \cellcolor{heat!23}0.36 & \cellcolor{heat!54}0.74 & \cellcolor{heat!39}0.56 & \cellcolor{heat!16}0.23 & \cellcolor{heat!39}0.56 & \cellcolor{heat!8}0.13 & \cellcolor{heat!54}0.69 & \cellcolor{heat!47}0.61 & 0.09 & \cellcolor{heat!16}0.20 & \cellcolor{heat!8}0.12 & \cellcolor{heat!54}0.72 & \cellcolor{heat!39}0.52 & \cellcolor{heat!39}0.49 & \cellcolor{heat!31}0.41 & \cellcolor{heat!8}0.18 & \cellcolor{heat!54}0.79 & \cellcolor{heat!39}0.52 \\
LNSP$_{\mathrm{min}}$ & \cellcolor{heat!47}0.67 & 0.05 & \cellcolor{heat!23}0.37 & \cellcolor{heat!54}0.76 & \cellcolor{heat!39}0.55 & \cellcolor{heat!16}0.24 & \cellcolor{heat!47}0.66 & \cellcolor{heat!8}0.17 & \cellcolor{heat!54}0.71 & \cellcolor{heat!54}0.70 & 0.07 & \cellcolor{heat!23}0.33 & \cellcolor{heat!16}0.22 & \cellcolor{heat!62}0.81 & \cellcolor{heat!39}0.51 & \cellcolor{heat!47}0.60 & \cellcolor{heat!31}0.48 & \cellcolor{heat!16}0.26 & \cellcolor{heat!54}0.76 & \cellcolor{heat!39}0.52 \\
LNSP$_{\mathrm{last}}$ & \cellcolor{heat!39}0.56 & 0.04 & \cellcolor{heat!16}0.25 & \cellcolor{heat!47}0.68 & \cellcolor{heat!31}0.47 & \cellcolor{heat!23}0.33 & \cellcolor{heat!39}0.56 & \cellcolor{heat!16}0.24 & \cellcolor{heat!54}0.77 & \cellcolor{heat!54}0.74 & \cellcolor{heat!23}0.31 & \cellcolor{heat!16}0.28 & \cellcolor{heat!8}0.10 & \cellcolor{heat!54}0.74 & \cellcolor{heat!39}0.53 & \cellcolor{heat!39}0.51 & \cellcolor{heat!23}0.39 & \cellcolor{heat!8}0.16 & \cellcolor{heat!47}0.62 & \cellcolor{heat!39}0.53 \\
ATN@5$_{\mathrm{first}}$ & \cellcolor{heat!47}0.66 & 0.06 & \cellcolor{heat!23}0.30 & \cellcolor{heat!47}0.65 & \cellcolor{heat!47}0.63 & \cellcolor{heat!16}0.28 & \cellcolor{heat!47}0.63 & \cellcolor{heat!16}0.24 & \cellcolor{heat!47}0.63 & \cellcolor{heat!54}0.77 & \cellcolor{heat!8}0.16 & \cellcolor{heat!16}0.26 & \cellcolor{heat!16}0.20 & \cellcolor{heat!47}0.68 & \cellcolor{heat!47}0.63 & \cellcolor{heat!31}0.45 & \cellcolor{heat!39}0.56 & \cellcolor{heat!16}0.21 & \cellcolor{heat!54}0.76 & \cellcolor{heat!47}0.61 \\
ATN@5$_{\mathrm{mean}}$ & \cellcolor{heat!47}0.68 & 0.05 & \cellcolor{heat!23}0.38 & \cellcolor{heat!54}0.74 & \cellcolor{heat!39}0.56 & \cellcolor{heat!16}0.21 & \cellcolor{heat!39}0.56 & \cellcolor{heat!8}0.13 & \cellcolor{heat!54}0.69 & \cellcolor{heat!47}0.61 & 0.09 & \cellcolor{heat!16}0.20 & \cellcolor{heat!8}0.12 & \cellcolor{heat!54}0.72 & \cellcolor{heat!39}0.52 & \cellcolor{heat!31}0.48 & \cellcolor{heat!31}0.40 & \cellcolor{heat!8}0.17 & \cellcolor{heat!54}0.79 & \cellcolor{heat!39}0.51 \\
ATN@5$_{\mathrm{min}}$ & \cellcolor{heat!47}0.66 & 0.05 & \cellcolor{heat!23}0.37 & \cellcolor{heat!54}0.77 & \cellcolor{heat!39}0.55 & \cellcolor{heat!16}0.24 & \cellcolor{heat!47}0.66 & \cellcolor{heat!8}0.17 & \cellcolor{heat!54}0.71 & \cellcolor{heat!47}0.68 & 0.08 & \cellcolor{heat!23}0.31 & \cellcolor{heat!23}0.30 & \cellcolor{heat!54}0.78 & \cellcolor{heat!39}0.51 & \cellcolor{heat!47}0.61 & \cellcolor{heat!31}0.47 & \cellcolor{heat!16}0.25 & \cellcolor{heat!54}0.76 & \cellcolor{heat!39}0.52 \\
ATN@5$_{\mathrm{last}}$ & \cellcolor{heat!39}0.58 & 0.04 & \cellcolor{heat!16}0.27 & \cellcolor{heat!47}0.68 & \cellcolor{heat!31}0.47 & \cellcolor{heat!23}0.33 & \cellcolor{heat!39}0.56 & \cellcolor{heat!16}0.24 & \cellcolor{heat!54}0.77 & \cellcolor{heat!54}0.73 & \cellcolor{heat!23}0.31 & \cellcolor{heat!23}0.29 & \cellcolor{heat!8}0.10 & \cellcolor{heat!54}0.74 & \cellcolor{heat!39}0.55 & \cellcolor{heat!39}0.51 & \cellcolor{heat!23}0.38 & \cellcolor{heat!8}0.16 & \cellcolor{heat!47}0.62 & \cellcolor{heat!39}0.52 \\
\cmidrule[0.01em](lr){1-21}
Prop-Dist & \cellcolor{heat!47}0.66 & 0.05 & \cellcolor{heat!23}0.35 & \cellcolor{heat!54}0.74 & \cellcolor{heat!39}0.55 & \cellcolor{heat!16}0.23 & \cellcolor{heat!39}0.57 & \cellcolor{heat!8}0.14 & \cellcolor{heat!47}0.69 & \cellcolor{heat!47}0.62 & 0.08 & \cellcolor{heat!16}0.21 & \cellcolor{heat!8}0.11 & \cellcolor{heat!54}0.71 & \cellcolor{heat!39}0.51 & \cellcolor{heat!39}0.51 & \cellcolor{heat!31}0.45 & \cellcolor{heat!16}0.21 & \cellcolor{heat!54}0.75 & \cellcolor{heat!39}0.50 \\
Prop-HMM & \cellcolor{heat!47}0.64 & 0.05 & \cellcolor{heat!23}0.35 & \cellcolor{heat!54}0.72 & \cellcolor{heat!39}0.53 & \cellcolor{heat!16}0.23 & \cellcolor{heat!47}0.60 & \cellcolor{heat!8}0.17 & \cellcolor{heat!54}0.72 & \cellcolor{heat!47}0.62 & 0.07 & \cellcolor{heat!16}0.24 & \cellcolor{heat!8}0.10 & \cellcolor{heat!47}0.67 & \cellcolor{heat!39}0.50 & \cellcolor{heat!39}0.51 & \cellcolor{heat!31}0.45 & \cellcolor{heat!16}0.21 & \cellcolor{heat!54}0.75 & \cellcolor{heat!31}0.48 \\
\cmidrule[0.01em](lr){1-21}
P(True) & \cellcolor{heat!54}\textbf{0.75} & \cellcolor{heat!8}0.12 & \cellcolor{heat!47}\textbf{0.62} & \cellcolor{heat!62}\textbf{0.83} & \cellcolor{heat!54}0.75 & \cellcolor{heat!16}0.24 & \cellcolor{heat!47}0.62 & \cellcolor{heat!8}0.15 & \cellcolor{heat!47}0.68 & \cellcolor{heat!62}0.82 & \cellcolor{heat!31}\textbf{0.48} & \cellcolor{heat!23}0.36 & \cellcolor{heat!23}0.29 & \cellcolor{heat!70}0.89 & \cellcolor{heat!62}0.80 & \cellcolor{heat!47}0.64 & \cellcolor{heat!39}0.55 & \cellcolor{heat!16}0.25 & \cellcolor{heat!62}0.80 & \cellcolor{heat!47}0.64 \\
VC & \cellcolor{heat!54}0.72 & \cellcolor{heat!8}\textbf{0.16} & \cellcolor{heat!31}0.39 & \cellcolor{heat!54}0.69 & \cellcolor{heat!62}\textbf{0.83} & \cellcolor{heat!16}0.25 & \cellcolor{heat!47}0.67 & \cellcolor{heat!8}0.18 & \cellcolor{heat!54}0.72 & \cellcolor{heat!54}0.77 & \cellcolor{heat!16}0.23 & \cellcolor{heat!23}0.34 & \cellcolor{heat!16}0.21 & \cellcolor{heat!62}0.85 & \cellcolor{heat!54}0.71 & \cellcolor{heat!31}0.46 & \cellcolor{heat!47}0.62 & \cellcolor{heat!8}0.18 & \cellcolor{heat!54}0.72 & \cellcolor{heat!47}0.65 \\
\cmidrule[0.01em](lr){1-21}
NCP & \cellcolor{heat!47}0.63 & 0.07 & \cellcolor{heat!23}0.37 & \cellcolor{heat!54}0.72 & \cellcolor{heat!47}0.65 & \cellcolor{heat!23}0.30 & \cellcolor{heat!54}0.74 & \cellcolor{heat!16}0.20 & \cellcolor{heat!70}\textbf{0.93} & \cellcolor{heat!62}0.82 & \cellcolor{heat!8}0.10 & \cellcolor{heat!16}0.24 & \cellcolor{heat!16}0.20 & \cellcolor{heat!70}\textbf{0.94} & \cellcolor{heat!47}0.64 & \cellcolor{heat!47}0.62 & \cellcolor{heat!54}\textbf{0.70} & \cellcolor{heat!16}0.23 & \cellcolor{heat!62}0.82 & \cellcolor{heat!54}0.76 \\
FAC & \cellcolor{heat!39}0.57 & \cellcolor{heat!8}0.12 & \cellcolor{heat!16}0.28 & \cellcolor{heat!47}0.69 & \cellcolor{heat!39}0.54 & \cellcolor{heat!16}0.26 & \cellcolor{heat!47}0.60 & \cellcolor{heat!16}0.20 & \cellcolor{heat!54}0.79 & \cellcolor{heat!54}0.77 & \cellcolor{heat!8}0.10 & \cellcolor{heat!16}0.29 & \cellcolor{heat!8}0.16 & \cellcolor{heat!62}0.80 & \cellcolor{heat!47}0.61 & \cellcolor{heat!23}0.37 & \cellcolor{heat!31}0.48 & \cellcolor{heat!8}0.16 & \cellcolor{heat!47}0.68 & \cellcolor{heat!39}0.55 \\
ASC & \cellcolor{heat!47}0.67 & \cellcolor{heat!8}0.12 & \cellcolor{heat!31}0.48 & \cellcolor{heat!54}0.75 & \cellcolor{heat!54}0.75 & \cellcolor{heat!23}0.30 & \cellcolor{heat!54}0.79 & \cellcolor{heat!8}0.15 & \cellcolor{heat!54}0.79 & \cellcolor{heat!62}0.85 & 0.09 & \cellcolor{heat!39}\textbf{0.49} & \cellcolor{heat!23}\textbf{0.32} & \cellcolor{heat!70}0.93 & \cellcolor{heat!62}\textbf{0.86} & \cellcolor{heat!39}0.51 & \cellcolor{heat!47}0.63 & \cellcolor{heat!23}0.36 & \cellcolor{heat!62}0.80 & \cellcolor{heat!54}0.76 \\
ADC & \cellcolor{heat!47}0.68 & 0.05 & \cellcolor{heat!39}0.49 & \cellcolor{heat!54}0.77 & \cellcolor{heat!54}0.77 & \cellcolor{heat!31}0.42 & \cellcolor{heat!62}\textbf{0.80} & \cellcolor{heat!8}0.16 & \cellcolor{heat!62}0.80 & \cellcolor{heat!62}0.87 & \cellcolor{heat!8}0.13 & \cellcolor{heat!31}0.48 & \cellcolor{heat!16}0.23 & \cellcolor{heat!70}0.90 & \cellcolor{heat!62}0.84 & \cellcolor{heat!39}0.59 & \cellcolor{heat!47}0.63 & \cellcolor{heat!23}\textbf{0.36} & \cellcolor{heat!62}0.83 & \cellcolor{heat!54}\textbf{0.79} \\
AEC & \cellcolor{heat!47}0.69 & 0.05 & \cellcolor{heat!31}0.48 & \cellcolor{heat!54}0.78 & \cellcolor{heat!54}0.76 & \cellcolor{heat!31}\textbf{0.42} & \cellcolor{heat!54}0.79 & \cellcolor{heat!8}0.18 & \cellcolor{heat!62}0.85 & \cellcolor{heat!70}\textbf{0.90} & \cellcolor{heat!8}0.11 & \cellcolor{heat!39}0.49 & \cellcolor{heat!8}0.15 & \cellcolor{heat!62}0.80 & \cellcolor{heat!47}0.67 & \cellcolor{heat!47}\textbf{0.66} & \cellcolor{heat!47}0.63 & \cellcolor{heat!23}0.31 & \cellcolor{heat!62}\textbf{0.84} & \cellcolor{heat!54}0.76 \\
TER & \cellcolor{heat!47}0.68 & 0.07 & \cellcolor{heat!31}0.45 & \cellcolor{heat!54}0.74 & \cellcolor{heat!54}0.72 & \cellcolor{heat!23}0.34 & \cellcolor{heat!47}0.65 & \cellcolor{heat!16}0.25 & \cellcolor{heat!70}0.92 & \cellcolor{heat!62}0.82 & 0.09 & \cellcolor{heat!23}0.33 & \cellcolor{heat!16}0.20 & \cellcolor{heat!70}0.91 & \cellcolor{heat!54}0.77 & \cellcolor{heat!39}0.57 & \cellcolor{heat!39}0.58 & \cellcolor{heat!16}0.23 & \cellcolor{heat!62}0.84 & \cellcolor{heat!54}0.73 \\
\midrule[\heavyrulewidth]
Success rate & 54\% & 5\% & 26\% & 65\% & 49\% & 24\% & 52\% & 20\% & 74\% & 74\% & 11\% & 25\% & 11\% & 80\% & 61\% & 37\% & 45\% & 16\% & 68\% & 55\% \\
Minority $n$ & 92 & 10$^\dagger$ & 53 & 70 & 98 & 12$^\dagger$ & 24 & 10$^\dagger$ & 13$^\dagger$ & 13$^\dagger$ & 12$^\dagger$ & 28 & 13$^\dagger$ & 23 & 45 & 42 & 51 & 18$^\dagger$ & 36 & 51 \\
\bottomrule
\end{tabular}
}
\caption{Area under the precision-recall curve (AUPRC) for predicting trajectory success on BFCL-v4 and the three $\tau^2$-bench domains, all scorers on the greedy reference basis matching the main table: white-box (SP, LNSP, ATN@5), SAUP-inspired propagation controls, and reflexive (P(True), VC) rows score the greedy trajectory, and consistency rows (NCP, FAC, ASC, ADC, AEC, TER) compare it with its three sampled draws where available. Bold marks the highest point estimate per column. The bottom row gives each model's success rate. AUPRC (average precision for predicting success) complements AUROC under the class imbalance of the low- and high-success domains; the no-skill baseline is a column's success rate (bottom row), not 0.5. Cell shading is the AUPRC relative to the whole table. Higher is better.}
\label{tab:auprc}
\end{table*}

\begin{table*}[t]
\centering
\scriptsize
\setlength{\tabcolsep}{1.2pt}
\resizebox{\textwidth}{!}{%
\begin{tabular}{lcccccccccccccccccccc}
\toprule
 & \multicolumn{5}{c}{BFCL-v4} & \multicolumn{5}{c}{Airline} & \multicolumn{5}{c}{Telecom} & \multicolumn{5}{c}{Retail} \\
\cmidrule(lr){2-6}
\cmidrule(lr){7-11}
\cmidrule(lr){12-16}
\cmidrule(lr){17-21}
Scorer & \rotatebox{90}{\texttt{gpt-4o-mini}} & \rotatebox{90}{\texttt{gpt-oss-20b}} & \rotatebox{90}{\texttt{Qwen2.5-7B}} & \rotatebox{90}{\texttt{Qwen3.5-9B}} & \rotatebox{90}{\texttt{MiniMax-M3}} & \rotatebox{90}{\texttt{gpt-4o-mini}} & \rotatebox{90}{\texttt{gpt-oss-20b}} & \rotatebox{90}{\texttt{Qwen2.5-7B}} & \rotatebox{90}{\texttt{Qwen3.5-9B}} & \rotatebox{90}{\texttt{MiniMax-M3}} & \rotatebox{90}{\texttt{gpt-4o-mini}} & \rotatebox{90}{\texttt{gpt-oss-20b}} & \rotatebox{90}{\texttt{Qwen2.5-7B}} & \rotatebox{90}{\texttt{Qwen3.5-9B}} & \rotatebox{90}{\texttt{MiniMax-M3}} & \rotatebox{90}{\texttt{gpt-4o-mini}} & \rotatebox{90}{\texttt{gpt-oss-20b}} & \rotatebox{90}{\texttt{Qwen2.5-7B}} & \rotatebox{90}{\texttt{Qwen3.5-9B}} & \rotatebox{90}{\texttt{MiniMax-M3}} \\
\midrule
SP$_{\mathrm{mean}}$ & \cellcolor{heat!62}\textbf{0.13} & \cellcolor{heat!39}0.46 & \cellcolor{heat!47}0.31 & \cellcolor{heat!70}\textbf{0.06} & \cellcolor{heat!70}\textbf{0.07} & \cellcolor{heat!54}\textbf{0.22} & \cellcolor{heat!47}0.32 & \cellcolor{heat!54}0.27 & \cellcolor{heat!54}0.21 & \cellcolor{heat!39}0.42 & \cellcolor{heat!47}0.35 & \cellcolor{heat!54}0.23 & \cellcolor{heat!70}\textbf{0.10} & \cellcolor{heat!39}0.42 & \cellcolor{heat!47}0.38 & \cellcolor{heat!62}\textbf{0.14} & \cellcolor{heat!47}0.36 & \cellcolor{heat!54}0.25 & \cellcolor{heat!70}\textbf{0.09} & \cellcolor{heat!62}\textbf{0.12} \\
LNSP$_{\mathrm{mean}}$ & \cellcolor{heat!39}0.44 & 0.89 & \cellcolor{heat!16}0.70 & \cellcolor{heat!47}0.31 & \cellcolor{heat!31}0.48 & \cellcolor{heat!16}0.70 & \cellcolor{heat!47}0.38 & \cellcolor{heat!16}0.74 & \cellcolor{heat!54}0.22 & \cellcolor{heat!54}0.22 & \cellcolor{heat!8}0.83 & \cellcolor{heat!23}0.60 & \cellcolor{heat!8}0.79 & \cellcolor{heat!62}0.15 & \cellcolor{heat!47}0.34 & \cellcolor{heat!23}0.57 & \cellcolor{heat!39}0.43 & \cellcolor{heat!8}0.78 & \cellcolor{heat!47}0.29 & \cellcolor{heat!39}0.41 \\
\cmidrule[0.01em](lr){1-21}
NCP & \cellcolor{heat!39}0.42 & \cellcolor{heat!16}0.68 & \cellcolor{heat!23}0.63 & \cellcolor{heat!47}0.31 & \cellcolor{heat!39}0.40 & \cellcolor{heat!23}0.61 & \cellcolor{heat!47}0.35 & \cellcolor{heat!31}0.55 & \cellcolor{heat!54}0.19 & \cellcolor{heat!62}0.11 & \cellcolor{heat!16}0.73 & \cellcolor{heat!23}0.57 & \cellcolor{heat!16}0.67 & \cellcolor{heat!62}0.15 & \cellcolor{heat!47}0.29 & \cellcolor{heat!39}0.42 & \cellcolor{heat!47}0.30 & \cellcolor{heat!23}0.57 & \cellcolor{heat!54}0.22 & \cellcolor{heat!47}0.31 \\
FAC & \cellcolor{heat!39}0.42 & \cellcolor{heat!39}0.45 & \cellcolor{heat!16}0.68 & \cellcolor{heat!47}0.29 & \cellcolor{heat!39}0.42 & \cellcolor{heat!23}0.57 & \cellcolor{heat!39}0.41 & \cellcolor{heat!23}0.57 & \cellcolor{heat!54}0.20 & \cellcolor{heat!54}0.20 & \cellcolor{heat!16}0.68 & \cellcolor{heat!47}0.33 & \cellcolor{heat!16}0.70 & \cellcolor{heat!54}0.20 & \cellcolor{heat!39}0.39 & \cellcolor{heat!23}0.58 & \cellcolor{heat!47}0.36 & \cellcolor{heat!23}0.64 & \cellcolor{heat!47}0.32 & \cellcolor{heat!39}0.43 \\
ASC & \cellcolor{heat!39}0.41 & \cellcolor{heat!39}0.44 & \cellcolor{heat!23}0.59 & \cellcolor{heat!54}0.27 & \cellcolor{heat!47}0.30 & \cellcolor{heat!31}0.54 & \cellcolor{heat!62}\textbf{0.12} & \cellcolor{heat!31}0.52 & \cellcolor{heat!62}0.15 & \cellcolor{heat!62}\textbf{0.10} & \cellcolor{heat!31}0.49 & \cellcolor{heat!54}0.23 & \cellcolor{heat!47}0.29 & \cellcolor{heat!70}0.09 & \cellcolor{heat!62}0.17 & \cellcolor{heat!39}0.47 & \cellcolor{heat!54}0.21 & \cellcolor{heat!31}0.51 & \cellcolor{heat!54}0.24 & \cellcolor{heat!47}0.31 \\
ADC & \cellcolor{heat!39}0.44 & \cellcolor{heat!31}0.54 & \cellcolor{heat!23}0.65 & \cellcolor{heat!47}0.31 & \cellcolor{heat!39}0.41 & \cellcolor{heat!23}0.65 & \cellcolor{heat!62}0.16 & \cellcolor{heat!23}0.63 & \cellcolor{heat!62}0.19 & \cellcolor{heat!62}0.15 & \cellcolor{heat!23}0.65 & \cellcolor{heat!31}0.48 & \cellcolor{heat!16}0.72 & \cellcolor{heat!62}0.18 & \cellcolor{heat!47}0.33 & \cellcolor{heat!31}0.55 & \cellcolor{heat!47}0.35 & \cellcolor{heat!16}0.69 & \cellcolor{heat!47}0.29 & \cellcolor{heat!39}0.38 \\
AEC & \cellcolor{heat!39}0.40 & \cellcolor{heat!39}0.39 & \cellcolor{heat!31}0.54 & \cellcolor{heat!54}0.24 & \cellcolor{heat!47}0.31 & \cellcolor{heat!39}0.43 & \cellcolor{heat!62}0.19 & \cellcolor{heat!47}0.29 & \cellcolor{heat!62}0.11 & \cellcolor{heat!62}0.14 & \cellcolor{heat!47}0.32 & \cellcolor{heat!54}\textbf{0.19} & \cellcolor{heat!39}0.40 & \cellcolor{heat!62}0.15 & \cellcolor{heat!70}\textbf{0.07} & \cellcolor{heat!47}0.30 & \cellcolor{heat!70}\textbf{0.09} & \cellcolor{heat!47}0.35 & \cellcolor{heat!62}0.12 & \cellcolor{heat!62}0.15 \\
TER & \cellcolor{heat!47}0.31 & \cellcolor{heat!62}\textbf{0.18} & \cellcolor{heat!47}0.35 & \cellcolor{heat!54}0.23 & \cellcolor{heat!62}0.18 & \cellcolor{heat!47}0.30 & \cellcolor{heat!54}0.21 & \cellcolor{heat!54}\textbf{0.21} & \cellcolor{heat!70}\textbf{0.07} & \cellcolor{heat!47}0.33 & \cellcolor{heat!39}0.43 & \cellcolor{heat!54}0.21 & \cellcolor{heat!47}0.29 & \cellcolor{heat!70}\textbf{0.08} & \cellcolor{heat!54}0.19 & \cellcolor{heat!62}0.18 & \cellcolor{heat!54}0.20 & \cellcolor{heat!54}0.26 & \cellcolor{heat!62}0.15 & \cellcolor{heat!54}0.22 \\
\cmidrule[0.01em](lr){1-21}
P(True) & \cellcolor{heat!54}0.24 & \cellcolor{heat!54}0.26 & \cellcolor{heat!54}\textbf{0.23} & \cellcolor{heat!62}0.15 & \cellcolor{heat!62}0.18 & \cellcolor{heat!47}0.29 & \cellcolor{heat!31}0.49 & \cellcolor{heat!31}0.55 & \cellcolor{heat!39}0.45 & \cellcolor{heat!31}0.56 & \cellcolor{heat!62}0.10 & \cellcolor{heat!54}0.25 & \cellcolor{heat!62}0.12 & \cellcolor{heat!39}0.43 & \cellcolor{heat!31}0.51 & \cellcolor{heat!54}0.25 & \cellcolor{heat!39}0.38 & \cellcolor{heat!54}\textbf{0.25} & \cellcolor{heat!54}0.23 & \cellcolor{heat!47}0.32 \\
VC & \cellcolor{heat!54}0.21 & \cellcolor{heat!47}0.31 & \cellcolor{heat!47}0.29 & \cellcolor{heat!47}0.29 & \cellcolor{heat!62}0.17 & \cellcolor{heat!54}0.26 & \cellcolor{heat!39}0.45 & \cellcolor{heat!39}0.38 & \cellcolor{heat!39}0.46 & \cellcolor{heat!31}0.48 & \cellcolor{heat!70}\textbf{0.05} & \cellcolor{heat!54}0.22 & \cellcolor{heat!54}0.26 & \cellcolor{heat!31}0.48 & \cellcolor{heat!47}0.38 & \cellcolor{heat!54}0.25 & \cellcolor{heat!54}0.28 & \cellcolor{heat!31}0.56 & \cellcolor{heat!47}0.34 & \cellcolor{heat!54}0.26 \\
\midrule[\heavyrulewidth]
Success rate & 54\% & 5\% & 26\% & 65\% & 49\% & 24\% & 52\% & 20\% & 74\% & 74\% & 11\% & 25\% & 11\% & 80\% & 61\% & 37\% & 45\% & 16\% & 68\% & 55\% \\
Minority $n$ & 92 & 10$^\dagger$ & 53 & 70 & 98 & 12$^\dagger$ & 24 & 10$^\dagger$ & 13$^\dagger$ & 13$^\dagger$ & 12$^\dagger$ & 28 & 13$^\dagger$ & 23 & 45 & 42 & 51 & 18$^\dagger$ & 36 & 51 \\
\bottomrule
\end{tabular}
}
\caption{Expected calibration error (ECE, 10 equal-width bins) for the scorers whose values lie in $[0,1]$. The remaining white-box scorers, including the SAUP-inspired propagation controls, use raw magnitudes off any probability scale and are omitted; SP and LNSP are per-token sequence probabilities and consistency agreement rates are not calibrated success probabilities, so their ECE reflects raw-confidence miscalibration. Cell shading is inverted so lower (better) ECE is darker. Lower is better; bold marks the lowest point estimate per column. The bottom row gives each model's success rate.}
\label{tab:ece}
\end{table*}

\section{Cheap Trajectory Baselines}
\label{app:trivial-trajectory-baselines}

Table~\ref{tab:trivial-trajectory-baselines} reports two zero-cost trajectory
baselines on the $\tau^2$ greedy trajectories: the number of scored agent turns and
the number of tool calls. We negate both counts so that larger scores predict
success, matching the confidence-score convention used elsewhere. These
baselines are not UQ methods, but they quantify how much of the trajectory
success label can be predicted from simple loop statistics alone.

\begin{table*}[t]
\centering
\small
\setlength{\tabcolsep}{4.0pt}
\begin{tabular}{llrrrrrr}
\toprule
Model & Domain & \multicolumn{3}{c}{Turn count} & \multicolumn{3}{c}{Tool calls} \\
\cmidrule(lr){3-5}\cmidrule(lr){6-8}
 & & AUROC & PRR & Succ/Fail mean & AUROC & PRR & Succ/Fail mean \\
\midrule
\texttt{gpt-4o-mini} & Airline & 0.67 & 0.37 & 9.7/14.0 & 0.69 & 0.29 & 4.7/8.3 \\
\texttt{gpt-4o-mini} & Telecom & 0.40 & -0.06 & 16.1/14.1 & 0.69 & 0.16 & 8.1/9.1 \\
\texttt{gpt-4o-mini} & Retail & 0.74 & 0.52 & 11.1/15.7 & 0.65 & 0.27 & 6.6/8.4 \\
\addlinespace
\texttt{gpt-oss-20b} & Airline & 0.70 & 0.52 & 9.2/11.2 & 0.77 & 0.57 & 5.1/8.0 \\
\texttt{gpt-oss-20b} & Telecom & 0.67 & 0.33 & 14.1/17.9 & 0.72 & 0.29 & 5.8/8.2 \\
\texttt{gpt-oss-20b} & Retail & 0.55 & -0.02 & 11.5/12.3 & 0.54 & -0.10 & 6.1/7.1 \\
\addlinespace
\texttt{Qwen2.5-7B} & Airline & 0.70 & 0.44 & 12.9/20.9 & 0.72 & 0.45 & 6.6/12.8 \\
\texttt{Qwen2.5-7B} & Telecom & 0.83 & 0.57 & 14.2/24.0 & 0.53 & 0.16 & 4.5/5.1 \\
\texttt{Qwen2.5-7B} & Retail & 0.66 & 0.34 & 17.0/19.6 & 0.62 & 0.19 & 10.7/12.4 \\
\addlinespace
\texttt{Qwen3.5-9B} & Airline & 0.70 & 0.30 & 10.5/13.1 & 0.69 & 0.33 & 6.1/7.9 \\
\texttt{Qwen3.5-9B} & Telecom & 0.81 & 0.39 & 17.2/25.4 & 0.67 & 0.60 & 6.4/7.2 \\
\texttt{Qwen3.5-9B} & Retail & 0.59 & 0.33 & 12.2/13.4 & 0.56 & 0.20 & 7.4/7.9 \\
\addlinespace
\texttt{MiniMax-M3} & Airline & 0.78 & 0.68 & 11.7/19.2 & 0.76 & 0.65 & 4.8/11.7 \\
\texttt{MiniMax-M3} & Telecom & 0.52 & -0.03 & 20.1/22.0 & 0.47 & -0.03 & 3.3/3.2 \\
\texttt{MiniMax-M3} & Retail & 0.51 & 0.06 & 12.8/13.2 & 0.38 & -0.31 & 6.6/5.4 \\
\bottomrule
\end{tabular}
\caption{Cheap trajectory-feature baselines on $\tau^2$ greedy trajectories. Scores are negated counts so higher values predict success; Succ/Fail mean reports the raw feature mean among successful and failed trajectories. These baselines quantify how much of the trajectory success label is predictable from simple trajectory length and tool-use statistics alone; they are compared with the UQ scorers side by side, not entered jointly, so they do not establish incremental signal either way.}
\label{tab:trivial-trajectory-baselines}
\end{table*}

\section{Statistical Uncertainty for Main Comparisons}
\label{app:statistical-uncertainty}

\paragraph{Per-cell intervals for the main grid.}
Table~\ref{tab:auroc-main-ci} repeats every cell of the main table with a 95\%
task-clustered bootstrap interval (1000 resamples of whole tasks), and the
numeric grid is released alongside it. The median interval half-width is 0.10
in columns whose minority class holds at least 20 trajectories and 0.17 in the
eight columns below that threshold, which we mark in both tables. Three
consequences bound how the main table should be read. First, per-cell
rankings are mostly unresolved: the leading scorers within a column typically
have overlapping marginal intervals, which is why we treat the family-level
pattern across cells, and the paired comparisons below, as the evidence rather
than any single cell. Second, the best-scoring method per column does separate
from chance in 18 of the 20 columns; the exceptions are two airline columns
whose minority class holds 12 and 10 successes (\texttt{gpt-4o-mini}, TER 0.647 [0.461, 0.810]; \texttt{Qwen2.5-7B},
SP$_{\mathrm{last}}$ 0.585 [0.347, 0.814]). Third, a flagged column is not
automatically uninformative: \texttt{gpt-oss-20b} on BFCL-v4 has only 10 successes yet VC
reaches 0.833 [0.749, 0.906], and \texttt{gpt-4o-mini} on telecom has 12 successes with
P(True) at 0.852 [0.732, 0.950]. The flag marks columns where an interval is
wide enough that a bolded point estimate should not be read as a resolved
ranking, not columns whose results are discarded.

The intervals also qualify the sub-chance white-box readings. Of the 240
white-box cells in the main grid, 91 fall below 0.5, but only 21 have
intervals lying entirely below 0.5, and 10 of those sit in columns that are
not flagged for minority-class size. Sign inversion is therefore an
established effect in a minority of cells rather than a property of every
sub-chance entry, and we describe it that way in
Section~\ref{sec:results}. The resolved cases concentrate in the mean and min
aggregations and on telecom, which holds 15 of the 21 against 5 on airline,
the pattern Figure~\ref{fig:whitebox-agg} shows.

\paragraph{Paired comparisons.}
Table~\ref{tab:airline-paired-bootstrap} reports the canonical marginal
intervals alongside task-paired intervals for the main airline differences,
where consistency is central to the argument and the task count is smallest.
For each model, we select the highest-AUROC reflexive and consistency scorer by
the observed airline point estimate, then hold those scorers fixed. The
difference intervals jointly resample both scorers over the same tasks. They
show that the consistency point estimates
are higher than the reflexive point estimates for all five models, but only the
\texttt{Qwen3.5-9B} gap separates cleanly from zero. Within the consistency family, the
ranking is also model-dependent: TER is the highest-AUROC airline consistency
scorer for three models, while ASC is highest for \texttt{gpt-oss-20b} and AEC
for \texttt{MiniMax-M3}, so the judged trajectory-equivalence variant does not uniformly
beat the non-judge consistency scores.



\begin{table*}[t]
\centering
\small
\setlength{\tabcolsep}{3.5pt}
\begin{tabular}{lcccc}
\toprule
Model & Top reflexive & Top consistency & Cons. $-$ refl. & TER $-$ top non-TER \\
\midrule
\texttt{gpt-4o-mini} & VC 0.550 [0.401, 0.707] & TER 0.647 [0.461, 0.810] & 0.096 [-0.142, 0.335] & 0.083 [-0.130, 0.323] \\
\texttt{gpt-oss-20b} & VC 0.703 [0.548, 0.827] & ASC 0.819 [0.685, 0.930] & 0.116 [-0.059, 0.293] & -0.156 [-0.349, 0.029] \\
\texttt{Qwen2.5-7B} & VC 0.400 [0.242, 0.558] & TER 0.581 [0.378, 0.759] & 0.181 [-0.035, 0.400] & 0.081 [-0.179, 0.334] \\
\texttt{Qwen3.5-9B} & VC 0.430 [0.292, 0.583] & TER 0.868 [0.728, 0.977] & 0.438 [0.244, 0.614] & 0.036 [-0.070, 0.142] \\
\texttt{MiniMax-M3} & P(True) 0.659 [0.435, 0.865] & AEC 0.742 [0.587, 0.877] & 0.083 [-0.181, 0.362] & -0.077 [-0.285, 0.120] \\
\bottomrule
\end{tabular}
\caption{Bootstrap uncertainty for the main airline comparison. For each model, the highest-AUROC reflexive and consistency scorers are selected by their observed airline point estimates. Their brackets reproduce the canonical marginal intervals from Table~\ref{tab:auroc-main-ci}. The two difference columns use a separate task-paired bootstrap with the scorers fixed and airline tasks resampled jointly. The final column compares TER against the highest-AUROC non-TER consistency scorer (NCP, FAC, ASC, ADC, or AEC) for the same model.}
\label{tab:airline-paired-bootstrap}
\end{table*}


\section{Trajectory Equivalence Judge}
\label{app:ter-prompt}

The Trajectory Equivalence Rate (TER) scorer uses \texttt{gemini-flash-lite}
(a distinct model family from every scored agent, so it raises no
self-preference concern) as the judge $J$, at temperature 0. For each task
the judge compares the greedy reference trajectory against each of the three
$T{=}0.7$ draws and returns a binary equivalence decision, which we average
into the rate. The judge sees only the two full transcripts, rendered as
alternating user/agent/tool messages with tool arguments included and tool
outputs truncated; it never sees the hidden task instructions or the
benchmark reward. The prompt
instructs the judge to decide net task-relevant outcome equivalence rather
than correctness: two trajectories are equivalent when they resolve the same
underlying request to the same net final state and answer, allowing
different clarifying questions, tool orderings, retries, and wording, and
requiring every part of a multi-part task to match. Judgments that fail to
parse as the expected JSON are dropped, and a task's rate averages its
remaining draws.

\paragraph{Judge validation against database state.} We validate the judge
against an objective ground truth on the database-graded domains, retail and
airline; telecom is graded on environment assertions rather than database state
and is excluded. Using $\tau^2$'s own database check, a pair of trajectories that
both reach the gold database state is equivalent, and a pair where exactly one
reaches it is not, which labels $1{,}365$ judged pairs across 545 tasks with
certainty; pairs where neither trajectory reaches gold cannot be separated by
the database check alone and are excluded. Against this ground truth the judge
attains precision 0.88 (95\% CI $[0.85, 0.90]$), recall 0.93 ($[0.90, 0.95]$),
and F1 0.90 ($[0.88, 0.92]$), with intervals from a task-clustered bootstrap.
The judge therefore agrees closely with what the database actually did and, in
particular, rarely misses genuinely equivalent trajectories. This validates the
state dimension of the judge's decisions; equivalence of the final natural
language answer is not captured by database state.

\paragraph{Sensitivity to the judge.} To test whether the choice of judge
affects the results, we re-score every airline pair, where TER is one of the
highest-ranking consistency scorers, with the larger \texttt{gemini-flash} in place of
\texttt{gemini-flash-lite}. The two judges agree on 91\% of pairs (Cohen's
$\kappa = 0.81$), and TER's airline AUROC is stable under the swap: it changes
by 0.021 on average across the five models and by at most 0.062, and the
headline result, \texttt{Qwen3.5-9B} at 0.868, holds at 0.894 under the stronger judge.
The stronger judge gives an equal or higher AUROC in four of the five cells, so
the cheaper judge does not appear to inflate TER's airline performance, and the
airline TER result is robust to this particular judge swap.

\begin{figure*}[t]
\centering
\includegraphics[width=\textwidth]{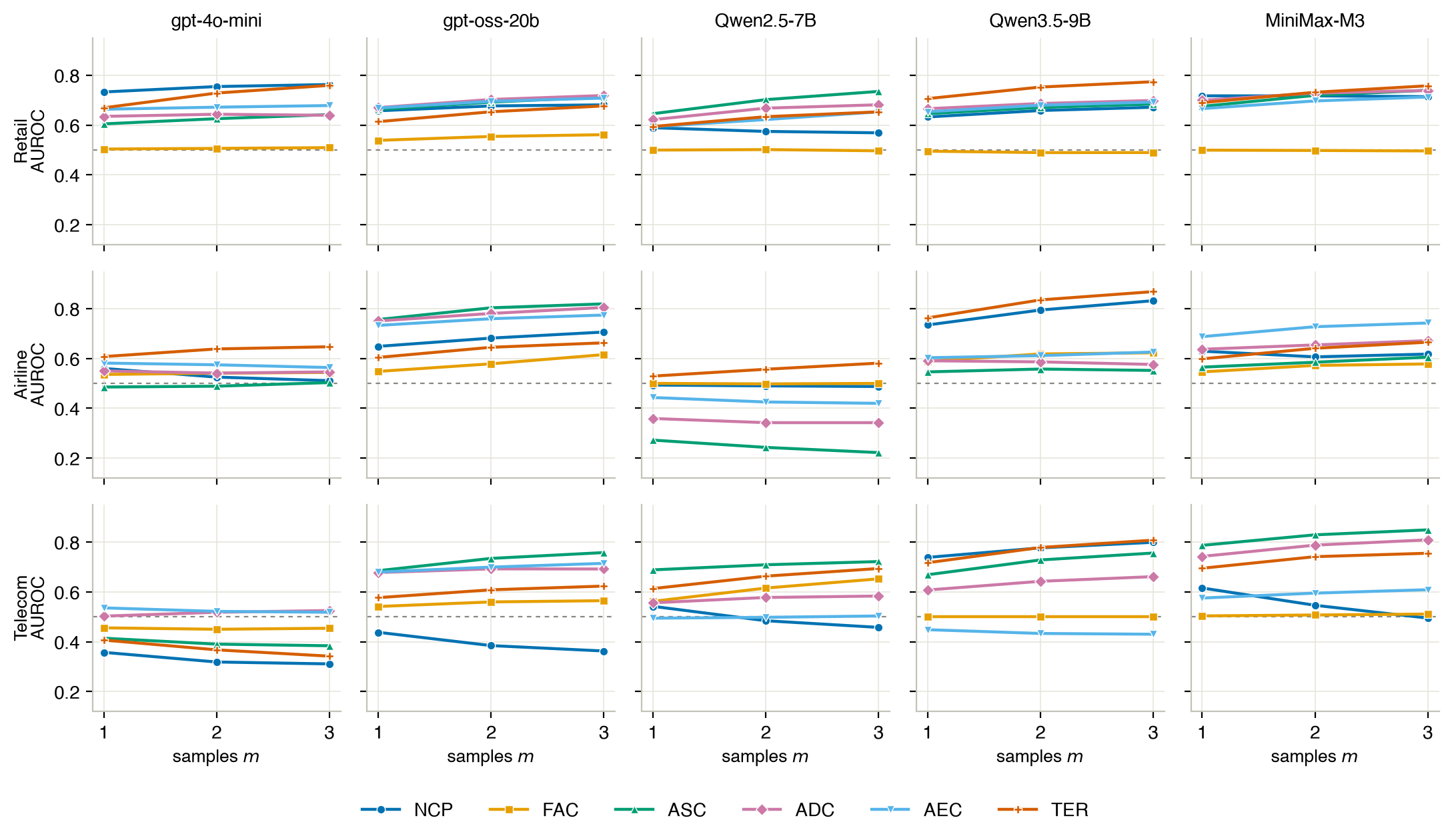}
\caption{Number-of-samples ablation for the six consistency scorers: AUROC for
predicting greedy-trajectory success as a function of the number of sampled
draws $m \in \{1,2,3\}$, per domain (rows) and model (columns). Each scorer is
recomputed from every $m$-subset of the three recorded draws and averaged over
subsets; the dashed line is chance. Discrimination generally rises with $m$ and
several curves are still improving at $m{=}3$ (TER and the action-set/edit scorers improve most),
mean first-action consistency remains close to chance, and the stuck-policy case
(\texttt{Qwen2.5-7B}, airline) drifts further below chance as $m$ grows.}
\label{fig:m-sensitivity}
\end{figure*}

\paragraph{Sensitivity to the number of samples.} The consistency scorers use
$m{=}3$ sampled draws. To gauge sensitivity to $m$, we recompute every
consistency scorer from each $m$-subset of the three recorded draws for $m \in
\{1, 2, 3\}$, averaging over subsets to remove draw-ordering dependence
(Figure~\ref{fig:m-sensitivity}). Discrimination generally increases with $m$:
for TER, mean AUROC across the fifteen $\tau^2$ model-domain cells rises from 0.63 at
$m{=}1$ to 0.68 at $m{=}3$ (airline 0.62 to 0.69, where consistency scorers have their largest point estimates, with
the headline \texttt{Qwen3.5-9B} airline cell climbing from 0.76 to 0.87), and the
action-set, distribution, and edit scorers improve similarly. Mean first-action
consistency remains close to chance, rising from 0.52 at $m{=}1$ to 0.54 at
$m{=}3$, because the first action is almost always a message to the user. The scorer ordering is stable
across $m$, and several curves are still improving at $m{=}3$, suggesting that
larger sampling budgets may yield further gains. The \texttt{Qwen2.5-7B} airline cell
drifts further below chance as $m$ grows, motivating the direct failure-failure diagnostic in
Appendix~\ref{app:stuck-policy-consistency}.
A full sweep to larger $m$ and to other decoding temperatures requires
additional sampling and is left to future work.

\section{Failure-Failure Consistency}
\label{app:stuck-policy-consistency}

Consistency can fail in two opposite ways: a model may be inconsistent because
it samples both successful and unsuccessful plans, or it may be highly
consistent because it repeats the same plan even when that plan fails. To
diagnose this stuck-policy mechanism directly, we partition each
original-vs-sample pair by the success labels of the two trajectories and
separately group tasks by whether all four trajectories succeed, some succeed,
or none succeed. Table~\ref{tab:stuck-policy-pair-types} reports pair-level
TER and AEC, and Table~\ref{tab:stuck-policy-task-groups} reports the same
scores averaged at the task level.

The diagnostic supports a cautious version of the stuck-policy interpretation.
Success-success pairs are by far the most consistent under TER (0.95 across
$\tau^2$, 0.87 on airline), but failure-failure pairs are not random: their TER is
0.40 across $\tau^2$ and 0.37 on airline, and AEC remains near 0.52 in both cases.
Never-solved tasks likewise retain moderate action consistency. For the
\texttt{Qwen2.5-7B} airline cell, where consistency AUROC is below chance, there is only
one always-solved task, while 31 of 50 tasks are never solved; AEC is similar
for never-solved and mixed tasks (0.50 vs. 0.46), so agreement often reflects
repeated unsuccessful behavior. The analysis does not show that stuck policies
dominate every consistency failure: failure-failure TER is still much lower
than success-success TER, and TER on \texttt{Qwen2.5-7B} airline is low overall. It does
show why agreement alone is ambiguous in agentic settings and why
outcome-aware or failure-mode-aware consistency is needed.


\begin{table*}[t]
\centering
\small
\begin{tabular}{llccc}
\toprule
Condition & Scorer & Success--success & Success--failure & Failure--failure \\
\midrule
All $\tau^2$ & TER & 0.95 (1219) & 0.22 (1029) & 0.40 (1918) \\
All $\tau^2$ & AEC & 0.74 (1219) & 0.58 (1030) & 0.52 (1921) \\
\midrule
Airline & TER & 0.87 (251) & 0.21 (174) & 0.37 (325) \\
Airline & AEC & 0.68 (251) & 0.52 (174) & 0.52 (325) \\
\midrule
\texttt{Qwen2.5-7B} airline & TER & 0.54 (13) & 0.17 (29) & 0.29 (108) \\
\texttt{Qwen2.5-7B} airline & AEC & 0.48 (13) & 0.47 (29) & 0.49 (108) \\
\bottomrule
\end{tabular}
\caption{Consistency by pair outcome. Entries are mean original-vs-sample consistency scores, with the number of judged or compared pairs in parentheses. TER is binary judged trajectory equivalence; AEC is action-edit consistency. Nonzero failure--failure agreement shows that consistency can reflect repeated failing behavior as well as repeated successful behavior, although failure--failure TER remains much lower than success--success TER.}
\label{tab:stuck-policy-pair-types}
\end{table*}

\begin{table*}[t]
\centering
\small
\begin{tabular}{llccc}
\toprule
Condition & Scorer & Always solved & Sometimes solved & Never solved \\
\midrule
All $\tau^2$ & TER & 0.96 (291) & 0.40 (597) & 0.40 (502) \\
All $\tau^2$ & AEC & 0.74 (291) & 0.60 (597) & 0.50 (502) \\
\midrule
Airline & TER & 0.91 (60) & 0.39 (101) & 0.35 (89) \\
Airline & AEC & 0.67 (60) & 0.57 (101) & 0.51 (89) \\
\midrule
\texttt{Qwen2.5-7B} airline & TER & 0.67 (1) & 0.24 (18) & 0.30 (31) \\
\texttt{Qwen2.5-7B} airline & AEC & 0.52 (1) & 0.46 (18) & 0.50 (31) \\
\bottomrule
\end{tabular}
\caption{Consistency by task solvability across the greedy trajectory and three sampled draws. Entries are mean task-level original-vs-sample consistency scores, with tasks in parentheses. Never-solved tasks retain moderate consistency, illustrating why consistency can be ambiguous between repeated success and repeated failure.}
\label{tab:stuck-policy-task-groups}
\end{table*}

\section{Agent-Only Prefix Consistency Ablation}
\label{app:prefix-action-consistency}

The full-trajectory consistency scorers in the main experiments resample both
the agent and the simulated user. Although the user simulator is fixed across
models and run at temperature 0, full-trajectory resampling still measures
variation in the joint agent--user interaction. To isolate the agent term, we
run a targeted prefix-conditioned ablation on the airline domain with
\texttt{gpt-4o-mini}, where consistency is central to the main results.

For each recorded greedy airline trajectory, we hold each observed agent prefix
fixed and resample only the next agent action five times at $T{=}0.7$. The
sampled actions are not executed, and the user simulator is not called. This
therefore estimates local policy stability,
$p(A_t \mid F_{<t})$, rather than full trajectory consistency. We compare each
sampled next action with the action actually taken in the recorded trajectory
using action type and action name agreement, then aggregate agreement over the
trajectory. These scores are evaluated against the original greedy trajectory's
success label.

\begin{table}[t]
\centering
\small
\begin{tabular}{lcc}
\toprule
Scorer & AUROC & Mean score \\
\midrule
Mean action-type agreement & 0.568 & 0.940 \\
Min action-type agreement & 0.581 & 0.560 \\
Mean action-name agreement & 0.588 & 0.920 \\
Min action-name agreement & 0.577 & 0.488 \\
Mean tool-name agreement & 0.580 & 0.912 \\
\bottomrule
\end{tabular}
\caption{Agent-only prefix-conditioned consistency ablation on $\tau^2$-bench
airline with \texttt{gpt-4o-mini}. We resample five next actions for each of 647
recorded agent decision points, holding the transcript prefix fixed and making
no simulator or environment calls. AUROC predicts the original greedy
trajectory's success over 50 tasks (12 successes, 38 failures).}
\label{tab:prefix-action-consistency}
\end{table}

Table~\ref{tab:prefix-action-consistency} shows that the agent-only local
stability point estimates are positive but modest: the largest is 0.588 AUROC
for mean action-name agreement. With only 50 tasks (12 successes), this
ablation is underpowered and should be read as a diagnostic check rather than
confirmatory evidence that the agent-only term is predictive. The result is
consistent with the possibility that some consistency signal remains when the
simulated user is held fixed, but it does not establish that local action-name
stability alone reliably predicts trajectory failure. The gap from
full-trajectory consistency is also expected because local next-action
agreement does not execute the sampled actions and cannot reflect downstream
recovery, cumulative state changes, or outcome equivalence across alternative
plans.

\section{Action Interface and Native Tool-Calling Ablation}
\label{app:native-tool-calling}

\paragraph{The action interface as measurement apparatus.}
Every white-box score in this study is an aggregate of token probabilities over
the tokens of a committed action, so the study needs an action whose tokens are
observable and delimited in the same way for every model. We therefore fix the
action interface exactly as we fix the user simulator, and for the same reason:
it is a measurement apparatus, and varying it across models would confound the
cross-model comparison. In our interface the domain policy and tool schemas are
given in the system prompt and the agent emits one action as text per turn, so
the action tokens appear in the generation and carry log probabilities. Native
tool calling did not expose a comparable action-token surface across our
model/provider stack, and where it was available its coverage was
provider-dependent, so it could not support the white-box part of the study.

This choice is usually left implicit in agent UQ. The estimators of
\citet{oh2026uncertaintyquantificationllmagents}, which our white-box scores
follow, are defined as per-token averages over the action of each turn, but the
procedure that makes those action tokens observable is not specified; this
matters because $\tau^2$-bench commonly supports native tool-calling agents, for
which provider APIs need not expose comparable token probabilities over tool
payloads. We therefore state the interface explicitly and report an ablation of
it rather than leaving the measurement surface undefined. The trade-off is the
one made visible here: the interface that makes action-token probabilities
available is not the interface most deployments use, which is itself a finding
about the practical availability of white-box UQ for agents
(Table~\ref{tab:cost}).

\paragraph{Interface ablation.}
To test whether the non-white-box results are an artifact of the text-action
representation, we ran a targeted interface ablation on $\tau^2$-bench retail with
\texttt{gpt-4o-mini}.

The ablation keeps the trajectory protocol fixed: one greedy reference
trajectory per task at $T{=}0$ and three sampled trajectories per task at
$T{=}0.7$. The only intended change is the agent interface. In the text-action
condition, the model emits textual actions that are parsed by the runner; in
the native condition, the model calls $\tau^2$ tools through the provider's native
tool-calling API. We do not use this ablation to evaluate white-box scores,
because the text-action interface is what makes action-token log probabilities
observable and comparable across models; native tool calls do not provide the same
action-token log-probability surface. The native ablation instead tests whether
the black-box consistency methods are sensitive to the text-action interface.

Native tool calling completed all 114 retail tasks. The greedy native run
succeeded on 46/114 tasks (40.4\%), and the three sampled native runs succeeded
on 137/342 trajectories (40.1\%). The corresponding text-action \texttt{gpt-4o-mini}
retail greedy run succeeded on 42/114 tasks (36.8\%). The 3.5 percentage-point
anchor gap is not statistically significant at the 5\% level under a
difference-of-proportions test, so for this model and domain we do not find
evidence that changing the interface substantially changes the base task
success rate.

\begin{table}[t]
\centering
\small
\begin{tabular}{lcc}
\toprule
Scorer & Text-action AUROC & Native AUROC \\
\midrule
FAC & 0.509 & 0.498 \\
ASC & 0.642 & 0.625 \\
ADC & 0.639 & 0.626 \\
AEC & 0.678 & 0.681 \\
NCP & 0.762 & 0.716 \\
TER & 0.759 & 0.761 \\
\bottomrule
\end{tabular}
\caption{Native tool-calling ablation on $\tau^2$-bench retail with \texttt{gpt-4o-mini}.
Each AUROC predicts the greedy trajectory's success from consistency between
the greedy trajectory and three $T{=}0.7$ samples. FAC, ASC, ADC, and AEC are
action-type consistency scores; NCP compares final messages with an NLI model;
TER uses the trajectory-equivalence judge. One \texttt{Gemini} empty response among the
342 native TER pair judgments was excluded from the TER average.}
\label{tab:native-tool-calling-ablation}
\end{table}

Table~\ref{tab:native-tool-calling-ablation} shows that the highest-AUROC
trajectory-structured signals transfer cleanly across interfaces. TER is
essentially unchanged, with AUROC 0.759 under the text-action interface and 0.761
under native tool calling. AEC, the highest-AUROC symbolic action-consistency
score in this ablation, is also stable (0.678 versus 0.681). The simpler
action-name measures remain in the same range, with FAC near chance in both
conditions. NCP decreases from 0.762 to 0.716, suggesting that final-message
consistency is more sensitive to interface-induced differences in dialogue
style or final-answer wording than the trajectory-level and action-structured
signals.

This ablation supports the interpretation that the non-white-box findings are
not merely artifacts of the text-action interface. It should be read as a
robustness check rather than a replacement experiment: it covers one model and
one domain, and native tool-calling reliability is itself model- and
provider-dependent. We did not extend the native ablation to the Together-hosted
open models because native execution fails to support the white-box measurement
we study and, in preliminary runs, was dominated by interface compatibility
rather than task solving: in a \texttt{Qwen3.5-9B} retail attempt, the agent rarely
produced usable native tool calls and task success was far below the text-action
condition. In that regime, a cross-model native comparison would partly measure
provider/tool-call compatibility for these model sizes rather than uncertainty
estimation.

\section{Prompts}
\label{app:prompts}


We reproduce the fixed prompts for the reflexive scorers and the judged
consistency scorer verbatim. For the reflexive scorers, the scored material is
the transcript prefix ending at the agent's last action, before any subsequent
user confirmation of the outcome. For the judge, the scored material is the two
full trajectories. In each case transcripts are rendered as alternating user,
agent, and tool messages with tool arguments included and tool outputs
truncated; the judge additionally wraps its two transcripts in
\texttt{<trajectory\_A>} and \texttt{<trajectory\_B>} tags.

\begin{table*}[tb]
\centering
\begin{tcolorbox}[colback=blue!5!white, colframe=blue!50!black,
  title={\textsc{P(True) Scoring Prompt}}]
\footnotesize\ttfamily
You are a binary evaluator for an AI customer-service agent. \\[1ex]
Your task is to decide whether the agent successfully completed the user's task. \\[1ex]
The agent succeeds only if it accomplished the user's goals correctly and completely, following all constraints in the task. Partial completion, wrong arguments, skipped confirmations, or actions the task did not ask for count as failure. \\[1ex]
Return exactly one word and nothing else: \\
TRUE if the agent successfully completed the task. \\
FALSE if the agent did not successfully complete the task. \\[1ex]
Do not explain. Do not show reasoning. Do not think step by step. The first output token must be either TRUE or FALSE.
\end{tcolorbox}
\caption{P(True) scoring prompt. Confidence is the probability the scoring
model places on the \texttt{TRUE} token (Section~\ref{sec:methods}).}
\label{tab:ptrue-prompt}
\end{table*}

\begin{table*}[tb]
\centering
\begin{tcolorbox}[colback=blue!5!white, colframe=blue!50!black,
  title={\textsc{Verbalized Confidence Prompt}}]
\footnotesize\ttfamily
You are evaluating an AI customer-service agent. \\[1ex]
Your task is to decide whether the agent successfully completed the user's task, and to state how confident you are. \\[1ex]
The agent succeeds only if it accomplished the user's goals correctly and completely, following all constraints in the task. Partial completion, wrong arguments, skipped confirmations, or actions the task did not ask for count as failure. \\[1ex]
Provide your best guess of whether the agent successfully completed the task (Yes or No) and the probability that this guess is correct, as a number from 0.0 to 1.0. Give ONLY the guess and probability, in exactly this format and with no other words or explanation: \\
Guess: <Yes or No>, Probability: <a number between 0.0 and 1.0>
\end{tcolorbox}
\caption{Verbalized confidence prompt. The success confidence is the stated
probability when the guess is \texttt{Yes} and its complement when the guess
is \texttt{No}.}
\label{tab:vc-prompt}
\end{table*}

\begin{table*}[tb]
\centering
\begin{tcolorbox}[colback=blue!5!white, colframe=blue!50!black,
  title={\textsc{Trajectory Equivalence Judge Prompt}}]
\footnotesize\ttfamily
You compare two completed AI customer-service agent trajectories. \\[1ex]
The trajectories come from $\tau^2$-style tasks: a user chats with an agent, and the agent may call tools/APIs that inspect or modify customer-service state such as reservations, orders, accounts, devices, bills, settings, or tickets. The two trajectories start from the same underlying task, but the user is simulated separately in each one, so the user's wording, tone, and the order in which information comes up will differ even when the request is the same. The labels A and B are arbitrary; presentation order carries no information. \\[1ex]
Your job is trajectory-only outcome equivalence. Do NOT judge whether either trajectory is correct for the original hidden task, and do NOT infer the official benchmark reward. You are only comparing the two trajectories against each other. \\[1ex]
Two trajectories are equivalent if, based only on what is shown, they resolve the same underlying request(s) to the same net task-relevant outcome. They can be equivalent even if they ask different clarifying questions, inspect extra information, call tools in a different order, repeat checks, retry failed calls, or use different wording. \\[1ex]
Judge the NET final effect, not process identity: \\
- Compare the net state changes after any corrections: a write that is later reversed or amended counts only by its end result. \\
- If both trajectories make state changes, they are equivalent when the net changes have the same effect on the same relevant entities. If the task has several parts (for example a cancellation and an exchange and a question), every task-relevant part must reach a matching outcome. \\
- If both trajectories give final answers and make no state changes, they are equivalent when the answers agree on the task-relevant facts (amounts, eligibility, identities, outcomes), regardless of wording or level of detail. \\
- Extra harmless reads, checks, confirmations, or explanatory wording never break equivalence. \\
- If NEITHER trajectory resolves the request (no task-relevant state change and no substantive answer), they are equivalent only when their final dispositions match: both escalate/transfer, both refuse for the same stated reason, or both end unresolved at the same apparent point. An unresolved or truncated trajectory is never equivalent to one that resolves the request. \\[1ex]
Treat trajectories as not equivalent if the users make substantively different requests (not just different phrasing), the final answers disagree on a task-relevant fact, one refuses/escalates while the other completes, one performs a net task-relevant state change that the other omits, or their net changes touch different entities/fields in ways that would lead to different customer-service outcomes. \\[1ex]
Return ONLY valid JSON with exactly this schema: \\
\{ \\
\quad "equivalent": true or false, \\
\quad "confidence": number between 0 and 1, your probability that this judgment is right \\
\}
\end{tcolorbox}
\caption{Trajectory equivalence judge prompt. TER averages the judge's binary
decision over the sampled draws (Appendix~\ref{app:ter-prompt}).}
\label{tab:ter-judge-prompt}
\end{table*}

\end{document}